\documentclass[a4paper,fleqn]{cas-sc}

\usepackage{float}
\usepackage{graphicx}
\usepackage{subcaption}
\usepackage{amsmath}
\usepackage{graphicx}

\usepackage[authoryear]{natbib}

\def\tsc#1{\csdef{#1}{\textsc{\lowercase{#1}}\xspace}}
\tsc{WGM}
\tsc{QE}

\begin{document}
\let\WriteBookmarks\relax
\def\floatpagepagefraction{1}
\def\textpagefraction{.001}

\shorttitle{}    

\shortauthors{Yinxiang Yu et~al.}

\title [mode = title]{HD-DinoMoE: A Class-Aware Hierarchical Dual Mixture-of-Experts Network for Scleral Anomaly Segmentation in Complex Acquisition Scenarios}  

\tnotemark[1,2]

\tnotetext[1]{This work was supported by the Natural Science Foundation of Liaoning Province 
(Nos. 2025-MS-134 and 2022-MS-353) and the Basic Scientific Research Project of the 
Education Department of Liaoning Province (Nos. LJ232410146053 and LJKMZ20220640).}

\tnotetext[2]{This work has been submitted to \textit{Medical Image Analysis} for possible publication.}

%

\author[1]{Yinxiang Yu}[orcid=0009-0003-2318-7189]
\credit{Conceptualization, Methodology, Software, Data curation, Formal analysis, Investigation, Visualization, Writing -- original draft}
\ead{xiangzi_pull_car@163.com}

\author[1]{Maoxiang Chu}[orcid=0000-0003-2358-9455]
\cormark[1]
\credit{Conceptualization, Resources, Supervision, Project administration, Writing -- review \& editing}
\ead{chu52_2004@163.com}

\author[1]{Qi Niu}
\credit{Data curation, Investigation}
\ead{25210854060556.ustl@vip.163.com}

\author[1]{Guanghu Liu}
\credit{Writing -- review \& editing}
\ead{huia104@163.com}

\author[1]{Wei Xu}
\credit{Data curation, Investigation}
\ead{xuwei020203@gmail.com}

\author[1]{Haotian Wang}
\credit{Data curation, Investigation}
\ead{wanyanwang59@gmail.com}

\author[1]{Zhi Chen}
\credit{Data curation, Investigation}
\ead{cz1073755937@gmail.com}

\author[1]{Yutian Zhu}
\credit{Data curation, Investigation}
\ead{ytz1024201@163.com}

\author[1]{Yuelong Fan}
\credit{Data curation, Investigation}
\ead{fyl010912@163.com}

\author[1]{Guanghao Liao}
\credit{Data curation, Investigation}
\ead{24210811000450@stu.ustl.edu.cn}

\affiliation[1]{organization={School of Electronic and Information Engineering, University of Science and Technology Liaoning},
            addressline={No.189 Qianshan Middle Road, Lishan District}, 
            city={Anshan},
            postcode={114051}, 
            state={Liaoning},
            country={China}}
\cortext[1]{Maoxiang Chu}


\begin{abstract}
Traditional Chinese Medicine (TCM) ocular inspection provides empirical cues for assessing scleral surface anomalies, but its clinical use remains subjective and difficult to quantify. 
To support intelligent and quantifiable ocular inspection, this study presents the TCM-inspired Artificial Intelligence Ocular Auxiliary Diagnosis System (TAO) and focuses on pixel-level scleral surface anomaly segmentation. For clinical and user-acquired images affected by multi-source distributional discrepancies, diverse anomaly morphologies, and scleral specular reflection (SSR), we propose HD-DinoMoE, a class-aware hierarchical dual mixture-of-experts network. HD-DinoMoE combines class-aware dual-stream DINOv3 feature fusion with class-specific multi-expert decoding to segment Vessels, Yellow and Black Spots, and Blood Spots. 
A three-stage backbone-frozen routing strategy stabilizes dual-backbone adaptation; Progressive Confidence Penalty (PCP) Loss reduces high-confidence false positives and segmentation leakage in SSR regions; and Class-Aware Adaptive Sample Weighting (CA-ASW) balances sample- and class-level training contributions. We further construct the Multi-label Scleral Anomaly Segmentation Dataset (ML-SASD), a new benchmark with Clinical, Wild, and Mix settings and pixel-wise annotations for three anomaly categories. 
On ML-SASD-Mix, HD-DinoMoE achieves a mean Dice of 72.11\% and a mean Intersection-over-Union of 58.44\%, while maintaining favorable boundary localization and specular-region false-positive control. 
It also shows competitive generalization on the Vessels subset of the public SBVPI dataset. 
These results indicate that HD-DinoMoE provides a feasible segmentation solution for TAO under complex acquisition scenarios. 
The code and data access information are available at \url{https://github.com/FX-CMX/HD-DinoMoE}.

\end{abstract}




\begin{keywords}
Scleral segmentation \sep Semantic segmentation \sep Vision foundation model \sep Mixture-of-experts
\end{keywords}

\maketitle

\section{Introduction}\label{Introduction}
The sclera and its superficial microvascular network serve as critical visual indicators of ocular health and have also been linked to various systemic diseases.
Accumulating medical evidence indicates that variations in scleral biomechanical properties, superficial vascular congestion, microvascular morphology, and localized hemorrhage can provide important clues for the non-invasive auxiliary analysis of conditions such as glaucoma, diabetic microvascular complications, anemia, and cardiovascular risk\citep{wenAdvances2026, linImpact2025, yaoArtificial2025, diagnostics13040648, DIMAURO2023104489, ramos-sotoNoninvasive2025, shiDeeplearning2025}.
Therefore, robust, fine-grained, and quantifiable pixel-level segmentation of scleral surface anomalies constitutes a critical foundation for translating ocular image-based auxiliary diagnostic analysis into practical clinical applications.

Within the framework of traditional Chinese medicine (TCM), scleral surface anomalies are also considered to hold important diagnostic value.
For instance, the theoretical framework of Zheng's ocular inspection seeks to map specific ocular regions to organ-related functional states in TCM by evaluating abnormal collateral vessels, pigmentation, and blood spots across distinct scleral sectors.
 
Concurrently, Chinese ocular acupuncture partitions the ocular region into ``eight regions and thirteen acupoints'', thereby establishing the sclera and its surrounding structures as an important spatial reference framework for TCM diagnosis and treatment\citep{zheng_traditional_eye_diagnosis, cheFormation2005}.
However, whether identifying early-stage lesions in modern medicine or performing empirical syndrome differentiation through TCM ocular inspection, the assessment remains heavily contingent upon clinicians' subjective expertise.
Consequently, there remains a clear lack of objective methodologies for automated localization, regional quantification, and auxiliary diagnostic analysis of scleral surface anomalies.
Therefore, transitioning scleral anomaly evaluation from empirical judgment toward intelligent and quantifiable analysis has emerged as a critical imperative for modern computer-aided ocular diagnostic systems.

In recent years, interdisciplinary research at the intersection of artificial intelligence (AI) and ophthalmology has steadily expanded from disease screening to anatomical structure segmentation, subtle lesion detection, and multimodal ophthalmic image analysis.
Particularly in fundus image analysis, deep learning models have demonstrated robust performance across tasks such as diabetic retinopathy screening, macular lesion grading, fine-grained retinal vessel segmentation, and subtle lesion recognition against complex backgrounds.
These advances highlight the strong potential of computer vision models in supporting computer-aided ophthalmic diagnosis\citep{linImpact2025, guoSAUNet2021, wangNew2024, xieMultimodal2025}.
However, existing ophthalmic AI studies have primarily focused on fundus analysis and conventional scleral region/contour segmentation, leaving the pixel-level parsing of scleral surface anomalies largely underexplored.
Consequently, a dedicated multi-class and multi-label segmentation framework tailored to the automated workflow of TCM ocular inspection remains noticeably lacking.

\begin{figure}[pos=htbp,width=\textwidth]
	\centering
	\includegraphics[width=\textwidth]{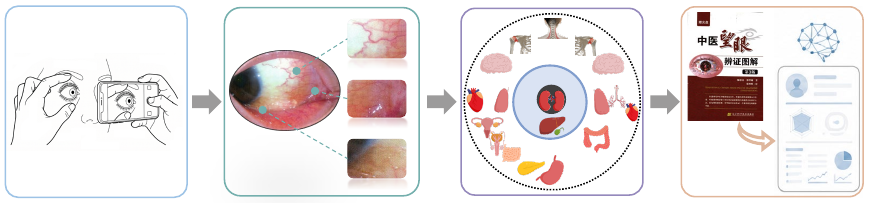}
	\caption{Functional overview of the TCM-inspired AI Ocular Auxiliary Diagnosis System}
	\label{fig:TAO-1}
\end{figure}

To facilitate the intelligent application of TCM ocular inspection, this study introduces the Traditional Chinese Medicine-inspired Artificial Intelligence Ocular Auxiliary Diagnosis System (TAO).
As illustrated in Fig.~\ref{fig:TAO-1}, the proposed pipeline comprises four interconnected components.
First, multi-gaze ocular images are captured using mobile devices to improve the coverage of the scleral surface.
Second, a cascaded semantic segmentation strategy is adopted for fine-grained scleral anomaly detection.
Specifically, the initial phase performs ocular structure segmentation to isolate the sclera region and filter out irrelevant structures such as the iris and background, whereas the subsequent phase identifies and delineates heterogeneous anomaly patterns within the sclera region in a multi-label manner, thereby characterizing their spatial distributions.
Third, the system constructs a pupil-centered polar coordinate system and, following TCM ocular inspection, establishes a potential eye-organ mapping model between scleral surface anomalies and human organs, producing regional anomaly scores.
Fourth, the system provides ocular images, segmentation results, regional scoring information, and basic user information, such as self-reported symptoms, as structured inputs to a multimodal large language model (MLLM) with knowledge-enhanced fine-tuning.
Combined with authoritative TCM literature as the knowledge base, the MLLM generates a medical report with auxiliary recommendations.

However, translating the TAO system into real-world deployment faces multifaceted challenges spanning both data and algorithmic dimensions:

\textbf{1) Extreme scarcity of pathology-oriented datasets:}
Although several ocular surface and sclera segmentation datasets have been publicly released\citep{vitekComprehensive2020, proencaUBIRISv22010, demarsicoMobile2015, dasMultimodal2017, dasMultiangle2014, vitekSSBC2020}, they were predominantly tailored for ocular biometrics or eye-tracking tasks.
Consequently, their annotations are generally centered on macro-anatomical scleral regions or contours, rather than multi-class pixel-level semantic masks for scleral surface anomalies.
As a result, benchmark datasets dedicated to fine-grained pathological scleral analysis remain exceedingly rare.

\textbf{2) Generalization and accuracy degradation in complex scenarios:}
In practical applications, the TAO system is implemented on user-facing mobile devices.
Users may acquire ocular images under various unconstrained daily conditions, such as indoor low light, backlighting, and strong outdoor illumination, which can introduce imaging artifacts including illumination variation, local defocus, and motion blur caused by hand tremor\citep{demarsicoMobile2015, proencaUBIRISv22010, vitekSSBC2020}.
These factors lead to pronounced cross-environment, cross-device, and cross-quality heterogeneity in mobile ocular images.
When facing data heterogeneity and domain shift caused by non-professional user-acquired imaging, existing medical image segmentation models are prone to brittle feature representations and degraded generalization performance.
This problem is particularly challenging for scleral anomaly regions with subtle morphologies and irregular boundaries, where models still struggle to achieve robust and accurate segmentation\citep{10.1007/978-3-030-32245-8_69, guanDomain2022, Lee_2020_CVPR, gaoMedical2025}.

\textbf{3) Spatial occlusion and segmentation leakage caused by Scleral Specular Reflection (SSR):}
In user-acquired ocular images, specular reflection on the scleral surface is a major confounding factor for accurate boundary localization of abnormal regions.
The key challenge of specular reflection patches lies in their strong spatial overlap with, and occlusion of, true lesion regions.
To preserve the connectivity and structural integrity of predicted target regions, existing medical segmentation networks are prone to segmentation leakage when handling such spatial overlap.
Specifically, specular-reflection-occluded regions may be incorrectly merged with true anomaly regions as a single connected target, leading to substantial boundary expansion and topological distortion of lesion masks\citep{s23020974, 10.1007/978-3-032-05127-1_19}.

Since the acquisition module of the TAO system adopts a multi-view acquisition mechanism, anomaly regions truncated by specular reflection in one view can be cross-compensated in other views through subsequent spatial mapping, thereby improving the system's fault tolerance.
Therefore, the model is not required to perform high-risk speculative inference within specular reflection regions.
Instead, the core requirement of the system is to preserve high-confidence true anomaly regions as much as possible while reducing the risk of false positives and mis-segmentation caused by ambiguous regions such as specular reflections.

To address these challenges, this study introduces HD-DinoMoE, a Class-Aware Hierarchical Dual Mixture-of-Experts Network tailored for complex acquisition scenarios, and comprehensively evaluates its effectiveness on the newly constructed Multi-label Scleral Anomaly Segmentation Dataset (ML-SASD) and the public SBVPI benchmark.
Quantitative evaluations demonstrate that HD-DinoMoE achieves strong segmentation performance and shows promising cross-scenario generalization under heterogeneous multi-source data settings.
The primary contributions of this study are outlined as follows:
\begin{itemize}
\item \textbf{Class-Aware Dual-Stream Gated Fusion Encoder (CA-DSGF) and Three-Stage Backbone-Frozen Routing Alignment Strategy (TS-BFRA):}
To address the structural, textural, and cross-scenario heterogeneity of different scleral anomaly categories, CA-DSGF integrates two types of pretrained weights, namely SAT-DINOv3-L and LVD-DINOv3-L, in parallel, and uses a class-aware gating mechanism to adaptively learn class-specific dual-stream feature fusion weights. Furthermore, TS-BFRA improves the adaptation stability of the dual DINOv3 branches on small-scale medically annotated data through phase-wise backbone freezing and routing optimization.

\item \textbf{Class-Specific Multi-Expert Decoder (CS-MED):}
To improve decoding adaptability under diverse acquisition environments and complex anomaly morphologies, this study constructs a Mixture-of-Experts (MoE) decoding framework comprising multiple heterogeneous decoding experts. Based on the class-fused features output by CA-DSGF, CS-MED performs expert-weighted prediction and adaptively adjusts expert contributions according to input features, thereby improving the model's segmentation capability for multi-scale and multi-morphology scleral anomaly regions.

\item \textbf{Progressive Confidence Penalty Loss (PCP Loss):}
To address segmentation leakage caused by Scleral Specular Reflection (SSR), PCP Loss treats SSR regions as special negative-sample regions and applies progressive loss weighting according to the model's prediction confidence in these regions. This design suppresses high-confidence false positives within SSR regions while reducing the risk of over-suppressing true lesion regions.

\item \textbf{Class-Aware Adaptive Sample Weighting (CA-ASW):}
To address imbalanced training contributions caused by class distribution, sample difficulty, and acquisition-environment differences in ML-SASD, CA-ASW dynamically adjusts loss weights from both sample- and class-level dimensions. This strategy encourages the model to focus more on representative hard samples and cross-scenario samples during training, thereby improving training stability and generalization performance under multi-source data conditions.

\item \textbf{Multi-label Scleral Anomaly Segmentation Dataset (ML-SASD):}
To address the scarcity of pixel-level annotated data for multi-class scleral surface anomalies in this field, this study constructs and releases a new multi-label scleral surface anomaly segmentation dataset for mobile-based auxiliary diagnostic scenarios. The dataset contains three anomaly categories, namely Vessels (Ve), Yellow and Black Spots (YBS), and Blood Spots (BS), and provides a new data benchmark for subsequent related research.

\end{itemize}

\section{Related Works}\label{related}
\subsection{Scleral Image Analysis and Medical Auxiliary Diagnostic Applications}

Sclera segmentation is a foundational task in ocular image analysis and has received sustained attention in ocular biometrics, eye tracking, and ocular surface structure analysis\citep{vitekComprehensive2020, maquilingZeroShot2024, maLayered2018}.
Early studies mainly focused on scleral region localization under unconstrained environments, aiming to accurately extract the scleral region from ocular images containing interference from the periocular area, iris, eyelashes, and background, thereby providing a stable structural prior for subsequent identity recognition or gaze estimation.
Notably, \citet{vitekComprehensive2020} released the SBVPI (Sclera Blood Vessels, Periocular and Iris) dataset and organized the Sclera Segmentation Benchmarking Competition (SSBC), which provided an important public benchmark for this field and substantially promoted the robustness of sclera segmentation models under complex acquisition environments.

In recent years, several studies have begun to explore the use of scleral or conjunctival-scleral imagery for the non-invasive auxiliary diagnostic analysis of systemic diseases.
\citet{diagnostics13040648} released scleral image data for the benign-malignant classification of pulmonary tumors and constructed a multi-instance learning framework based on these data.
Based on the Eyes-defy-anemia dataset, \citet{DIMAURO2023104489} and \citet{ramos-sotoNoninvasive2025} investigated anemia detection using scleral/vascular color characteristics and Vision Transformer representations, respectively.
\citet{shiDeeplearning2025} further constructed a coronary heart disease risk assessment system using more than 5,000 scleral photographs.
Collectively, these studies suggest that visual cues in scleral imagery, including vascular patterns, color characteristics, and pigmentation, may have potential value for auxiliary diagnostic applications.

\subsection{Sclera-related Datasets}
In recent years, with the growing interest in ocular feature analysis in computer vision, several high-quality ocular and scleral datasets have been released.
For example, UBIRIS.v2\citep{proencaUBIRISv22010} and MICHE-I\citep{demarsicoMobile2015} were among the first to explore ocular image acquisition under unconstrained environments, such as mobile, long-distance, and mobile-device scenarios, which greatly promoted the development of mobile biometrics.
To address the limitations of single-view acquisition and restricted feature representation, datasets such as SMD\citep{dasMultimodal2017} and MASD\citep{dasMultiangle2014} were proposed to explore multi-angle and multimodal scleral feature extraction.
The MOBIUS dataset further provides large-scale benchmark data for unconstrained mobile natural scenarios.
These early datasets laid a solid foundation for coarse-grained scleral region localization and ocular biometrics.
Among various sclera-related datasets, the release of the SBVPI dataset\citep{vitekComprehensive2020} represents an important milestone.
Unlike previous datasets that mainly focused on the overall scleral contour or iris region, SBVPI first introduced fine-grained pixel-level annotations for the scleral microvascular network, namely Vessels.
Since superficial vascular congestion and morphological changes in the sclera are often important clinical indicators of systemic conditions, such as metabolic abnormalities or organ-related functional imbalance in TCM, the vessel annotations provided by SBVPI offer valuable data support for exploring sclera-based auxiliary diagnostic analysis and establish an early feature benchmark for the scleral pathological analysis task considered in this study.

\subsection{Applications of Vision Foundation Models in Medical Image Segmentation}

Vision Foundation Models (VFMs), with general visual representations learned from large-scale pretraining, offer a promising paradigm for alleviating the generalization bottlenecks caused by limited medical image annotations.
A representative foundation model is the Segment Anything Model (SAM) introduced by Meta\citep{kirillovSegment2023}, which demonstrated notable cross-domain potential in medical images through large-scale natural-image pretraining and prompt engineering.
Recently, \citet{wuOneprompt2024} proposed a SAM adaptation paradigm for medical images, further supporting the applicability of foundation models across diverse medical modalities.
The recent DINOv3 architecture provides a promising direction for dense prediction tasks.
Unlike conventional segmentation networks that heavily rely on task-specific supervision, DINOv3 learns representations through self-supervised learning on massive unlabeled natural images and extracts rich, high-dimensional dense feature representations, showing robust generalization under illumination variations, scale changes, and unseen-domain data\citep{simeoniDINOv32025}.

Recent studies have begun to introduce the dense prediction capability of DINOv3 into medical image segmentation and have shown encouraging progress in several related research directions.
For example, \citet{liuDoes2025} conducted a comprehensive benchmark evaluation of the zero-shot generalization capability of DINOv3 on multimodal medical images, showing that its robust visual features outperformed medical-domain foundation models such as BiomedCLIP in several segmentation tasks.
\citet{heIncentivizing2025} proposed DINOv3-FD, a feature-disentanglement-based adaptation framework that decouples task-relevant and task-irrelevant feature subspaces through a dual-stream adapter, thereby helping bridge the domain gap between natural and medical images.
\citet{xuExploiting2026} further proposed the DINO-AugSeg framework, which combines the stable self-supervised representations of DINOv3 with wavelet-domain augmentation to address few-shot segmentation under limited clinical annotations.
Collectively, these studies provide useful references for baseline model selection and methodological refinement in this work.

\subsection{Mixture-of-Experts Architecture}

Mixture-of-Experts (MoE) is a dynamically parameterized network architecture for modeling complex tasks. Its core idea is to adaptively select or weight multiple expert subnetworks according to input features through a gating router\citep{jacobsAdaptive1991}.
Unlike conventional shared-parameter networks that use a uniform computational pathway for all samples and categories, MoE routes different inputs through differentiated feature-processing paths.
This conditional computation expands the representational capacity of the model without substantially increasing per-inference computational overhead.
This mechanism enables more flexible feature disentanglement and parameter utilization under complex data distributions, multi-task learning, and highly heterogeneous scenarios.

Due to its dynamic routing and conditional computation mechanisms, MoE shows flexible adaptive modeling capability when handling cross-domain, cross-scenario, and highly heterogeneous data distributions\citep{liSparse2022}.
This mechanism is well suited to capturing data heterogeneity jointly caused by imaging devices, acquisition environments, lesion morphologies, and noise interference.
For example, \citet{chenAdaMVMoE2023} proposed AdaMV-MoE and showed that multi-expert conditional routing can effectively mitigate gradient interference among network parameters during the joint optimization of highly heterogeneous multi-task data.

Conventional medical image segmentation models often rely on static weighting strategies and treat pixels in a largely uniform manner, limiting their adaptability to morphologically diverse pathological scleral features and severe environmental noise.
In contrast, the MoE architecture has inherent advantages in dynamic feature disentanglement, which can improve generalization and robustness against noise and acquisition artifacts.

\subsection{Specular Reflection Artifact Interference}

Specular reflections, overexposed regions, and saturated areas in medical images are prevalent forms of imaging artifacts that can compromise automated image analysis.
\citet{aliDeep2021} proposed a deep learning framework for endoscopic video quality assessment and image restoration, showing that artifacts such as motion blur, bubbles, specular reflections, floating objects, and pixel saturation can affect clinical visual interpretation and automated analysis. The framework further includes modules for artifact detection, instance segmentation, quality scoring, and local restoration.
Existing methods for handling specular reflection artifacts mostly follow a pipeline of detection followed by removal or restoration.
\citet{DAHER2023102994} proposed a temporal learning-based specularity inpainting method, which uses temporal information from adjacent frames to infer the tissue appearance beneath specular-reflection-occluded regions, and evaluated the influence of specular reflection restoration on downstream tasks such as optical flow estimation, feature matching, and camera motion estimation.
\citet{s23020974} proposed an endoscopic specular reflection detection and removal method based on luminance classification, which localizes high-intensity reflection regions and performs subsequent restoration to reduce the influence of specular reflections on medical image analysis.
Beyond explicit image restoration, some medical segmentation methods attempt to enhance feature discriminability under complex backgrounds by constructing contrastive constraints or hard negative samples. For example, ConDSeg improves the discriminative ability between target regions and interfering regions through conditional contrastive learning\citep{leiConDSeg2025}.

However, the above methods are mainly designed for endoscopic image restoration, image quality enhancement, or general complex-background interference modeling. Their objectives are typically to restore visual content occluded by specular reflections or to improve the overall discrimination between target and background regions.
In contrast, in the multi-label scleral anomaly segmentation task considered in this study, specular reflection regions and true anomaly regions often exhibit spatial proximity or partial occlusion. When preserving the topological connectivity of predicted masks, conventional segmentation models may incorrectly absorb specular reflection regions into anomaly masks, resulting in segmentation leakage.

Different from explicit image restoration pipelines, PCP Loss focuses on the implicit suppression of high-confidence false-positive predictions within specular reflection regions. 
Specifically, scleral specular reflection regions are treated as special negative-sample regions, and progressive confidence-based constraints are imposed on anomaly predictions within these regions. 
This design aims to reduce the influence of specular reflection artifacts on anomaly boundary localization while mitigating boundary expansion caused by segmentation leakage.

\section{Datasets}
\label{datasets}

Although existing datasets, especially SBVPI, provide a solid foundation for basic segmentation of the sclera, they fall short of meeting the needs of this study for more in-depth clinical auxiliary diagnostic analysis, particularly for multi-class anomaly analysis based on TCM ocular inspection theory.
Their annotations primarily cover macro-anatomical categories such as canthus, eyelashes, iris, periocular region, pupil, and overall sclera, while the only category closely aligned with the scope of this study is the Vessels (Ve) subset in the SBVPI dataset.
Consequently, these datasets still lack dense semantic annotations for multi-class composite anomalies on the scleral surface, such as heterogeneous manifestations of Yellow and Black Spots (YBS) and pathological Blood Spots (BS).

In addition, real-world anomaly patterns often exhibit spatial overlap and topological ambiguity. For example, dense Ve patterns may be intertwined with BS features, making conventional mutually exclusive single-label mask annotation schemes insufficient for practical application needs.
To address this annotation limitation, this study constructs and introduces ML-SASD for complex acquisition scenarios. The structure of the dataset is summarized in Table~\ref{tab:Datasets Structure}.

\begin{table}[pos=htbp,width=\textwidth]
  \centering
  \caption{ML-SASD Dataset Structure}
  \label{tab:Datasets Structure}
  \resizebox{\textwidth}{!}{%
    \begin{tabular}{cccccc}
    \toprule
    Subset Name & Environment & Total & First-stage Annotated & Second-stage Annotated & Negative Labels \\
    \midrule
    ML-SASD-Clinical & Controlled / Ideal & 1075  & 1075  & 265   & 265 \\
    ML-SASD-Wild & Unconstrained / Glare & 1068  & 250   & 250   & 250 \\
    ML-SASD-Mix & Diverse & 2143  & 1325  & 515   & 515 \\
    \bottomrule
    \end{tabular}%
  }
\end{table}

\subsection{Dataset Acquisition}

\begin{figure}[pos=htbp,width=\textwidth]
    \centering

    \begin{minipage}[t]{0.31\textwidth}
        \centering
        \includegraphics[width=\linewidth]{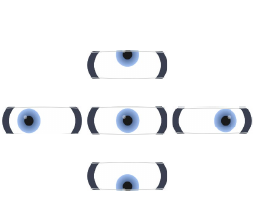}
        \vspace{2pt}
        \textbf{(a)}
    \end{minipage}
    \hfill
    \begin{minipage}[t]{0.31\textwidth}
        \centering
        \includegraphics[width=\linewidth]{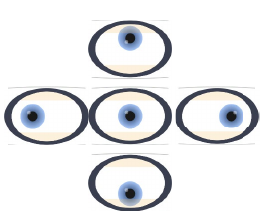}
        \vspace{2pt}
        \textbf{(b)}
    \end{minipage}
    \hfill
    \begin{minipage}[t]{0.31\textwidth}
        \centering
        \includegraphics[width=\linewidth]{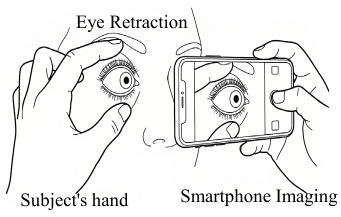}
        \vspace{2pt}
        \textbf{(c)}
    \end{minipage}

    \caption{Illustration of the eye image acquisition protocol. 
    (a) and (b) show the Five-eye Positioning Method for exposing different scleral regions, and (c) shows the schematic illustration of eye image acquisition.}
    \label{fig:eye-acquisition-protocol}
\end{figure}

To improve the coverage of complex acquisition conditions in ML-SASD, we constructed a multi-source data acquisition framework consisting of medical atlases and images captured using commercial mobile devices, thereby providing a data basis for evaluating cross-scenario model generalization.

\begin{figure}[pos=!t,width=\textwidth]
	\centering
	\includegraphics[width=0.8\textwidth]{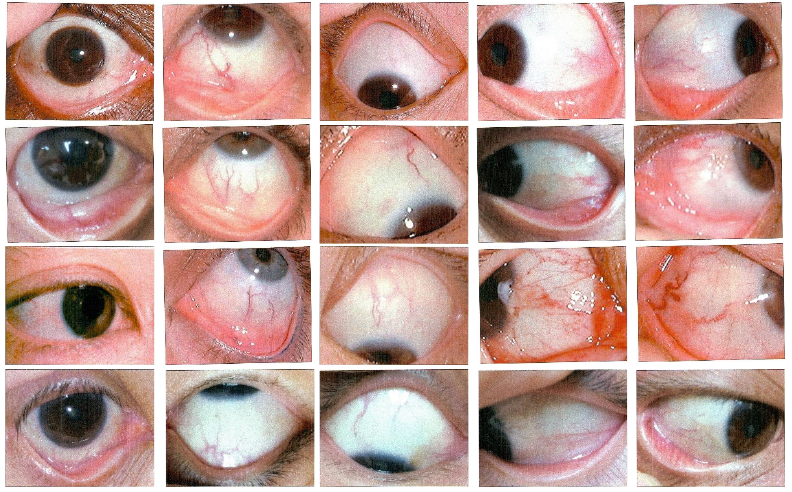}
	\caption{Illustrations from \textit{Traditional Chinese Medicine Eye Diagnosis} (3rd Edition) \citep{zheng_traditional_eye_diagnosis}}
	\label{figs/Illustrations-book}
\end{figure}

\begin{figure}[pos=!t,width=\textwidth]
	\centering
	\includegraphics[width=0.8\textwidth]{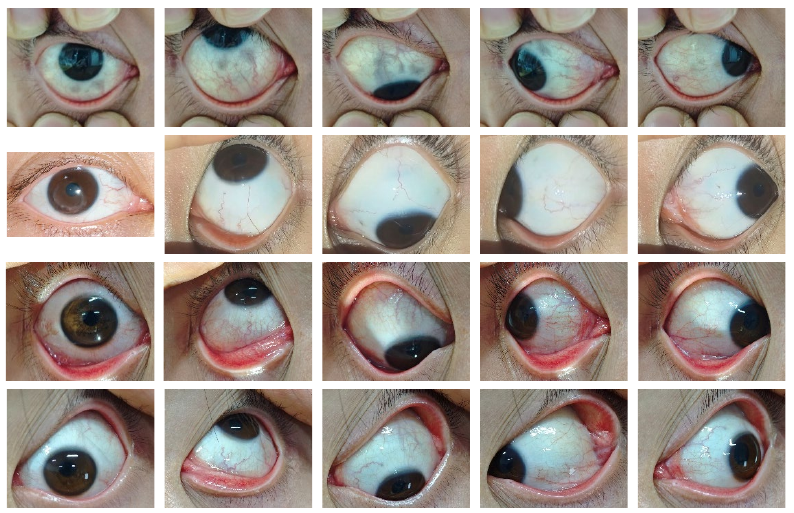}
	\caption{Representative image examples from the ML-SASD-Wild subset}
	\label{figs/ML-SASD-Wild}
\end{figure}

\textbf{(1) ML-SASD-Clinical (controlled clinical subset):}
This subset was selected from the medical atlas \textit{Illustrated TCM Ocular Inspection and Syndrome Differentiation}\citep{zheng_traditional_eye_diagnosis}, which consists of ocular images with diverse morphologies captured by Dr. Zheng using a digital camera during years of clinical practice.
The acquisition protocol follows the Five-eye Positioning Method, as shown in Fig.~\ref{fig:eye-acquisition-protocol}(b), where subjects were photographed under five gaze directions, namely forward, upward, downward, leftward, and rightward, to cover different regions of the ocular surface.

As shown in Fig.~\ref{fig:eye-acquisition-protocol}(a), when ocular images are captured under natural, unassisted eye-opening conditions, the eyelids often occlude the superior and inferior scleral regions, which may lead to information loss during subsequent anomaly localization.
To obtain more complete scleral anomaly features, Dr. Zheng incorporated a manually assisted eyelid-retraction step into the imaging workflow.
As shown in Fig.~\ref{fig:eye-acquisition-protocol}(b), while the subject rotated the eyes, the operator gently retracted the eyelids superiorly and inferiorly through slight manual traction, effectively exposing the regions originally hidden in the conjunctival fornix (the hidden regions are marked in pale yellow). Fig.~\ref{fig:eye-acquisition-protocol}(c) shows a schematic illustration of the ocular imaging workflow.
This manually assisted step substantially increased the effective visible area of the sclera and provided a reliable basis for subsequent high-precision segmentation.

After organization and screening, ML-SASD-Clinical contains 1,075 high-resolution images, with representative examples shown in Fig.~\ref{figs/Illustrations-book}.

\textbf{(2) ML-SASD-Wild:}
To simulate unconstrained daily environments, we used multiple mobile imaging devices, including smartphones, tablets, and cameras, to capture images under varied illumination conditions, including natural light and indoor artificial lighting.
The data acquisition protocol was consistent with that of the ML-SASD-Clinical subset.
The subjects in this subset ranged in age from 18 to 60 years, ensuring broad diversity in ocular biometric characteristics and anomaly-related manifestations.
Notably, during sample recruitment, we intentionally included a large number of students and researchers who had long been engaged in laboratory work.
Due to prolonged high-intensity visual load and irregular daily routines, this population may be more likely to show ocular manifestations associated with potential disease-related or subclinical conditions, thereby improving the anomaly diversity of this dataset.
Finally, we collected 1,068 images with varying levels of background noise and resolution differences, forming the ML-SASD-Wild subset. Representative examples of the ML-SASD-Wild subset are shown in Fig.~\ref{figs/ML-SASD-Wild}.

\subsection{Data Annotation Rules}

\begin{figure*}[pos=htbp,width=\textwidth]
    \centering

    \begin{subfigure}[t]{0.48\textwidth}
        \centering
        \includegraphics[width=0.85\linewidth]{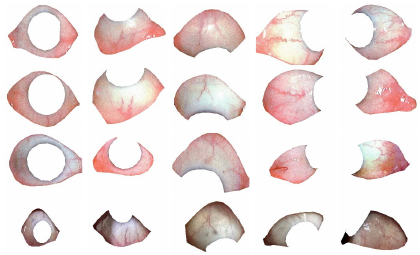}
        \caption{Representative annotated sclera examples from the ML-SASD-Clinical subset.}
        \label{fig:ML-SASD-Clinical-sclera}
    \end{subfigure}
    \hfill
    \begin{subfigure}[t]{0.48\textwidth}
        \centering
        \includegraphics[width=0.85\linewidth]{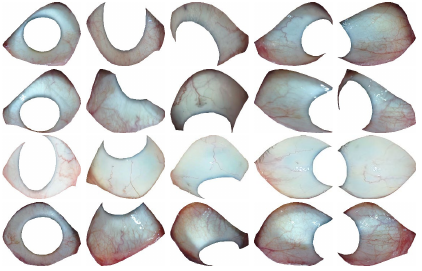}
        \caption{Representative annotated sclera examples from the ML-SASD-Wild subset.}
        \label{fig:ML-SASD-Wild-sclera}
    \end{subfigure}

    \caption{Representative annotated sclera examples from the ML-SASD dataset (First-stage Annotated)}
    \label{fig:ML-SASD-sclera}
\end{figure*}

\begin{figure}[pos=!t,width=\textwidth]
	\centering
	\begin{minipage}[t]{0.48\textwidth}
		\centering
		\includegraphics[width=\linewidth]{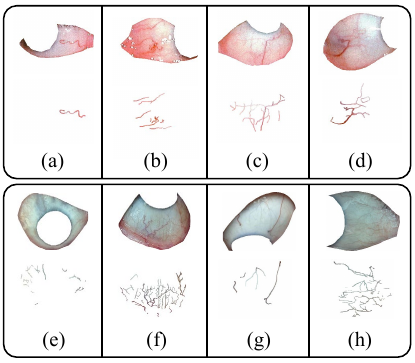}
		\par\vspace{2pt}
		\small\textbf{Left:} Examples of Ve
	\end{minipage}
	\hfill
	\begin{minipage}[t]{0.48\textwidth}
		\centering
		\includegraphics[width=\linewidth]{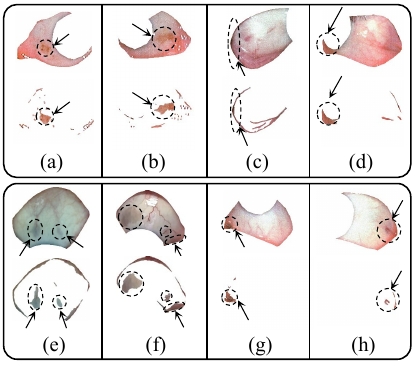}
		\par\vspace{2pt}
		\small\textbf{Right:} Examples of YBS
	\end{minipage}
	\caption{Representative examples of Ve and YBS from the Second-stage Annotated subset. Left: various examples of Ve. Right: various examples of YBS. The black dashed contours and arrows in the YBS examples are used only to highlight typical locations and morphological characteristics and do not represent precise annotation boundaries.}
	\label{fig:examples-Ve-YBS}
	\label{fig:examples-Ve}
	\label{fig:examples-YBS}
\end{figure}

\begin{figure}[pos=!t,width=\textwidth]
	\centering
	\includegraphics[width=0.45\textwidth]{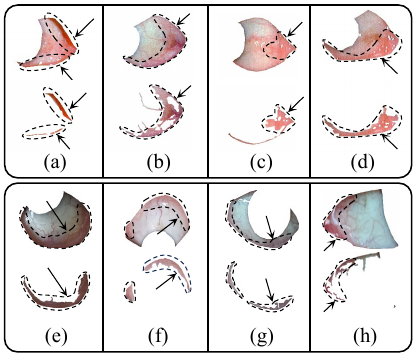}
	\caption{Various examples of BS (Second-stage Annotated): (a) marginal bright-red clot-like BS; (b) reticular congestion with marginal hemorrhage; (c) localized congestion in the perilimbal conjunctiva; (d) diffuse marginal congestion; (e) large patch-like diffuse dark-red BS; (f)(g) marginal light-red congestion; (h) localized congestion in the perilimbal conjunctiva obscured by specular reflection. The black dashed contours and arrows are used only to highlight the typical locations and morphological characteristics of BS and do not represent precise annotation boundaries.}
	\label{fig:examples-BS}
\end{figure}

The annotation protocols of ML-SASD were designed in accordance with the cascaded semantic segmentation strategy in the TAO system, resulting in two annotation sets: First-stage Annotated Images and Second-stage Annotated Images.
For First-stage Annotated Images, the task objective is to segment the sclera from ocular images; therefore, the target semantic classes include Periocular, Iris, and Sclera.
Since this study mainly focuses on the scleral region, the iris was not further subdivided into the pupil category as in the SBVPI dataset\citep{vitekComprehensive2020}.
The first-stage annotation set contains 1,325 images.
Fig.~\ref{fig:ML-SASD-Clinical-sclera} and Fig.~\ref{fig:ML-SASD-Wild-sclera} show representative examples of the annotated Sclera masks from ML-SASD-Clinical and ML-SASD-Wild.

For Second-stage Annotated Images, the task objective is fine-grained segmentation of scleral surface anomalies. Therefore, with reference to the major anomaly categories documented in \textit{Traditional Chinese Medicine Eye Diagnosis} (Third Edition)\citep{zheng_traditional_eye_diagnosis}, we defined three distinct and representative anomaly classes for dense annotation: Vessels (Ve), Yellow and Black Spots (YBS), and Blood Spots (BS).
Morphological and visual variants of Ve, YBS, and BS are illustrated in Fig.~\ref{fig:examples-Ve}(left), Fig.~\ref{fig:examples-YBS}(right), and Fig.~\ref{fig:examples-BS}, respectively.

\begin{figure}[pos=htbp,width=\textwidth]
	\centering
	\begin{minipage}[t]{0.48\textwidth}
		\centering
		\includegraphics[width=\linewidth]{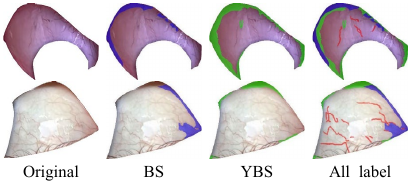}
		\par\vspace{2pt}
		\small\textbf{(a)} BS--YBS overlap
	\end{minipage}
	\hfill
	\begin{minipage}[t]{0.48\textwidth}
		\centering
		\includegraphics[width=\linewidth]{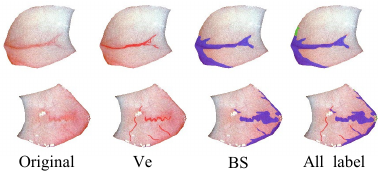}
		\par\vspace{2pt}
		\small\textbf{(b)} Ve--BS overlap
	\end{minipage}
	\caption{Examples of regional overlap among scleral anomaly categories (Second-stage Annotated). 
	(a) Regional overlap between BS and YBS, where blue and green highlighted regions indicate the BS and YBS annotations, respectively. 
	(b) Regional overlap between Ve and BS, where red and blue highlighted regions indicate the Ve and BS annotations, respectively. 
	In both panels, ``All label'' shows the combined visualization of all three categories.}
	\label{fig:between-category-overlap}
	\label{fig:between-BS-YBS}
	\label{fig:between-BS-Ve}
\end{figure}

It is worth noting that these three anomaly categories exhibit visual similarity and morphological overlap. 
Specifically, YBS and BS may show chromatic overlap; when BS appears markedly dark red, its color-texture characteristics may visually resemble those of YBS, as shown in Fig.~\ref{fig:examples-YBS}(c)(d)(g)(h) and Fig.~\ref{fig:between-BS-YBS}(a). 
Similarly, when Ve patterns are highly dense or prominent, their local geometric structures may resemble the morphological profiles of BS, as shown in Fig.~\ref{fig:examples-BS}(b)(c)(e)(f)(g)(h) and Fig.~\ref{fig:between-BS-Ve}(b).

\begin{figure}[pos=htbp,width=\textwidth]
	\centering
	\includegraphics[width=0.9\textwidth]{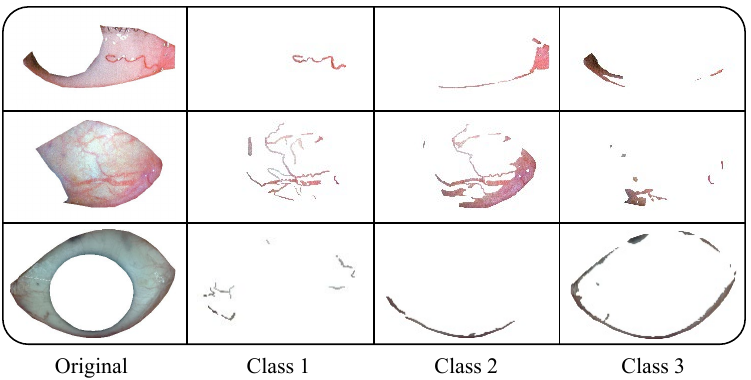}
	\caption{Comprehensive visualization of scleral anomaly annotations in ML-SASD-Mix (Second-stage Annotated): the ``Original'' panel shows the raw input image, while ``Class 1'', ``Class 2'', and ``Class 3'' correspond to the ground-truth masks of Ve, BS, and YBS, respectively.}
	\label{fig:ML-SASD-Mix-sclera}
\end{figure}

Overall, fine-grained annotation of many anomaly regions may involve semantic and topological ambiguity.
Conventional semantic segmentation networks usually adopt mutually exclusive pixel-level assignment, where each pixel is assigned to a single category.
Therefore, considering the inter-class overlap described above, the target model should be designed to allow a single pixel to be associated with multiple anomaly categories simultaneously, resulting in composite pixel-wise labeling.
Fig.~\ref{fig:ML-SASD-Mix-sclera} provides a comprehensive visualization of multi-label scleral anomaly masks from ML-SASD-Clinical and ML-SASD-Wild.

To improve annotation consistency, all masks in ML-SASD were produced following a unified annotation protocol. 
The annotations were independently prepared by trained annotators and reviewed by researchers with experience in ocular image analysis.
Uncertain cases were further reviewed with reference to expert feedback.
The annotation process first defined category-specific visual criteria according to the morphological descriptions and representative examples of Ve, YBS, and BS. 
The initial annotations were then checked and corrected through a multi-round review procedure. 
For regions with ambiguous boundaries, the annotators were required to follow the visible edge of the anomaly as closely as possible, while avoiding speculative expansion into unclear background areas. 
For spatially overlapping anomaly patterns, the multi-label annotation rule was adopted, allowing the same pixel or local region to be assigned to more than one anomaly category when supported by visual evidence. 
Disagreements or uncertain cases were resolved through consensus review based on the original image, category definitions, and adjacent contextual structures. 
This quality-control procedure was used to reduce annotation inconsistency caused by weak boundaries, chromatic similarity, and topological overlap among scleral surface anomalies.

\subsection{Scleral Specular Reflection Negative Samples}

\begin{figure}[pos=H,width=\textwidth]
	\centering
	\includegraphics[width=0.5\textwidth]{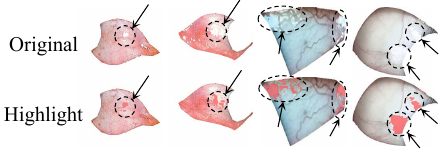}
	\caption{Examples of SSR (Second-stage Annotated): The first column shows the original images, and the second column highlights the SSR regions on the original images. The black dashed contours and arrows are used only to highlight the typical locations and morphological characteristics of SSR and do not represent precise annotation boundaries.}
	\label{fig:examples-SSR}
\end{figure}

As discussed in Section~\ref{Introduction}, specular reflection on the scleral surface is an important interference factor faced by the TAO system in real-world acquisition scenarios.
To explicitly characterize this issue, based on the annotation protocols of Second-stage Annotated Images, this study further constructs the \textbf{S}cleral \textbf{S}pecular \textbf{R}eflection Negative Sample Dataset (SSR).
In this annotation set, SSR regions are labeled as special negative-sample regions to describe their spatial distributions and morphological patterns in real images, as shown in Fig.~\ref{fig:examples-SSR}.
This resource is mainly used to provide a quantitative basis for validation experiments and to measure the extent of false-positive predictions produced by the model within SSR regions.

\subsection{Dataset Distribution Analysis}

To systematically characterize the challenges of ML-SASD and provide an empirical basis for the subsequent design of HD-DinoMoE, this study conducts statistical analyses from two perspectives: basic physical indicators and high-level semantic associations.
For consistency in statistical profiling, SSR is included in the statistics together with the three positive-sample categories.

\subsubsection{Dataset Partitioning and Image-Level Sample Distribution}

\begin{figure}[pos=htbp,width=\textwidth]
	\centering
	\includegraphics[width=0.5\textwidth]{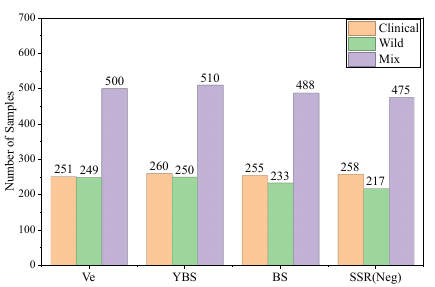}
	\caption{Image-level sample distribution statistics}
	\label{fig:Image-level-sample}
\end{figure}

Fig.~\ref{fig:Image-level-sample} summarizes the image-level frequency of each semantic class in the dataset.
Both the ML-SASD-Clinical and ML-SASD-Wild subsets maintain an approximately 8:1:1 partitioning ratio among the training, validation, and test splits.
Because the data acquisition scope intentionally included participants who may be more likely to show potential disease-related or subclinical ocular manifestations, the anomaly categories show high image-level coverage, with a relatively balanced sample distribution across categories.

\begin{figure}[pos=htbp,width=\textwidth]
	\centering
	\includegraphics[width=\textwidth]{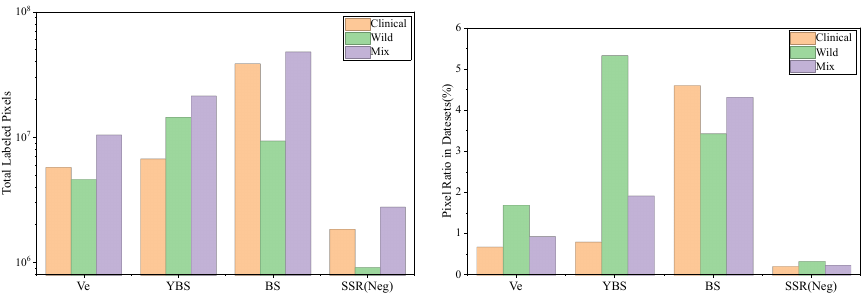}
	\caption{Pixel-level class distribution across Clinical-Wild-Mix datasets:The left chart displays the absolute number of labeled pixels on a logarithmic scale for each segmentation class. The right chart illustrates the relative pixel ratio (\%) of each class within the respective datasets.}
	\label{fig:Pixel-level}
\end{figure}

Although the image-level distribution of each class is relatively balanced, the pixel-level indicators reveal severe class imbalance.
In the Pixel Ratio chart in the right panel of Fig.~\ref{fig:Pixel-level}, all foreground classes together account for less than 10\% of the image pixels. For example, even the class with the highest pixel ratio accounts for only approximately 5\% in the ML-SASD-Wild subset, while most categories fall between 1\% and 4\%. More than 90\% of the pixels belong to the background.
Without targeted intervention during training, the model may suffer from background-biased prediction collapse.
Within the foreground classes, the pixel-level distribution is also highly skewed. In the Total Labeled Pixels chart in the left panel of Fig.~\ref{fig:Pixel-level}, BS and YBS contain the largest numbers of labeled pixels, approaching or exceeding the order of $ (10^7) $.
The pixel count and pixel ratio of SSR are both extremely low. However, substantial differences in class distribution exist across different data sources. In the ML-SASD-Clinical subset, BS is dominant, with a pixel ratio exceeding 4\%, whereas YBS is rare.
In the ML-SASD-Wild subset, YBS becomes dominant, with a pixel ratio exceeding 5\%, whereas the proportion of BS is relatively reduced.

\subsubsection{Resolution Discrepancy and the Small Object Characteristic}
\label{Small-Object}
\begin{figure}[pos=htbp,width=\textwidth]
	\centering
	\includegraphics[width=0.5\textwidth]{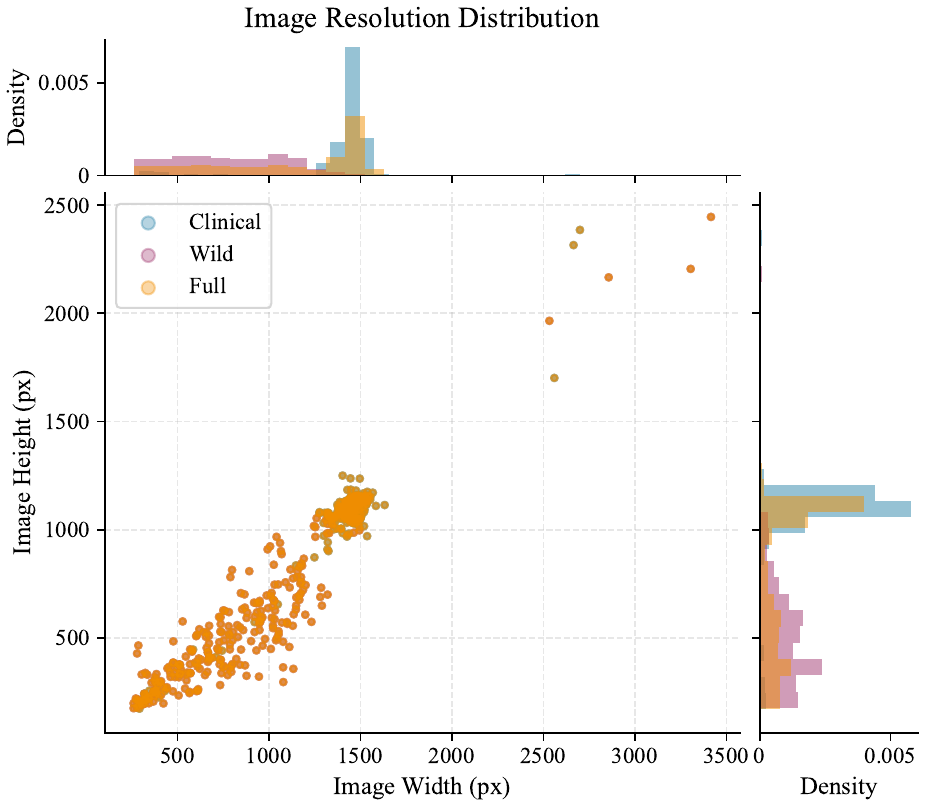}
	\caption{Image Resolution Distribution}
	\label{fig:image-size}
\end{figure}

As shown in Fig.~\ref{fig:image-size}, a substantial resolution discrepancy exists between the two subsets.
The ML-SASD-Clinical subset exhibits a relatively homogeneous and standardized distribution, with its data points concentrated within a narrow resolution band, approximately 1400--1500 px in width and 1000--1200 px in height.
In contrast, the ML-SASD-Wild subset shows a much broader resolution distribution, with data points widely scattered along the coordinate axes, ranging from less than 500 px to approximately 3500 px in width.
This pattern is consistent with real-world unconstrained acquisition scenarios, where images may be captured using different mobile devices and at varying capture distances.
These resolution differences indicate that the model should be able to handle substantial scale variations, namely scale invariance.

\begin{figure}[pos=htbp,width=\textwidth]
	\centering
	\includegraphics[width=0.5\textwidth]{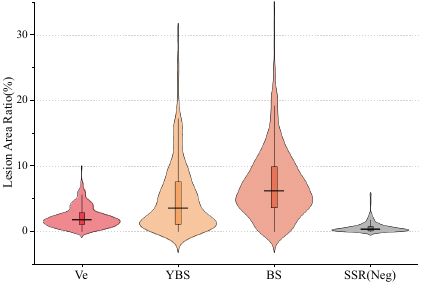}
	\caption{Per-Class Lesion Area Distribution}
	\label{fig:Lesion}
\end{figure}

As shown in Fig.~\ref{fig:Lesion}, the Lesion Area Ratio measures the pixel-wise area proportion of each category relative to the whole image. The maximum area ratio reaches only 35\%, while the bulk of the distribution remains between 0\% and 5\%. In real-world images, the spatial footprint of most anomaly regions may even fall below 1\%, indicating that these regions exhibit clear small-object characteristics. This poses a challenge for preserving fine-grained local features during segmentation.

\subsubsection{Semantic Concurrency and Spatial Priors}

\begin{figure}[pos=htbp,width=\textwidth]
	\centering
	\includegraphics[width=\textwidth]{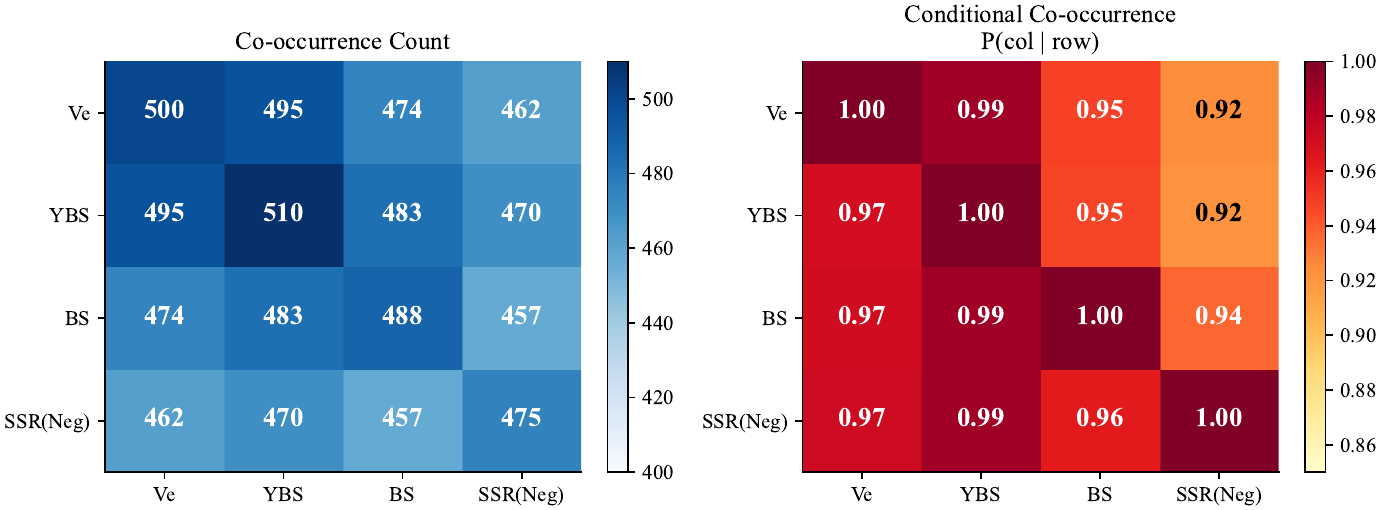}
	\caption{Class co-occurrence matrix}
	\label{fig:co-occurrence}
\end{figure}

As summarized in Fig.~\ref{fig:co-occurrence}, the class co-occurrence matrix of ML-SASD shows strong semantic concurrency among anomaly categories. The conditional probability matrix indicates that all categories exhibit high co-occurrence probabilities.
For example, the conditional probability of Ve and YBS appearing simultaneously is close to 1.00, suggesting that these two categories are highly coupled rather than isolated.
Further analysis suggests that this high-frequency concurrency is closely related to the sampling preference of this dataset. Specifically, the data acquisition mainly focused on participants who may be more likely to show potential disease-related or subclinical ocular manifestations, and such participants may also be more likely to exhibit multiple anomaly categories simultaneously.

\begin{figure}[pos=htbp,width=\textwidth]
	\centering
	\includegraphics[width=0.5\textwidth]{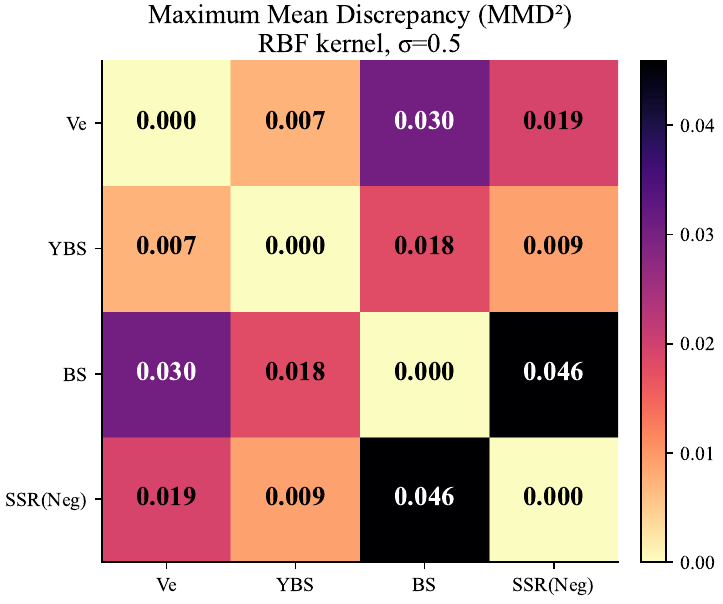}
	\caption{Maximum Mean Discrepancy matrix}
	\label{fig:mmd-matrix}
\end{figure}

Fig.~\ref{fig:mmd-matrix} presents the Maximum Mean Discrepancy (MMD) matrix, which quantifies class-specific differences in spatial distribution. The highest global discrepancy value (0.046) appears at the intersection between BS and SSR, indicating strong spatial exclusiveness between the two categories.
As observed in Fig.~\ref{fig:examples-SSR}, SSR regions are usually concentrated in the central scleral region, whereas BS more frequently occurs in the peripheral scleral region.
This statistical distribution pattern is consistent with the optical characteristics of real-world ocular image acquisition.
In contrast, the lowest discrepancy value (0.007) in the matrix appears between YBS and Ve, indicating a high degree of spatial co-occurrence in their topological distributions.
This data-driven finding is supported by the visual patterns in Fig.~\ref{fig:examples-YBS}(e), (f), and (h), where many YBS regions appear at Ve terminals.
This highly overlapping spatial distribution further suggests a potential association between these two anomaly categories.

\section{Methodology}
\subsection{System-Level Overview of the TAO Framework}

\begin{figure}[pos=htbp,width=\textwidth]
	\centering
	\includegraphics[width=\textwidth]{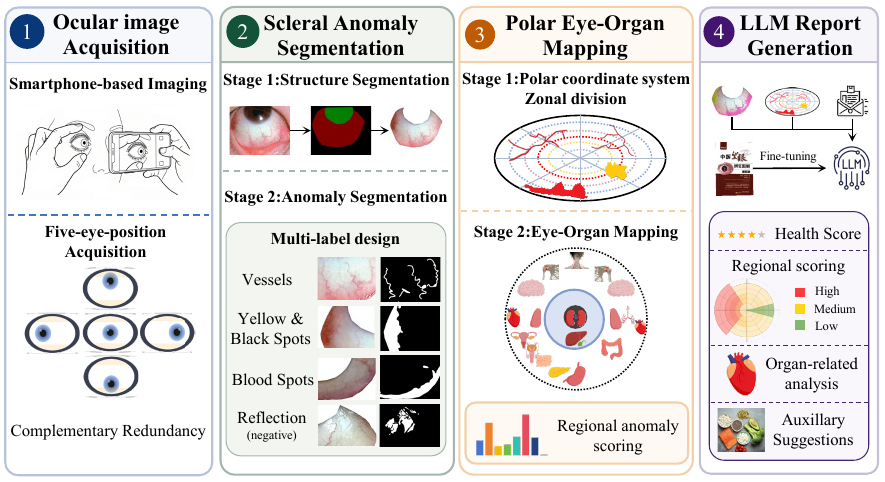}
	\caption{Technical architecture of the TCM-inspired AI Ocular Auxiliary Diagnosis System}
	\label{fig:TAO-4}
\end{figure}

To support the systematic transition of TCM ocular inspection from empirical observation toward intelligent and quantifiable analysis, this study introduces the \textbf{T}raditional Chinese Medicine-inspired \textbf{A}rtificial Intelligence \textbf{O}cular Auxiliary Diagnosis System (TAO) as a system-level technical framework for scleral surface anomaly analysis.
The system is not intended to replace clinicians in diagnostic decision-making. Instead, it aims to provide objective, visually grounded, and traceable auxiliary analytical evidence through standardized ocular image acquisition, pixel-level scleral anomaly segmentation, ocular region mapping, and knowledge-enhanced report generation.
As shown in Fig.~\ref{fig:TAO-4}, the overall workflow is organized into four components: ocular image acquisition, scleral anomaly segmentation, polar-coordinate eye-organ mapping and regional scoring, and knowledge-enhanced report generation.

In the first component, TAO captures multi-view ocular images by guiding subjects through different gaze directions, thereby expanding the visible scleral area as much as possible and reducing information loss caused by a single viewpoint.
In the current system design, the acquisition module guides subjects to complete image capture in five gaze directions, namely upward, leftward, central, rightward, and downward, through orbital-region detection, pupil detection, and position control, thereby forming multi-view inputs that cover different scleral regions.
For mobile or unconstrained scenarios, users can independently capture ocular images using mobile devices. This design provides the image-input basis for subsequent scleral region extraction and anomaly analysis.

In the second component, TAO adopts a coarse-to-fine cascaded semantic segmentation strategy to parse ocular image structures.
The initial phase extracts the scleral region from the raw ocular image by filtering out non-target anatomical regions such as the iris and periocular background; the subsequent phase further identifies scleral surface anomalies, including Ve, YBS, and BS, within the extracted scleral region.
Since these anomaly categories may involve morphological differences, spatial overlap, and occlusion caused by SSR, the accuracy of the subsequent-phase segmentation is critical to the reliability of downstream regional scoring and auxiliary analysis.
Therefore, this study focuses on the second-phase scleral anomaly segmentation task, and HD-DinoMoE is designed around this task.

In the third component, TAO establishes a spatial mapping relationship based on the segmented scleral anomaly regions, using the pupil center as the coordinate origin.
Specifically, the system maps anomaly regions under different gaze directions into a standardized ocular coordinate space and calculates the anomaly distribution across different ocular regions based on the radial coordinate, polar angle, and regional partitioning results.
By incorporating the eye-organ mapping concept in TCM ocular inspection, the system further converts the directional aggregation of scleral anomalies into regional anomaly scores or heatmaps, which are used to indicate ocular regions that may require closer attention.
The role of this module is to transform pixel-level segmentation results into region-level visual information that is more consistent with clinical observation workflows.

In the fourth component, TAO further introduces a knowledge-enhanced report generation module.
This module takes ocular images, scleral anomaly segmentation results, regional scoring results, and users' basic information as structured inputs, and incorporates literature related to TCM ocular inspection and domain-specific medical knowledge bases through Retrieval-Augmented Generation (RAG).
Specifically, the system retrieves relevant knowledge from digitized TCM textual corpora, ocular inspection diagnostic materials, and physician-annotated cases, and then generates an auxiliary analytical report using a multimodal large model.
This report is intended only as a reference for physicians and should not be used as an independent diagnostic conclusion.

This study primarily focuses on the empirical investigation and validation of the scleral anomaly segmentation component of the TAO system, including the construction of the ML-SASD dataset and the design of the HD-DinoMoE model.
The polar-coordinate eye-organ mapping, regional anomaly scoring, and knowledge-enhanced report generation modules are articulated in this work as a conceptual, system-level downstream application framework.
Their clinical validity, diagnostic consistency, and real-world deployment performance remain outside the immediate scope of the present methodological validation and should be further assessed in future studies through physician evaluation and multicenter data.

Through this design, this study places the pixel-level scleral anomaly segmentation task within a complete auxiliary diagnostic system workflow. Accordingly, the model optimization objective is not limited to improving dataset-level metrics, but is further aligned with the practical requirements of downstream regional localization, anomaly scoring, and knowledge-enhanced report generation in the TAO system.

\subsection{Overall Architecture of HD-DinoMoE}

\begin{figure}[pos=H,width=\textwidth]
	\centering
	\includegraphics[width=0.75\textwidth]{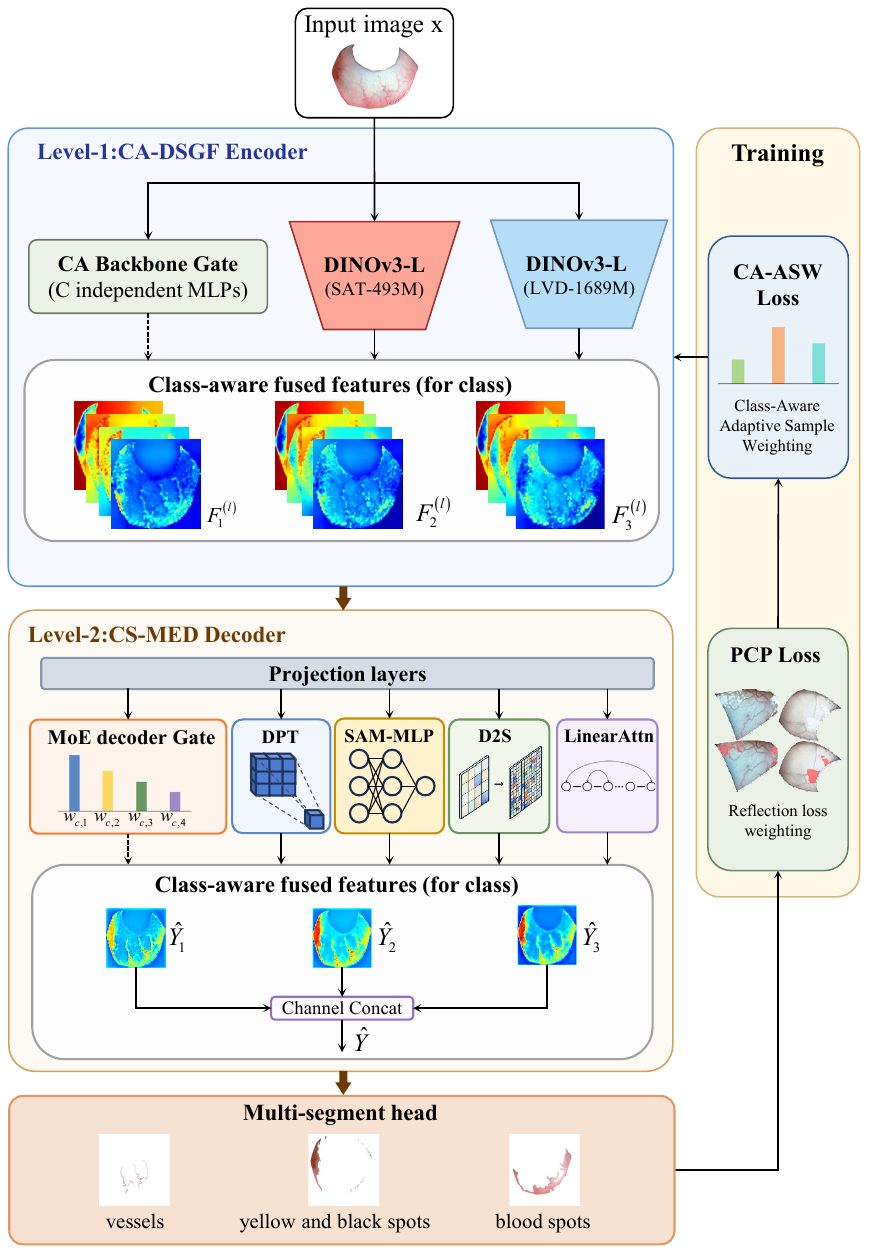}
	\caption{HD-DinoMoE Overall architecture}
	\label{fig:HD-DinoMoE-Overall-architecture}
\end{figure}

HD-DinoMoE is designed for the second-phase scleral surface anomaly segmentation task in the TAO system.
Given a pre-extracted scleral region of interest (ROI), the model predicts three anomaly categories, namely Ve, YBS, and BS, in a multi-label manner.
As shown in Fig.~\ref{fig:HD-DinoMoE-Overall-architecture}, HD-DinoMoE follows a class-aware hierarchical dual design, consisting of CA-DSGF for feature encoding, CS-MED for pixel decoding, and two training-level optimization components, PCP Loss and CA-ASW.

The core design principle of HD-DinoMoE is challenge-aware hierarchical dynamic adaptation. 
Specifically, the model introduces tailored adaptive mechanisms across three complementary levels: feature encoding, pixel decoding, and training optimization.

Architecturally, the Hierarchical Dual structure serves as the central design principle of HD-DinoMoE.
Specifically, the model implements a two-level adaptive design across feature encoding and pixel decoding.

At Level-1, HD-DinoMoE introduces CA-DSGF as its feature-encoding module.
CA-DSGF adopts a dual-stream parallel topology composed of two DINOv3-L visual foundation branches initialized with heterogeneous pretrained weights, namely the SAT branch and the LVD branch.
Through the Class-Aware Backbone Gate (CA Backbone Gate), the model dynamically adjusts the fusion weights of the two branches for each anomaly category, thereby improving representational robustness under multi-source domain heterogeneity.

At Level-2, HD-DinoMoE deploys CS-MED for class-specific pixel decoding.
This module equips each anomaly category with a heterogeneous expert pool composed of DPT, SAM-MLP, D2S, and LinearAttn.
Driven by the gated routing mechanism, CS-MED adaptively weights category-specific decoding paths according to the structural and morphological heterogeneity of individual anomaly categories.

During training, HD-DinoMoE incorporates two adaptive optimization mechanisms to improve training stability and cross-scenario robustness.

PCP Loss treats SSR regions as special negative-sample regions and penalizes high-confidence false-positive anomaly predictions, thereby reducing reflection-induced segmentation leakage near anomaly boundaries.

CA-ASW further adjusts loss weights along sample-wise and class-wise dimensions, enabling the model to balance heterogeneous training contributions across samples and anomaly categories.

\subsection{Class-Aware Dual-Stream Gated Fusion Encoder}
\label{CA-DSGF}

The three scleral surface anomaly categories exhibit distinct representational requirements: Ve emphasizes elongated topological structures, YBS involves low-frequency pigmentation or plaque-like texture changes, and BS often shows irregular high-contrast regions.
A uniform single-encoder representation may therefore be insufficient to capture category-specific structural and textural cues under heterogeneous acquisition conditions.

\begin{figure}[pos=htbp,width=\textwidth]
	\centering
	\includegraphics[width=0.75\textwidth]{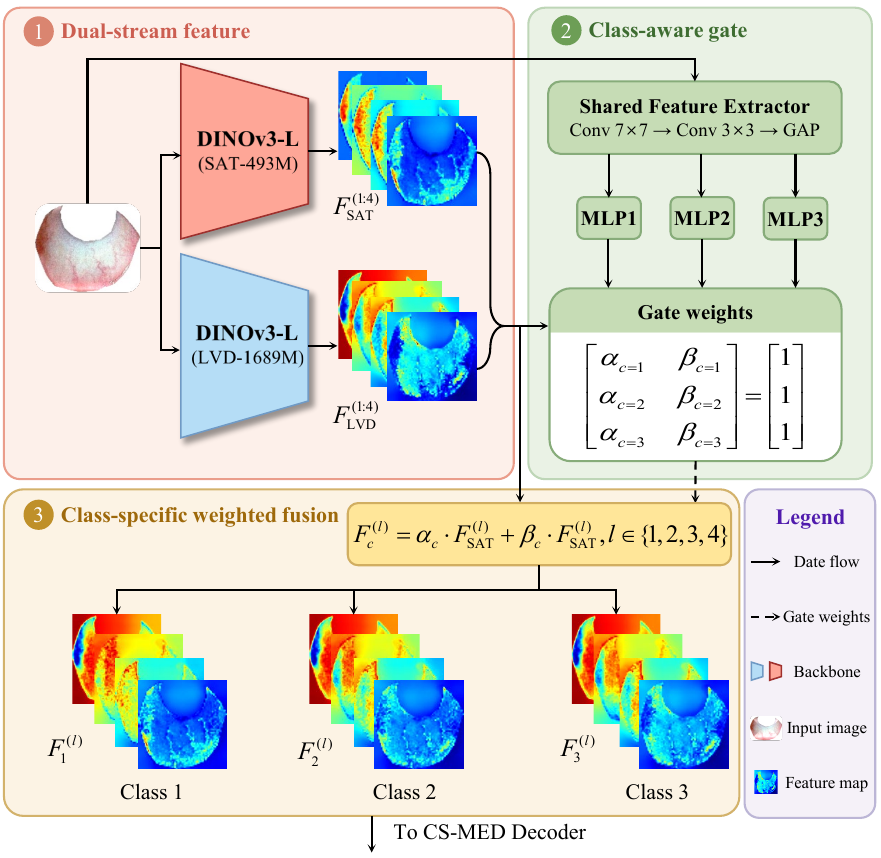}
	\caption{Architecture diagram of CA-DSGF}
	\label{fig:Architecture-CA-DSGF}
\end{figure}

To address this issue, this study proposes the \textbf{C}lass-\textbf{A}ware \textbf{D}ual-\textbf{S}tream \textbf{G}ated \textbf{F}usion Encoder (CA-DSGF).
Instead of relying on a single pretrained weight set, CA-DSGF adopts two heterogeneously pretrained DINOv3-L encoders and learns class-specific fusion weights through a class-aware gating mechanism.
Different from conventional dual-stream networks that use fixed concatenation or shared weighting for all categories, CA-DSGF leverages the multi-label formulation of this segmentation task to construct independent gating mappings for different anomaly categories, thereby generating category-specific encoded features, as shown in Fig.~\ref{fig:Architecture-CA-DSGF}.

Specifically, given a scleral image \(\mathbf{x}\in\mathbb{R}^{B\times3\times H\times W}\), the model extracts multi-scale features from the two pretrained encoders, as shown in Fig.~\ref{fig:Architecture-CA-DSGF}(1). 
The two pretrained weight sets are derived from different pretraining data, and their learned visual representations are complementary: the SAT branch tends to capture stronger spatial structures and long-range topological relationships, making it suitable for capturing elongated, topologically continuous structures such as Ve; whereas the LVD branch, pretrained on broader natural image corpora, yields more generic visual representations and stronger adaptability to complex scenarios.

The two encoders output features from predefined intermediate layers. To avoid confusion with subsequent feature-layer indices, this study uses (b) to denote the intermediate-layer index in the backbone and (l) to denote the corresponding multi-level feature index.

This study adopts the ViT-L variant of DINOv3, selects the intermediate-layer set
\({b}=\{4,11,17,23\}\), and sequentially maps it to
\(l\in\{1,2,3,4\}\). Therefore, the multi-level features output by the two encoders can be expressed as:

\begin{center}
\refstepcounter{equation}\label{eq:dual_dino_features}
\makebox[\linewidth]{%
\hfill
\makebox[0pt][c]{%
\begin{tabular}{r@{\;}c@{\;}l@{}l}
$F_{\mathrm{SAT}}^{(l)}$ & $=$ & $H_{\mathrm{SAT}}^{(b)}$ & \\[0.6em]
$F_{\mathrm{LVD}}^{(l)}$ & $=$ & $H_{\mathrm{LVD}}^{(b)}$ & 
\end{tabular}%
}%
\hfill
\llap{(\theequation)}%
}
\end{center}
where the output feature of each layer satisfies \(F^{(l)}\in\mathbb{R}^{B\times N\times D}\), where \(N=\frac{H}{16}\times\frac{W}{16}\) denotes the number of patches, and \(D=1024\) denotes the embedding dimension.

To provide the gating network with global contextual awareness while keeping computational overhead low, this study adopts a lightweight feature extractor comprising a shallow convolutional cascade and Global Average Pooling (GAP), as shown in Fig.~\ref{fig:Architecture-CA-DSGF}(2). This extractor compresses the input image into a compact image-level representation \(\mathbf{z}\):

\begin{center}
\refstepcounter{equation}\label{eq:ca_dsgf_descriptor}
\makebox[\linewidth]{%
\hfill
\makebox[0pt][c]{%
$\mathbf{z}
=
\operatorname{GAP}
\left(
\operatorname{ReLU}
\left(
\operatorname{BN}
\left(
\operatorname{Conv}_{3\times3}
\left(
\operatorname{ReLU}
\left(
\operatorname{BN}
\left(
\operatorname{Conv}_{7\times7}(\mathbf{x})
\right)
\right)
\right)
\right)
\right)
\right)
\in \mathbb{R}^{B\times64}$%
}%
\hfill
\llap{(\theequation)}%
}
\end{center}
where \(\operatorname{Conv}_{7\times7}(\cdot)\) and \(\operatorname{Conv}_{3\times3}(\cdot)\) denote the \(7\times7\) and \(3\times3\) convolution operations, respectively; \(\operatorname{BN}(\cdot)\) denotes batch normalization; \(\operatorname{ReLU}(\cdot)\) denotes the nonlinear activation function; and \(\operatorname{GAP}(\cdot)\) denotes global average pooling, which compresses spatial features into an image-level global descriptor. \(B\) denotes the batch size.
Subsequently, for each anomaly category \(c\in\{1,\ldots,C\}\), this study uses an independent MLP branch to compute the corresponding gating weight:

\begin{center}
\refstepcounter{equation}\label{eq:ca_dsgf_gate}
\makebox[\linewidth]{%
\hfill
\makebox[0pt][c]{%
\begin{tabular}{r@{\;}c@{\;}l}
$[\alpha_c,\beta_c]$ & $=$ & $\operatorname{Softmax}\left(\operatorname{MLP}_c(\mathbf{z})\right)\in\mathbb{R}^{B\times2}$\\[0.6em]
$\operatorname{MLP}_c(\mathbf{z})$ & $=$ & $\mathbf{W}_{c}^{(2)}\cdot
\operatorname{ReLU}\left(\mathbf{W}_{c}^{(1)}\cdot\mathbf{z}+\mathbf{b}_{c}^{(1)}\right)
+\mathbf{b}_{c}^{(2)}$
\end{tabular}%
}%
\hfill
\llap{(\theequation)}%
}
\end{center}
where \(C\) is the number of anomaly categories, and \(\{\mathbf{W} _{c}^{(1)},\mathbf{b}_{c}^{(1)},\mathbf{W}_{c}^{(2)},\mathbf{b}_{c}^{(2)}\}\) denotes the set of learnable parameters of the first and second layers of the MLP branch corresponding to category \(c\).
Since these parameters are explicitly indexed by category \(c\), the gating network does not use category-agnostic shared fusion weights across all categories; instead, it learns a category-specific mapping for each anomaly category.
In this way, the image-level representation \(\mathbf{z}\) can be decoupled and mapped into class-specific representational preferences.
After Softmax normalization, \([\alpha_c,\beta_c]\) satisfies \(\alpha_c+\beta_c=1\), representing the preference weights of anomaly category \(c\) for the SAT and LVD branch features, respectively.

Finally, for each category \(c\), category-aware weighted feature fusion is performed at each feature layer \(l\), as shown in Fig.~\ref{fig:Architecture-CA-DSGF}(3):

\begin{center}
\refstepcounter{equation}\label{eq:ca_dsgf_fusion}
\makebox[\linewidth]{%
\hfill
\makebox[0pt][c]{%
$F_{c}^{(l)}
=
\alpha_cF_{\mathrm{SAT}}^{(l)}
+
\beta_cF_{\mathrm{LVD}}^{(l)},
\;
l\in\{1,2,3,4\},
\;
c\in\{1,\ldots,C\}$%
}%
\hfill
\llap{(\theequation)}%
}
\end{center}
where \(F_{c}^{(l)}\) denotes the fused feature of category \(c\) at feature layer \(l\), while \(F_{\mathrm{SAT}}^{(l)}\) and \(F_{\mathrm{LVD}}^{(l)}\) correspond to the features output by the SAT branch and the LVD branch at this feature layer, respectively.
The gating weights \([\alpha_c,\beta_c]\) of category \(c\) can be viewed as an adaptive assessment of the informational utility of the two encoded feature streams.

Through this class-aware dual-stream gated fusion mechanism, CA-DSGF transforms shared visual representations into category-specific encoded representations during the encoding stage, providing a clearer feature basis for subsequent category routing and multi-expert selection in the decoder.The experimental validation is presented in Section~\ref{CA-DSGF-val}.

\subsection{Three-Stage Backbone-Frozen Routing Alignment Strategy}

\begin{figure}[pos=htbp,width=\textwidth]
	\centering
	\includegraphics[width=0.5\textwidth]{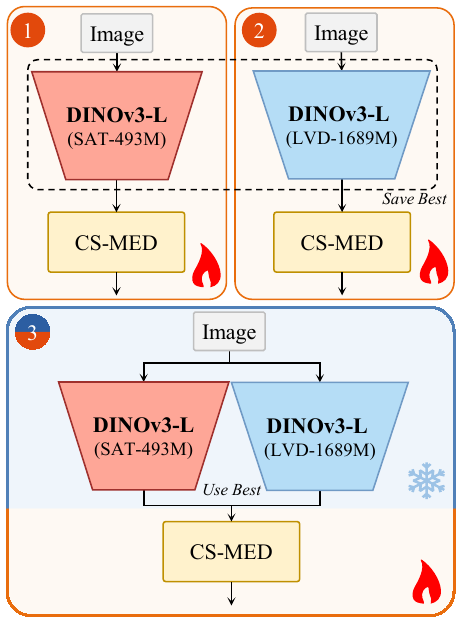}
	\caption{Design of the proposed TS-BFRA strategy: The snowflake icon denotes frozen parameters, while the flame icon denotes trainable parameters.}
	\label{fig:TS-BFRA-strategy}
\end{figure}

Given that CA-DSGF contains two large-scale DINOv3-L encoders, conventional end-to-end joint optimization is highly susceptible to systematic representation collapse induced by cross-feature gradient interference, especially before the two foundation branches have adapted to the sclera-specific data domain \citep{lyuAdvances2019, zophSTMoE2022}.

Inspired by the staged training procedure of domain-specific supervised fine-tuning (SFT) followed by global routing alignment in large-scale models \citep{NEURIPS2022_b1efde53, touvronLlama2023a}, we formulate a new training strategy, \textbf{T}hree-\textbf{S}tage \textbf{B}ackbone-\textbf{F}rozen \textbf{R}outing \textbf{A}lignment (TS-BFRA), for the dual visual foundation backbones used in this study, which collectively exceed two billion parameters.

Fig.~\ref{fig:TS-BFRA-strategy} illustrates the design of the TS-BFRA strategy. To improve the optimization stability of these large-scale foundation backbones, this study draws on the staged training strategy used in LLMs. Stage 1 and Stage 2 serve as domain-adaptation warm-up phases, in which the SAT and LVD branches are alternately frozen. By independently training the two single-branch paths, the strategy adapts the backbones from generic natural-image representations to scleral anomaly features, thereby establishing stable single-branch feature-mapping baselines and retaining the best-performing weights. Stage 3 is designed as a routing-alignment phase similar to the late-stage joint routing alignment used in MoE-based large-scale models \citep{jiangMixtral2024}. In this stage, the model freezes the backbones of both branches and exclusively fine-tunes the feature-fusion ratios in CA-DSGF and the expert-routing allocation in CS-MED, thereby enabling globally coordinated collaboration between feature fusion and expert routing. The experimental validation is presented in Section~\ref{TS-BFRA-study}.

\subsection{Class-Specific Multi-Expert Decoder}
\label{CS-MED}

\begin{figure}[pos=htbp,width=\textwidth]
	\centering
	\includegraphics[width=0.75\textwidth]{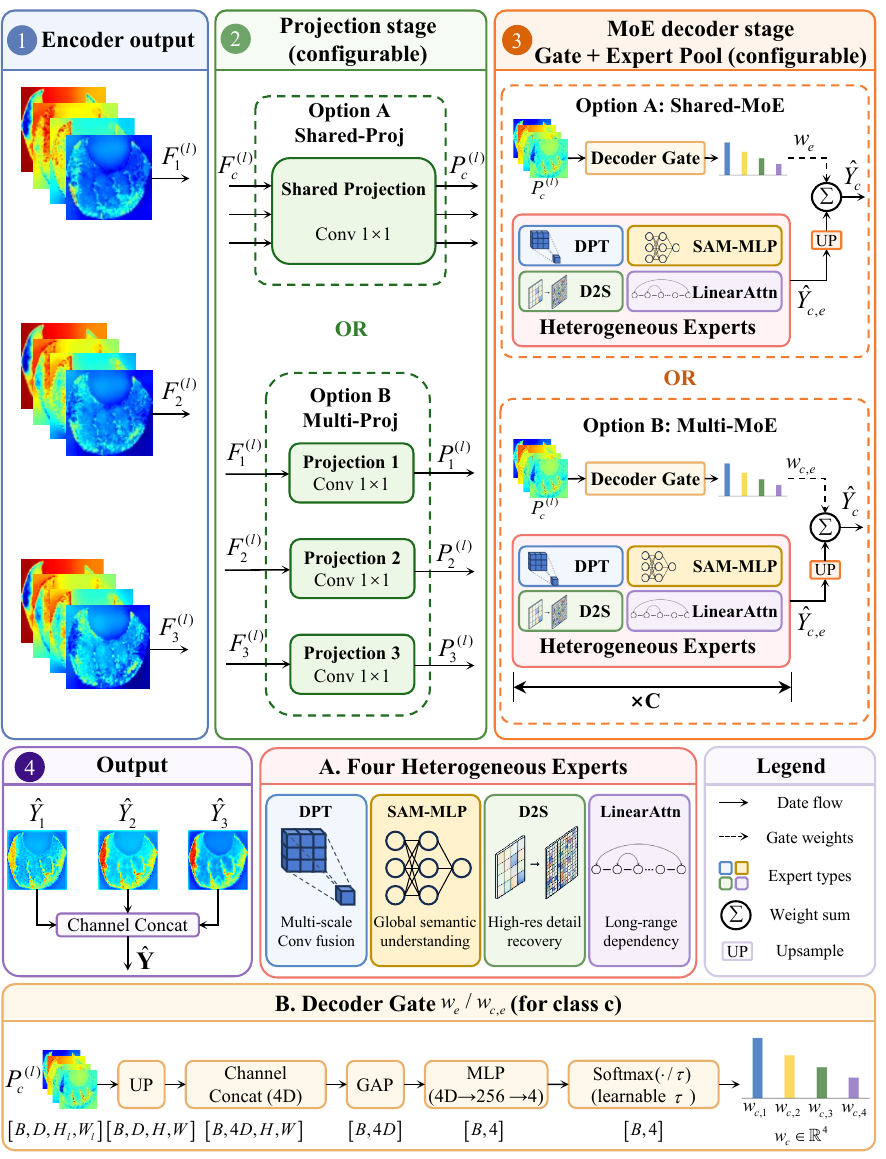}
	\caption{Architecture diagram of CS-MED}
	\label{fig:Architecture-CS-MED}
\end{figure}

As described in Section~\ref{CA-DSGF}, CA-DSGF generates class-fused features with category-level representational preferences.
However, dense prediction still requires these multi-level ViT token embeddings to be reshaped into 2D spatial feature maps and decoded in a morphology-aware manner.

Different anomaly categories also place distinct requirements on the decoder: Ve requires boundary-continuous detail recovery, YBS relies more on color-texture discrimination, and BS requires multi-scale region modeling and boundary localization.
Therefore, a single decoder with a fixed inductive bias may have limited flexibility in simultaneously addressing these differentiated segmentation requirements.

To this end, this study proposes the \textbf{C}lass-\textbf{S}pecific \textbf{M}ulti-\textbf{E}xpert \textbf{D}ecoder (CS-MED). This module adopts a set of decoding experts with distinct inductive biases and uses an MoE gating network to adaptively assign expert weights for different anomaly classes. It should be noted that ``Class-Specific'' in CS-MED does not mean that all decoding parameters are independent across classes by default. Instead, this specificity first arises from the use of class-fused features generated by CA-DSGF as inputs. Building on this, the projection layer and the MoE gating network can be further configured as either shared or class-independent, allowing us to evaluate whether additional class-level parameter specialization is still required at the decoding stage.

Fig.~\ref{fig:Architecture-CS-MED} shows the architecture of the CS-MED decoder. For each class \(c\) and each feature layer \(l\) (see Fig.~\ref{fig:Architecture-CS-MED}(1), CS-MED first reshapes the fused features from CA-DSGF into spatial feature maps and projects them through a \(1\times1\) convolution into a unified decoder input space (see Fig.~\ref{fig:Architecture-CS-MED}(2)):

\begin{center}
\refstepcounter{equation}\label{eq:csmed_projection}
\makebox[\linewidth]{%
\hfill
\makebox[0pt][c]{%
\(\displaystyle
P_c^{(l)}
=
\operatorname{Conv}_{1\times1}^{(c,l)}
\left(
\operatorname{Reshape}\left(F_c^{(l)}\right)
\right)
\in \mathbb{R}^{B\times D\times h\times w}
\)
}
\hfill
\llap{(\theequation)}%
}
\end{center}
where \(\operatorname{Reshape}(\cdot)\) restores the serialized output of \(\operatorname{ViT}\) from \(\mathbb{R}^{B\times N\times D}\) to a 2D spatial feature map \(\mathbb{R}^{B\times D\times h\times w}\). 
\(B\) denotes the batch size, \(D\) denotes the projected channel dimension, and \(h\) and \(w\) denote the spatial dimensions of the patch-level feature map. Under the DINOv3-L setting with a patch size of 16, \(h=H/16\) and \(w=W/16\) are usually obtained. 
When the \textbf{Shared-Proj} configuration is adopted, \(\operatorname{Conv}_{1\times1}^{(c,l)}\) is shared along the class dimension, meaning that different classes use the same projection parameters at feature layer \(l\). 
When the \textbf{Mult-Proj} configuration is adopted, this projection layer is class-independent.

Subsequently, the gating network computes the gating weights for the four experts based on the multi-scale projected features. 
Specifically, the projected features across multiple levels are first upsampled to a unified scale using bilinear interpolation and concatenated along the channel dimension. Global average pooling is then applied to obtain an image-level feature vector (see Fig.~\ref{fig:Architecture-CS-MED}B):

\begin{center}
\refstepcounter{equation}\label{eq:csmed_query_vector}
\makebox[\linewidth]{%
\hfill
\makebox[0pt][c]{%
\(\displaystyle
\mathbf{q}_c
=
\operatorname{Flatten}
\left(
\operatorname{GAP}
\left(
\operatorname{Concat}
\left(
\operatorname{Upsample}\left(P_c^{(1)}\right),
\ldots,
\operatorname{Upsample}\left(P_c^{(4)}\right)
\right)
\right)
\right)
\in \mathbb{R}^{B\times 4D}
\)
}
\hfill
\llap{(\theequation)}%
}
\end{center}

When the \textbf{Shared-MoE} configuration is adopted, a shared gating network \(G_{\mathrm{dec}}\) is used. 
When the \textbf{Mult-MoE} configuration is adopted, different classes use independent \(G_{\mathrm{dec},c}\).

Building on this, for each class \(c\), the decoder gating network \(G_{\mathrm{dec},c}\) uses a lightweight MLP to generate expert logits, which are further normalized into expert weights through a temperature-scaled \(\operatorname{Softmax}\) operation.
Specifically, under the \textbf{Shared-MoE} configuration, all classes share the same expert-weight vector \(w_e\); under the \textbf{Mult-MoE} configuration, each class adopts a class-specific expert-weight vector \(w_{c,e}\).

\begin{center}
\refstepcounter{equation}\label{eq:csmed_expert_weight_comparison}
\makebox[\linewidth]{%
\hfill
\makebox[0pt][c]{%
\begin{tabular}{r@{\;}c@{\;}l}
$\displaystyle
w_e
$
& $=$ &
$\displaystyle
\operatorname{Softmax}
\left(
\frac{
G_{\mathrm{dec}}\left(\left\{P_c^{(l)}\right\}_{l=1}^{4}\right)
}{\tau}
\right)
\in \mathbb{R}^{B\times E}$\\[0.8em]
$\displaystyle
w_{c,e}
$
& $=$ &
$\displaystyle
\operatorname{Softmax}
\left(
\frac{
G_{\mathrm{dec},c}\left(\left\{P_c^{(l)}\right\}_{l=1}^{4}\right)
}{\tau}
\right)
\in \mathbb{R}^{B\times E}$
\end{tabular}%
}%
\hfill
\llap{(\theequation)}%
}
\end{center}
where \(e\in\{1,\ldots,E\}\) indexes each expert, \(E=4\) denotes the number of experts, and \(\tau\) is a learnable temperature parameter used to control the smoothness of the expert-weight distribution. The MLP computation in the gating network is formulated as follows:

\begin{center}
\refstepcounter{equation}\label{eq:csmed_gate_mlp_comparison}
\makebox[\linewidth]{%
\hfill
\makebox[0pt][c]{%
\begin{tabular}{r@{\;}c@{\;}l}
$\displaystyle
G_{\mathrm{dec}}(\mathbf{q}_c)$
& $=$ &
$\displaystyle
\mathbf{W}_{2}
\operatorname{ReLU}
\left(
\mathbf{W}_{1}\mathbf{q}_c+\mathbf{b}_{1}
\right)
+\mathbf{b}_{2}
\in \mathbb{R}^{B\times E}$\\[0.8em]
$\displaystyle
G_{\mathrm{dec},c}(\mathbf{q}_c)$
& $=$ &
$\displaystyle
\mathbf{W}_{c,2}
\operatorname{ReLU}
\left(
\mathbf{W}_{c,1}\mathbf{q}_c+\mathbf{b}_{c,1}
\right)
+\mathbf{b}_{c,2}
\in \mathbb{R}^{B\times E}$
\end{tabular}%
}%
\hfill
\llap{(\theequation)}%
}
\end{center}
where \(\{\mathbf{W}_{1}, \mathbf{b}_{1}, \mathbf{W}_{2}, \mathbf{b}_{2}, \mathbf{W}_{c,1}, \mathbf{b}_{c,1}, \mathbf{W}_{c,2}, \mathbf{b}_{c,2}\}\) denotes the learnable parameter set of the lightweight MLP in the decoder gating network, including the weight matrices and bias terms of its two linear layers.

This gating mechanism enables the model to adaptively select different experts according to class-specific features. When the \textbf{Shared-MoE} configuration is adopted, a shared gating network \(G_{\mathrm{dec}}\) is used; when the \textbf{Mult-MoE} configuration is adopted, different classes use independent \(G_{\mathrm{dec},c}\).

The expert pool of CS-MED consists of four heterogeneous decoders, namely DPT, SAM-MLP, D2S, and LinearAttn, as shown in Fig.~\ref{fig:Architecture-CS-MED}A. These four experts provide complementary decoding capabilities from different perspectives. The DPT expert adopts lightweight convolution and multi-scale fusion structures, focusing on basic spatial structure recovery and local feature aggregation. The SAM-MLP expert uses a two-layer MLP with GELU activation for token-wise decoding, emphasizing global semantic understanding and mask-region discrimination. The D2S expert employs PixelShuffle for sub-pixel upsampling, thereby mitigating artifacts introduced by interpolation-based upsampling and making it better suited to high-resolution boundary detail recovery. The LinearAttn expert adopts a linear attention mechanism to model long-range dependencies with lower computational complexity, making it suitable for large-area or diffuse anomaly regions.

Therefore, the four experts are not simply duplicated homogeneous decoders; instead, they are tailored to multi-scale fusion, semantic-level understanding, detail recovery, and long-range contextual modeling, respectively. Through MoE-based weight fusion, the model dynamically aggregates the outputs of different experts conditioned on the current class-specific features.

For each class \(c\), the four experts decode the multi-scale projected features in parallel:

\begin{center}
\refstepcounter{equation}\label{eq:csmed_expert_prediction}
\makebox[\linewidth]{%
\hfill
\makebox[0pt][c]{%
\(\displaystyle
\hat{\mathbf{Y}}_{c,e}
=
D_e
\left(
\left\{\mathbf{P}_c^{(l)}\right\}_{l=1}^{4}
\right)
\in \mathbb{R}^{B\times 1\times H'\times W'},
\quad
e\in\{\mathrm{DPT},\mathrm{SAM},\mathrm{D2S},\mathrm{LA}\}
\)
}
\hfill
\llap{(\theequation)}%
}
\end{center}
where \(D_e\) denotes the \(e\)-th expert decoder, and \(\hat{\mathbf{Y}}_{c,e}\) represents the single-channel prediction map generated by this expert for class \(c\).

Subsequently, CS-MED upsamples the prediction map produced by each expert and performs expert-level weighted fusion according to the gating weights, thereby obtaining the final prediction result for class \(c\) (see Fig.~\ref{fig:Architecture-CS-MED}(3)). 
Under the \textbf{Shared-MoE} configuration, all classes share the expert weight \(w_e\); under the \textbf{Mult-MoE} configuration, each class \(c\) uses class-specific expert weights \(w_{c,e}\). 
The computation is formulated as follows:

\begin{center}
\refstepcounter{equation}\label{eq:csmed_weighted_fusion_comparison}
\makebox[\linewidth]{%
\hfill
\makebox[0pt][c]{%
\begin{tabular}{r@{\;}c@{\;}l}
$\displaystyle
\hat{\mathbf{Y}}_c
$
& $=$ &
$\displaystyle
\sum_{e=1}^{E}
w_{e}
\operatorname{Upsample}
\left(
\hat{\mathbf{Y}}_{c,e}
\right)
\in \mathbb{R}^{B\times 1\times H\times W}$\\[0.8em]
$\displaystyle
\hat{\mathbf{Y}}_c
$
& $=$ &
$\displaystyle
\sum_{e=1}^{E}
w_{c,e}
\operatorname{Upsample}
\left(
\hat{\mathbf{Y}}_{c,e}
\right)
\in \mathbb{R}^{B\times 1\times H\times W}$
\end{tabular}%
}%
\hfill
\llap{(\theequation)}%
}
\end{center}

Finally, the prediction results of all classes are concatenated along the channel dimension to obtain the overall segmentation output (see Fig.~\ref{fig:Architecture-CS-MED}(4)):

\begin{center}
\refstepcounter{equation}\label{eq:csmed_final_output}
\makebox[\linewidth]{%
\hfill
\makebox[0pt][c]{%
\(\displaystyle
\hat{\mathbf{Y}}
=
\operatorname{Concat}
\left(
\hat{\mathbf{Y}}_1,
\hat{\mathbf{Y}}_2,
\ldots,
\hat{\mathbf{Y}}_C
\right)
\in \mathbb{R}^{B\times C\times H\times W}
\)
}
\hfill
\llap{(\theequation)}%
}
\end{center}

Since ML-SASD adopts multi-label pixel-wise annotations, a single pixel can simultaneously belong to multiple anomaly classes. Accordingly, this study does not apply \(\operatorname{Softmax}\) along the class dimension for mutually exclusive classification, but instead applies \(\operatorname{Sigmoid}\) prediction independently to each class channel. This setting better accommodates the potential spatial overlap and semantic co-occurrence among Ve, YBS, and BS, as analyzed in Fig.~\ref{fig:co-occurrence} and Fig.~\ref{fig:mmd-matrix}.

To systematically evaluate whether additional class-level parameter specialization is required at the decoding stage, this study designs several optional architectural configurations in CS-MED. These configurations primarily vary in the degree of class dependence of the projection layer, the gating network, and the decoder experts:

\begin{center}
\begin{tabular}{ll}
\textbf{Configuration I:} & Shared Projection + Shared MoE Gating Network (Shared-Proj + Shared-MoE) \\
\textbf{Configuration II:} & Shared Projection + Multiple MoE Gating Networks (Shared-Proj + Mult-MoE) \\
\textbf{Configuration III:} & Multiple Projections + Multiple MoE Gating Networks (Mult-Proj + Mult-MoE)
\end{tabular}
\end{center}

The above configurations form a progressive decoding design from weak to strong class-level specialization. In this study, \textbf{Shared-MoE} indicates that all classes share the same gating network and the same four expert decoders, whereas \textbf{Mult-MoE} indicates that each class is assigned an independent MoE gating network and four expert decoders. \textbf{Shared-Proj+Shared-MoE} mainly relies on the class-aware features generated at the encoding stage and emphasizes parameter sharing and class-level knowledge reuse. \textbf{Shared-Proj+Mult-MoE} enhances class-level expert scheduling within a shared projection space. \textbf{Mult-Proj+Mult-MoE} further introduces class-independent parameters in both the projection layer and the MoE decoding stage. The ablation experiments in Section~\ref{CS-MED-study} evaluate these configurations to determine the CS-MED topology that is most suitable for the data distribution of scleral surface anomalies.

\subsection{Progressive Confidence Penalty Loss}
\label{PCP-Loss}

Before introducing \textbf{P}rogressive \textbf{C}onfidence \textbf{P}enalty Loss (PCP Loss), we first define the baseline objective for multi-label scleral anomaly segmentation.
To jointly account for pixel-level classification accuracy and region-level structural completeness under class imbalance, we formulate \(\mathcal{L}_{\mathrm{Base}}\) as a combination of binary cross-entropy (BCE) loss \citep{ronnebergerUnet2015} and Dice loss \citep{ed278621-dc3e-343f-ae66-540d8990b60d}.

The binary cross-entropy (BCE) loss (\(\mathcal{L}_{\mathrm{BCE}}\)) independently evaluates the discrepancy between the predicted probabilities and the ground-truth labels at the pixel level. 
For a multi-label segmentation task with \(C\) anomaly classes, the global BCE loss is defined as the average over all classes and all pixels:

\begin{center}
\refstepcounter{equation}\label{eq:bce_loss}
\makebox[\linewidth]{%
\hfill
\makebox[0pt][c]{%
\(\displaystyle
\mathcal{L}_{\mathrm{BCE}}
=
-\frac{1}{C\times |S|}
\sum_{c=1}^{C}
\sum_{x\in S}
\left[
y_c(x)\log \hat{p}_c(x)
+
\left(1-y_c(x)\right)\log\left(1-\hat{p}_c(x)\right)
\right]
\)
}
\hfill
\llap{(\theequation)}%
}
\end{center}
where \(S\) denotes the spatial pixel set of the image, and \(|S|=H\times W\) is the total number of pixels. 
\(y_c(x)\in\{0,1\}\) and \(\hat{p}_c(x)\in(0,1)\) represent the ground-truth label and the predicted Sigmoid probability of pixel \(x\) belonging to class \(c\), respectively.

To compensate for the pixel-wise locality limitation of BCE loss under class imbalance, we introduce the standard Dice loss \(\mathcal{L}_{\mathrm{Dice}}\). 
Dice loss measures the region-level overlap between the predicted mask and the ground-truth mask:

\begin{center}
\refstepcounter{equation}\label{eq:dice_loss}
\makebox[\linewidth]{%
\hfill
\makebox[0pt][c]{%
\(\displaystyle
\mathcal{L}_{\mathrm{Dice}}
=
\frac{1}{C}
\sum_{c=1}^{C}
\left(
1-
\frac{
2\sum_{x\in S}\hat{p}_c(x)y_c(x)+\epsilon
}{
\sum_{x\in S}\hat{p}_c(x)
+
\sum_{x\in S}y_c(x)
+\epsilon
}
\right)
\)
}
\hfill
\llap{(\theequation)}%
}
\end{center}
where \(\epsilon\) is a smoothing term set to \(1\times10^{-5}\) to prevent division by zero and stabilize gradients during the early stages of training.

\(\mathcal{L}_{\mathrm{Dice}}\) ignores the absolute number of background pixels and directly optimizes the foreground Intersection-over-Union (IoU) metric \citep{everinghamPascal2015}. 
This enables the network to focus more on sparse anomaly regions and guides the model to learn continuous morphological structures.

Leveraging the complementary advantages of these two losses, the baseline multi-label segmentation network adopts an equally weighted joint loss function for end-to-end supervised optimization:

\begin{center}
\refstepcounter{equation}\label{eq:base_loss}
\makebox[\linewidth]{%
\hfill
\makebox[0pt][c]{%
\(\displaystyle
\mathcal{L}_{\mathrm{Base}}
=
\lambda_1\mathcal{L}_{\mathrm{BCE}}
+
\lambda_2\mathcal{L}_{\mathrm{Dice}}
\)
}
\hfill
\llap{(\theequation)}%
}
\end{center}
where the loss weights are set to \(\lambda_1=\lambda_2=0.5\).

However, in complex real-world diagnostic settings, the above baseline loss function \(\mathcal{L}_{\mathrm{Base}}\) remains limited in handling spatially heterogeneous misclassification costs: it assumes a relatively uniform misclassification cost across different spatial locations. When strong Scleral Specular Reflection (SSR) interference is present in an image, the model may produce extremely high-confidence false-positive predictions within the reflection regions. In this case, the baseline BCE and Dice losses are optimized solely based on the overall discrepancy between predicted probabilities and ground-truth labels, making it difficult to impose more targeted constraints on high-confidence false positives within SSR regions. To address this issue, building on \(\mathcal{L}_{\mathrm{Base}}\), we incorporate a mask-guided and confidence-decoupled mechanism to construct \textbf{P}rogressive \textbf{C}onfidence \textbf{P}enalty Loss (PCP Loss).

The core objective of PCP Loss is to adaptively increase the penalty strength in SSR negative-sample regions conditioned on the model prediction confidence, thereby suppressing high-confidence false-positive predictions within these regions. 
This objective is implemented by introducing a dynamic pixel-level penalty weight \(w(x)\). 
As the model prediction confidence \(\hat{p}\) increases, the penalty weight increases nonlinearly.

Given the probability map predicted by the model \(\hat{p}=\sigma(z)\), and the pre-extracted binary mask of the reflection regions \(M_g\in\{0,1\}^{H\times W}\), the pixel-level progressive penalty weight is defined as follows:

\begin{center}
\refstepcounter{equation}\label{eq:pcp_weight}
\makebox[\linewidth]{%
\hfill
\makebox[0pt][c]{%
\(\displaystyle
w(x)
=
1+
(\lambda-1)\cdot
\hat{p}_d(x)^{\gamma}
\cdot
\mathbf{M}_g(x)
\)
}
\hfill
\llap{(\theequation)}%
}
\end{center}
where \(\lambda\) controls the maximum penalty magnitude and is generally set to \(2\)-\(4\), while \(\gamma\) denotes the steepness decay factor that controls the shape of the penalty curve.
\(\hat{p}_d(\cdot)\) represents the predicted probability subjected to a stop-gradient operation. Directly incorporating the predicted probability without a stop-gradient operation into the weight function would cause the weight term itself to participate in backpropagation, thereby introducing an unintended gradient path. 
To prevent the dynamic weight function from being directly optimized by the model, this study applies a stop-gradient operation to the predicted probability, allowing it to participate in loss computation only as a pixel-level reweighting factor in the current iteration. 
This design improves training stability.

Upon obtaining the dynamic weight map \(w(x)\), we reformulate the baseline joint loss by introducing spatial weighting.

The weighted binary cross-entropy (BCE) loss \(\mathcal{L}_{\mathrm{BCE}}^w\) is formulated by incorporating the adaptive weights into the standard BCE loss:

\begin{center}
\refstepcounter{equation}\label{eq:weighted_bce_loss}
\makebox[\linewidth]{%
\hfill
\makebox[0pt][c]{%
\(\displaystyle
\mathcal{L}_{\mathrm{BCE}}^w
=
-\frac{1}{C\times |S|}
\sum_{c=1}^{C}
\sum_{x\in S}
w(x)
\left[
y_c(x)\log \hat{p}_c(x)
+
\left(1-y_c(x)\right)\log\left(1-\hat{p}_c(x)\right)
\right]
\)
}
\hfill
\llap{(\theequation)}%
}
\end{center}

The weighted Dice loss \(\mathcal{L}_{\mathrm{Dice}}^w\) applies spatial weighting to both the intersection and union terms of the Dice coefficient for each class \(c\), thereby optimizing the morphological structure of the overall segmentation mask:

\begin{center}
\refstepcounter{equation}\label{eq:weighted_dice_score}
\makebox[\linewidth]{%
\hfill
\makebox[0pt][c]{%
\(\displaystyle
\mathrm{Dice}_c^w
=
\frac{
2\sum_{x\in S}w(x)\cdot \hat{p}_c(x)\cdot y_c(x)+\epsilon
}{
\sum_{x\in S}w(x)\cdot \hat{p}_c(x)
+
\sum_{x\in S}w(x)\cdot y_c(x)
+\epsilon
}
\)
}
\hfill
\llap{(\theequation)}%
}
\end{center}

\begin{center}
\refstepcounter{equation}\label{eq:weighted_dice_loss}
\makebox[\linewidth]{%
\hfill
\makebox[0pt][c]{%
\(\displaystyle
\mathcal{L}_{\mathrm{Dice}}^w
=
\frac{1}{C}
\sum_{c=1}^{C}
\left(
1-\mathrm{Dice}_c^w
\right)
\)
}
\hfill
\llap{(\theequation)}%
}
\end{center}

The final PCP Loss is formulated as a combination of the weighted BCE loss and the weighted Dice loss, serving as a reflection-suppression enhancement term for the baseline segmentation objective during the specified training stage:

\begin{center}
\refstepcounter{equation}\label{eq:pcp_loss}
\makebox[\linewidth]{%
\hfill
\makebox[0pt][c]{%
\(\displaystyle
\mathcal{L}_{\mathrm{PCP}}
=
\lambda_1\mathcal{L}_{\mathrm{BCE}}^w
+
\lambda_2\mathcal{L}_{\mathrm{Dice}}^w
\)
}
\hfill
\llap{(\theequation)}%
}
\end{center}
where the weight coefficients are set to \(\lambda_1=\lambda_2=0.5\). 
The ablation experiments on \(\lambda\) and \(\gamma\) are further conducted in Section~\ref{PCP-loss-ex} to determine the PCP Loss parameters suitable for the data distribution of scleral surface anomalies.

\subsection{Class-Aware Adaptive Sample Weighting}

\begin{figure}[pos=htbp,width=\textwidth]
	\centering
	\includegraphics[width=0.8\textwidth]{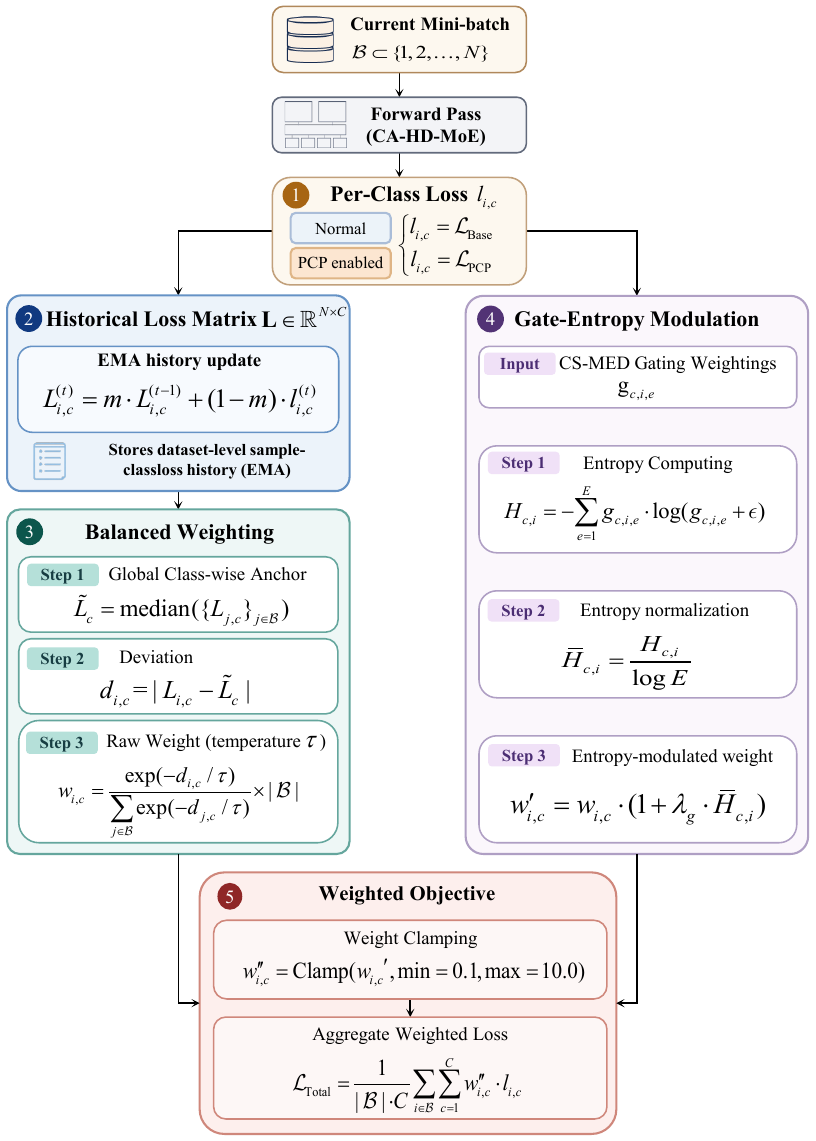}
	\caption{CA-ASW Loss Overall Process}
	\label{fig:CA-ASW}
\end{figure}

In multi-label medical image segmentation, sample-wise and class-wise difficulty may interact with each other.
Conventional sample-weighting strategies usually maintain a scalar-valued loss vector \(\mathbf{L}\in\mathbb{R}^{N}\) and average the errors across all classes, which may obscure class-specific difficulty disparities.
After PCP Loss is introduced, directly averaging class-wise losses may further dilute the spatial noise-suppression information encoded in \(\mathcal{L}_{\mathrm{PCP}}\).

To address this, this study proposes \textbf{C}lass-\textbf{A}ware \textbf{A}daptive \textbf{S}ample \textbf{W}eighting (CA-ASW), which extends sample difficulty estimation from a scalar-valued loss space \(\mathbf{L}\in\mathbb{R}^{N}\) to a matrix-valued loss space \(\mathbf{L}\in\mathbb{R}^{N\times C}\).
By maintaining a sample-by-class historical loss matrix and using class-wise training loss as the basis for difficulty estimation, CA-ASW can simultaneously adjust loss weights along both the sample and class dimensions. Instead of assigning a single weight to each image, CA-ASW assigns an independent weight to each "sample-class" combination, thereby preserving class-specific difficulty profiles in multi-label segmentation.

Fig.~\ref{fig:CA-ASW} illustrates the overall process of the CA-ASW strategy. 
Specifically, this process starts from the ``Current Mini-batch'' module, the model samples a mini-batch for forward optimization, while CA-ASW constructs a reference set \(\mathcal{B}\) from the stored historical loss matrix for weight estimation.

The input to the CA-ASW module is \(l_{i,c}\), which is derived from the loss functions \(\mathcal{L}_{\mathrm{Base}}\) and \(\mathcal{L}_{\mathrm{PCP}}\) in Section~\ref{PCP-Loss}. 
When PCP Loss is not used, the input is defined as:

\begin{center}
\refstepcounter{equation}\label{eq:ca_asw_base_input}
\makebox[\linewidth]{%
\hfill
\makebox[0pt][c]{%
\(\displaystyle
l_{i,c}
=
\mathcal{L}_{\mathrm{Base}}^{(i,c)}
\)
}
\hfill
\llap{(\theequation)}%
}
\end{center}

When PCP Loss is used, the input is defined as:

\begin{center}
\refstepcounter{equation}\label{eq:ca_asw_pcp_input}
\makebox[\linewidth]{%
\hfill
\makebox[0pt][c]{%
\(\displaystyle
l_{i,c}
=
\mathcal{L}_{\mathrm{PCP}}^{(i,c)}
\)
}
\hfill
\llap{(\theequation)}%
}
\end{center}

To track the long-term, class-specific learning difficulty of individual samples and filter out the random noise introduced by a single batch, CA-ASW maintains a persistent global historical loss matrix \(\mathbf{L}\in\mathbb{R}^{N\times C}\). 
In each iteration, this matrix is updated via an exponential moving average (EMA) using the incoming PCP Loss value:

\begin{center}
\refstepcounter{equation}\label{eq:ca_asw_ema_update}
\makebox[\linewidth]{%
\hfill
\makebox[0pt][c]{%
\(\displaystyle
L_{i,c}^{(t)}
=
\begin{cases}
l_{i,c}^{(t)}, & t=1, \\[0.4em]
m\cdot L_{i,c}^{(t-1)}+(1-m)\cdot l_{i,c}^{(t)}, & t\geq 2.
\end{cases}
\)
}
\hfill
\llap{(\theequation)}%
}
\end{center}
where \(m\in[0,1)\) denotes the momentum coefficient, with a default value of \(m=0.9\).

Upon obtaining the smoothed historical loss matrix, CA-ASW assigns sample-class weights to the samples in the current mini-batch by comparing their historical losses with class-wise anchors estimated from the global loss history.
In the CA-ASW framework, sample-class weights can be estimated using different strategies. 
This study considers various weighting schemes, including Hard, Simple, Balanced, and Curriculum, and analyzes their respective impacts on performance through ablation experiments in Section~\ref{CA-ASW study}.

This study adopts the Balanced strategy for weight allocation. For each class, the strategy first estimates a robust historical anchor from the global sample-class loss matrix rather than from the samples in the current mini-batch. The deviation between each sample-class historical loss and the corresponding class-wise anchor is then used to determine its adaptive weight.
Specifically, for each class \(c\), CA-ASW computes the median of the historical losses within the reference set \(\mathcal{B}\) retrieved from the global loss history as the normal-difficulty baseline:

\begin{center}
\refstepcounter{equation}\label{eq:ca_asw_median}
\makebox[\linewidth]{%
\hfill
\makebox[0pt][c]{%
\(\displaystyle
\tilde{L}_{c}
=
\operatorname{median}
\left(
\left\{L_{j,c}\right\}_{j\in\mathcal{B}}
\right)
\)
}
\hfill
\llap{(\theequation)}%
}
\end{center}
where \(L_{j,c}\) denotes the historical loss of the \(j\)-th sample in the reference set \(\mathcal{B}\) for class \(c\).

The absolute deviation of each sample from the median baseline is then calculated as follows:

\begin{center}
\refstepcounter{equation}\label{eq:ca_asw_deviation}
\makebox[\linewidth]{%
\hfill
\makebox[0pt][c]{%
\(\displaystyle
d_{i,c}
=
\left|L_{i,c}-\tilde{L}_{c}\right|
\)
}
\hfill
\llap{(\theequation)}%
}
\end{center}

Weights are allocated based on each sample's absolute deviation, where a larger deviation yields a lower assigned weight:

\begin{center}
\refstepcounter{equation}\label{eq:ca_asw_balanced_weight}
\makebox[\linewidth]{%
\hfill
\makebox[0pt][c]{%
\(\displaystyle
w_{i,c}
=
\frac{
\exp\left(-d_{i,c}/\tau\right)
}{
\sum_{j\in\mathcal{B}}
\exp\left(-d_{j,c}/\tau\right)
}
\times
|\mathcal{B}|
\)
}
\hfill
\llap{(\theequation)}%
}
\end{center}

The scaling multiplier \(|\mathcal{B}|\) keeps the average weight over the reference set close to \(1\), thereby stabilizing the overall loss scale.

To establish an underlying collaborative feedback loop between sample weighting and the MoE expert-allocation behavior in Section~\ref{CS-MED}, this study introduces a routing-entropy modulation mechanism. 
In the CS-MED decoder described in Section~\ref{CS-MED}, the gating network for class \(c\) outputs the expert-routing weights for \(E\) experts, denoted as \(g_{c}=\operatorname{Softmax}(G_{\mathrm{dec},c}(\mathbf{q}_{c}))\).

Here, the expert-routing weight tensor output by CS-MED is defined as \(\mathbf{g}_{c}\in\mathbb{R}^{|\mathcal{B}|\times E}\). 
For a specific sample \(i\) and expert \(e\), the corresponding routing probability is denoted as \(g_{c,i,e}\).

We quantify this uncertainty by computing the Shannon entropy of the routing probabilities:

\begin{center}
\refstepcounter{equation}\label{eq:ca_asw_entropy}
\makebox[\linewidth]{%
\hfill
\makebox[0pt][c]{%
\(\displaystyle
H_{c,i}
=
-\sum_{e=1}^{E}
g_{c,i,e}\cdot
\log\left(g_{c,i,e}+\epsilon\right)
\)
}
\hfill
\llap{(\theequation)}%
}
\end{center}

To decouple the entropy scale from the number of experts \(E\), we divide the entropy by its theoretical maximum \(\log E\) to yield the normalized value \(\bar{H}_{c,i}\in[0,1]\):

\begin{center}
\refstepcounter{equation}\label{eq:ca_asw_normalized_entropy}
\makebox[\linewidth]{%
\hfill
\makebox[0pt][c]{%
\(\displaystyle
\bar{H}_{c,i}
=
\frac{H_{c,i}}{\log E}
\)
}
\hfill
\llap{(\theequation)}%
}
\end{center}

Finally, the normalized entropy is used as a modulation term and directly incorporated into the baseline Balanced sample weights:

\begin{center}
\refstepcounter{equation}\label{eq:ca_asw_entropy_modulation}
\makebox[\linewidth]{%
\hfill
\makebox[0pt][c]{%
\(\displaystyle
w'_{i,c}
=
w_{i,c}\cdot
\left(
1+\lambda_{g}\cdot \bar{H}_{c,i}
\right)
\)
}
\hfill
\llap{(\theequation)}%
}
\end{center}
where \(\lambda_g\) denotes the modulation-strength coefficient, with a default value of 0.5. When the CS-MED gating network shows high uncertainty for a given "sample-class" combination (\(\bar{H} _{c,i}\rightarrow1\)), CA-ASW increases the final sample weight \(w'_{i,c}\) for this combination according to the entropy modulation term \(\lambda_g\). 
This design assigns greater optimization emphasis to boundary samples with high expert-allocation uncertainty, thereby helping the gating network learn more decisive expert-allocation patterns in subsequent iterations.

After obtaining the final modulated weight \(w'_{i,c}\) through gating entropy, directly multiplying it with the inner loss risks inducing numerical instability during the early stages of training. 
Since the weights are dynamically computed based on the historical matrix and amplified through an exponential mapping, extreme distributions may produce excessively large weights, triggering gradient explosion, or excessively small weights, resulting in gradient vanishing or even causing certain samples to be entirely ignored.

To ensure numerical stability during optimization, this study introduces a hard weight-clamping mechanism before final loss aggregation:

\begin{center}
\refstepcounter{equation}\label{eq:ca_asw_weight_clamping}
\makebox[\linewidth]{%
\hfill
\makebox[0pt][c]{%
\(\displaystyle
w''_{i,c}
=
\operatorname{Clamp}
\left(
w'_{i,c},
\min=0.1,
\max=10.0
\right)
\)
}
\hfill
\llap{(\theequation)}%
}
\end{center}

For upper-bound clamping (\(\max=10.0\)), the maximum value prevents the model from overreacting to highly fluctuating samples, thereby reducing the risk of disrupting the feature space shaped by most samples.

Conversely, for lower-bound clamping (\(\min=0.1\)), the minimum value serves as an important regularization design. 
Even when the Balanced strategy down-weights an extreme outlier or an overly simple sample, the model does not allow its weight to decay to \(0\). 
Instead, a small residual weight is retained to maintain a basic tension in the feature space and preserve a minimum optimization contribution, thereby reducing the risk of sample omission.

After weight clamping, the dynamic sample-class weight matrix is multiplied element-wise by the inner mixed-loss matrix \(l_{i,c}\), followed by normalized aggregation over the batch and class dimensions:

\begin{center}
\refstepcounter{equation}\label{eq:ca_asw_total_loss}
\makebox[\linewidth]{%
\hfill
\makebox[0pt][c]{%
\(\displaystyle
\mathcal{L}_{\mathrm{Total}}
=
\frac{1}{|\mathcal{B}|\cdot C}
\sum_{i\in\mathcal{B}}
\sum_{c=1}^{C}
w''_{i,c}\cdot l_{i,c}
\)
}
\hfill
\llap{(\theequation)}%
}
\end{center}

Consequently, this study establishes a two-level loss framework that combines pixel-level reflection constraints with sample-class-level reweighting. 
The inner loss focuses on suppressing high-confidence false-positive predictions within reflection regions, whereas the outer CA-ASW module adaptively adjusts the training contribution of different sample-class combinations according to the historical loss distribution. 
This hierarchical design helps mitigate the influence of local reflection interference and extreme samples on the optimization process. 
In Section~\ref{CA-ASW study}, ablation experiments are conducted to compare different strategies and parameters, thereby determining an experimental configuration suitable for CA-ASW.

\section{Experiments}
\label{Experiments}

\subsection{Implementation Details}

The proposed method was implemented in PyTorch 2.4.1. All experiments were conducted on an Ubuntu 20.04 server equipped with eight NVIDIA Tesla V100 16GB GPUs, with the software environment comprising Python 3.10.16, CUDA 12.1, and cuDNN 9.1.0.70.

Unless otherwise specified, the ML-SASD-Clinical and ML-SASD-Wild subsets were each partitioned into training, validation, and test sets at an approximate ratio of 8:1:1. 
The ML-SASD-Mix subset was formed by aggregating the corresponding training, validation, and test sets from both subsets.

During dataset partitioning, subject-level and acquisition-sequence-level grouping was considered whenever subject or acquisition identifiers were available, ensuring that multi-gaze images from the same subject or the same five-eye-position acquisition sequence were assigned to the same split.

All images and masks were uniformly resized to \(1024 \times 1024\) pixels using bilinear and nearest-neighbor interpolation, respectively. 
The main implementation details and hyperparameter settings of HD-DinoMoE are summarized in Table~\ref{tab:Implementation}.

\begin{table}[pos=htbp,width=\textwidth]
  \centering
  \caption{Implementation details and hyperparameter settings of HD-DinoMoE}
    \begin{tabular}{ccc}
    \toprule
    Hyperparameters & Value & Note \\
    \midrule
    Backbone & DINOv3 ViT-L/16 & Pre-trained on SAT-493M \& LVD-1689M \\
    Decoder channels & 256   & 	Intermediate feature dimension for all decoders \\
    Input resolution & \(1024 \times 1024\) & Bilinear resize for images, nearest for masks \\
    Normalization & ImageNet \((\mu,\sigma)\) & \(\mu=(0.485, 0.456, 0.406)\), \(\sigma=(0.229, 0.224, 0.225)\) \\
    Optimizer & AdamW & 	With decoupled weight decay \\
    Learning rate & \(1\mathrm{e}{-4}\) & Stage 1 \& 2; lower for Stage 3 fine-tuning \\
    Weight decay & 	\(1\mathrm{e}{-4}\) & Applied to all trainable parameters \\
    LR schedule & Constant & No decay scheduler used \\
    Batch size & 1     & 	Due to large model size \\
    Epochs (Stage 1) & 50    & SAT branch training, LVD backbone frozen \\
    Epochs (Stage 2) & 50    & LVD branch training, SAT backbone frozen \\
    Epochs (Stage 3) & 50    & Routing alignment with both backbones frozen \\
    Model selection & Best val. mDice & Val. selects checkpoint; test reports final results \\
    Num workers & 	8    & 	DataLoader parallelism \\
    \bottomrule
    \end{tabular}%
  \label{tab:Implementation}%
\end{table}%

\subsection{Evaluation Metrics}
\label{Evaluation-Metrics}

To comprehensively evaluate the performance of HD-DinoMoE in the multi-label scleral anomaly segmentation task, this study adopts four core evaluation metrics: Mean Intersection-over-Union (mIoU)\citep{everinghamPascal2015}, Mean Dice Coefficient (mDice)\citep{ed278621-dc3e-343f-ae66-540d8990b60d}, Mean Boundary F1 Score (mBF1)\citep{perazziBenchmark2016}, and Mean Glare False Positive Rate (mGFPR).

Since the ML-SASD dataset features multi-label pixel-wise annotations, all metrics are computed independently for each class channel. Following \(\operatorname{Sigmoid}\) activation of the model outputs, a fixed threshold is applied to generate a binary prediction mask for each class, without using \(\operatorname{Softmax}\) or \(\operatorname{Argmax}\) operations along the class dimension.

\textbf{1) Region-level Segmentation Accuracy Metrics (mIoU \& mDice)}

mIoU and mDice are fundamental metrics for evaluating the overall region-level agreement of semantic segmentation models. For a task containing \(C\) classes, these metrics are computed class-wise and subsequently averaged. mIoU measures the overlap between the predicted region and the ground-truth region, and is calculated as follows:

\begin{center}
\refstepcounter{equation}\label{eq:miou}
\makebox[\linewidth]{%
\hfill
\makebox[0pt][c]{%
$\displaystyle
mIoU
=
\frac{1}{C}
\sum_{c=1}^{C}
\frac{TP_c}{TP_c+FP_c+FN_c}
$%
}%
\hfill
\llap{(\theequation)}%
}
\end{center}

mDice imposes a more balanced penalty on false positives (FPs) and false negatives (FNs), and is widely used in medical image segmentation to evaluate model sensitivity to small targets:

\begin{center}
\refstepcounter{equation}\label{eq:mdice}
\makebox[\linewidth]{%
\hfill
\makebox[0pt][c]{%
$\displaystyle
mDice
=
\frac{1}{C}
\sum_{c=1}^{C}
\frac{2\cdot TP_c}{2\cdot TP_c+FP_c+FN_c}
$%
}%
\hfill
\llap{(\theequation)}%
}
\end{center}
where \(TP_c\), \(FP_c\), and \(FN_c\) denote the numbers of true positives, false positives, and false negatives at the pixel level for class \(c\), respectively.

\textbf{2) Boundary Localization Metric (mBF1)}

In addition to conventional region-level overlap metrics, boundary localization accuracy is also important for subsequent auxiliary diagnostic analysis of scleral surface anomalies. In the downstream task of this study, subtle anomaly regions in different scleral areas are further used for eye-organ mapping, where even minor boundary shifts may lead to erroneous organ mapping. Therefore, this study introduces mBF1 to evaluate boundary segmentation quality more strictly.

Given the predicted boundary \(B_p\) and the ground-truth boundary \(B_g\), boundary precision and recall are defined within a tolerance distance threshold \(\theta\), which is set to 2 pixels in this study:

\begin{center}
\refstepcounter{equation}\label{eq:boundary_precision_recall}
\makebox[\linewidth]{%
\hfill
\makebox[0pt][c]{%
$\displaystyle
P_b
=
\frac{\left|B_p\cap B_g\right|}{\left|B_p\right|},
\qquad
R_b
=
\frac{\left|B_p\cap B_g\right|}{\left|B_g\right|}
$%
}%
\hfill
\llap{(\theequation)}%
}
\end{center}

The mBF1 score across all classes is then calculated as follows:

\begin{center}
\refstepcounter{equation}\label{eq:mbf1}
\makebox[\linewidth]{%
\hfill
\makebox[0pt][c]{%
$\displaystyle
mBF1
=
\frac{1}{C}
\sum_{c=1}^{C}
\frac{2\cdot P_{b,c}\cdot R_{b,c}}{P_{b,c}+R_{b,c}}
$%
}%
\hfill
\llap{(\theequation)}%
}
\end{center}

The mBF1 metric can mitigate the numerical inflation caused by large areas of correctly segmented internal pixels and more directly evaluate whether the model accurately delineates the true contours of scleral anomalies under complex scenarios and specular reflection interference.

\textbf{3) Mean Glare False Positive Rate (mGFPR)}

To quantitatively evaluate the model's ability to suppress SSR interference, this study introduces a customized metric, mGFPR, based on the concept of false positive rate (FPR)\citep{fawcettIntroduction2006}. This metric is defined as the average pixel ratio within SSR regions in each image that the model incorrectly classifies as any foreground anomaly class:

\begin{center}
\refstepcounter{equation}\label{eq:mgfpr}
\makebox[\linewidth]{%
\hfill
\makebox[0pt][c]{%
$\displaystyle
mGFPR
=
\frac{1}{N}
\sum_{i=1}^{N}
\frac{\left|G_i\cap P_{\mathrm{lesion},i}\right|}{\left|G_i\right|}
\times 100\%
$%
}%
\hfill
\llap{(\theequation)}%
}
\end{center}
where \(G_i\) denotes the pixel set of SSR regions in the \(i\)-th image, and \(P_{\mathrm{lesion},i}\) denotes the set of pixels predicted by the model as any foreground anomaly class, namely Ve, YBS, or BS. 
A lower mGFPR indicates stronger suppression of specular reflection interference.

Considering the resolution discrepancy and small-object characteristics of ML-SASD(see Section~\ref{Small-Object}), this study adopts a sample-level evaluation protocol in all quantitative experiments. 
Specifically, each metric is first computed independently for each image and then averaged over the held-out test set for final evaluation.
By contrast, a global-level evaluation protocol, which pools all predicted and ground-truth pixels before metric computation, risks biasing the overall scores toward categories with larger area proportions, such as large-area BS regions, thereby potentially obscuring missed detections of small and difficult anomalies, such as extremely thin Ve. 
Therefore, sample-level evaluation treats each image equally and more faithfully reflects the model's fine-grained segmentation capability for subtle anomaly features.

\subsection{Dataset Complexity and Source-wise Heterogeneity Analysis}

To objectively quantify the baseline difficulty and source-wise heterogeneity of ML-SASD, this study adopts SegDINO as a baseline evaluator and conducts quantitative evaluations across the ML-SASD-Clinical, ML-SASD-Wild, and ML-SASD-Mix subsets. Leveraging DINOv3 pretrained weights, SegDINO exhibits strong segmentation performance and can therefore serve as a relatively stable baseline model for analyzing the difficulty differences inherent to the dataset. To mitigate potential representation biases stemming from a single set of pretrained weights, this study further evaluates two configurations, namely SegDINO-SAT and SegDINO-LVD. The experimental results are shown in Table~\ref{tab:Data-Subsets}.For all tables, $\uparrow$ denotes higher-is-better and $\downarrow$ denotes lower-is-better.

\begin{table}[pos=htbp,width=\textwidth]
  \centering
  \caption{Experiments on Different Data Subsets}
    \begin{tabular}{cccccc}
    \toprule
    Weights & Dataset types & mDice($\uparrow$) & mIoU($\uparrow$) & mBF1($\uparrow$) & mGFPR($\downarrow$) \\
    \midrule
    \multirow{3}[2]{*}{LVD} & ML-SASD-Clinical & 65.94\% & 51.69\% & 33.64\% & 0.84\% \\
          & ML-SASD-Wild & 73.83\% & 60.39\% & 40.05\% & 1.45\% \\
          & ML-SASD-Mix & 70.26\% & 56.40\% & 37.96\% & 1.04\% \\
    \midrule
    \multirow{3}[2]{*}{SAT} & ML-SASD-Clinical & 65.97\% & 51.48\% & 33.68\% & 0.90\% \\
          & ML-SASD-Wild & 74.20\% & 61.00\% & 41.61\% & 1.37\% \\
          & ML-SASD-Mix & 70.18\% & 56.39\% & 38.57\% & 1.14\% \\
    \bottomrule
    \end{tabular}%
  \label{tab:Data-Subsets}%
\end{table}%

In terms of region-overlap metrics, ML-SASD-Wild achieves higher overall mDice, mIoU, and mBF1 scores than ML-SASD-Clinical. Taking SegDINO-LVD as an example, it obtains 73.83\% mDice, 60.39\% mIoU, and 40.05\% mBF1 on the ML-SASD-Wild subset, whereas the corresponding scores on the ML-SASD-Clinical subset are 65.94\%, 51.69\%, and 33.64\%, respectively. 
A similar trend is observed for SegDINO-SAT. These results suggest that although the ML-SASD-Clinical subset is acquired from relatively standardized sources, its anomaly morphology and boundary distribution are not necessarily simpler. 
By contrast, despite the more complex acquisition conditions of the ML-SASD-Wild subset, some anomaly regions exhibit stronger color contrast and regional saliency, leading to higher region-level and boundary-level segmentation scores under the current baseline models.

Compared with the single-source ML-SASD-Wild subset, the performance on ML-SASD-Mix decreases. 
For example, SegDINO-LVD achieves 70.26\% mDice and 56.40\% mIoU on ML-SASD-Mix, lower than its 73.83\% and 60.39\% on ML-SASD-Wild. 
SegDINO-SAT shows a similar decrease, with mDice and mIoU dropping from 74.20\% and 61.00\% on ML-SASD-Wild to 70.18\% and 56.39\% on ML-SASD-Mix, respectively. 
This indicates that multi-source mixing does not simply increase the number of training samples, but also introduces more complex cross-source distributional differences. 
The model needs to adapt simultaneously to appearance variations, scale differences, imaging-quality variations, and anomaly morphology differences between medical-atlas images and real smartphone-acquired images. 
Therefore, ML-SASD-Mix better reflects the combined challenges of practical deployment scenarios.

The comparison between SAT and LVD pretrained weights shows that the two configurations achieve similar overall performance, and neither consistently outperforms the other across all subsets and metrics.
This result suggests that a single visual foundation backbone is insufficient to fully cover the multi-source data distribution and multi-class anomaly morphology in ML-SASD. 
The slight performance differences between the two pretrained weights further support the design of introducing dual-stream DINOv3-L encoders in CA-DSGF and performing adaptive feature fusion through a class-aware gating mechanism.

In addition, the mGFPR results show that the false-positive rate in reflection regions is generally higher on the ML-SASD-Wild subset than on the ML-SASD-Clinical subset. For instance, SegDINO-LVD obtains an mGFPR of 1.45\% on ML-SASD-Wild, higher than 0.84\% on ML-SASD-Clinical; SegDINO-SAT also shows a higher mGFPR on ML-SASD-Wild than on ML-SASD-Clinical. 
This indicates that reflection artifacts in real smartphone-acquired images are more likely to be mis-segmented by the baseline models as anomalous foreground regions, providing experimental support for the design of PCP Loss.

Overall, ML-SASD-Mix contains both Clinical and Wild sources, and its difficulty arises not only from the increased sample scale but also from the combined effects of multi-source image distributions, anomaly morphology differences, scale variations, and reflection interference. 
Therefore, unless otherwise specified, the subsequent comparative experiments and ablation studies are conducted on ML-SASD-Mix under the same 8:1:1 training-validation-test split, where the validation set is used for checkpoint selection and the held-out test set is used for final quantitative evaluation. This protocol enables a more comprehensive assessment of model performance under multi-source heterogeneous scleral anomaly segmentation scenarios.

\subsection{Comparison with Representative Segmentation Methods}
\label{Comparison-Methods}

To systematically evaluate the effectiveness of the proposed method in the multi-source scleral anomaly segmentation task, this study selects 11 representative segmentation models as comparative baselines. These models cover different method families, including classical medical image segmentation networks, CNN/Transformer hybrid architectures, emerging network architectures, visual foundation model adaptation baselines, and medical MoE and interference-resistant segmentation methods. All experiments are conducted on ML-SASD-Mix, and the evaluation metrics include mDice, mIoU, mBF1, and mGFPR, as defined in Section~\ref{Evaluation-Metrics}.

Specifically, the selected baselines are categorized as follows: U-Net and nnU-Net are used as classical medical image segmentation baselines; TransUNet and SegNeXt represent CNN/Transformer hybrid architectures; and U-Mamba and U-KAN represent emerging network architectures. Considering that the proposed method is built upon the DINOv3 visual foundation model, this study further constructs four DINOv3 feature-adaptation baselines, namely SegDINO-SAT, SegDINO-LVD, DINOUNet-SAT, and DINOUNet-LVD, to represent visual foundation model adaptation baselines. In addition, UTANet and ConDSeg are selected to represent medical MoE and interference-resistant segmentation methods, respectively, to further compare their performance with HD-DinoMoE under complex acquisition scenarios.

\begin{figure}[pos=htbp,width=\textwidth]
	\centering
	\includegraphics[width=0.4\textwidth]{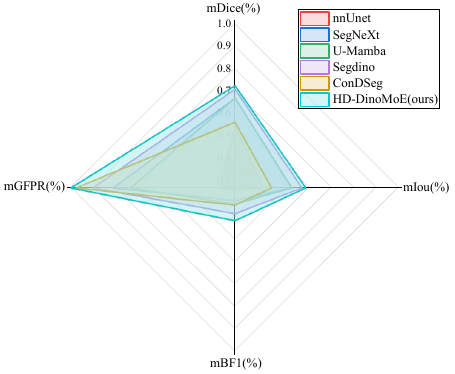}
	\caption{Radar chart comparison of the best-performing models from each group}
	\label{fig:Radar-comparison}
\end{figure}

To adapt the comparative models to ML-SASD, all baselines are uniformly configured in a multi-label segmentation form. 
Specifically, the segmentation heads of all models are set to three output channels corresponding to Ve, YBS, and BS, while class-wise Sigmoid activation and class-wise multi-head BCE + Dice loss are adopted at the output and optimization stages, respectively, instead of applying Softmax along the class dimension. 
To ensure a fair comparison, all models share the same training-validation-test split, input resolution, checkpoint-selection protocol, and evaluation metrics.
The four models in the visual foundation model category all use DINOv3 ViT-L/16 weights. 
Since both ConDSeg and HD-DinoMoE include negative-sample modeling mechanisms, the same SSR masks are used during training. 
Conversely, baselines lacking negative-sample input interfaces are trained only with regular segmentation annotations. 
Except for the necessary adaptations described above, the remaining architecture-related hyperparameters of each model adhere to the default configurations of their respective original methods wherever applicable. 
The experimental results are shown in Table~\ref{tab:Comparative-Experiments}, and Fig.~\ref{fig:Radar-comparison} presents the radar chart comparison of the best-performing models from each group.

\begin{table}[pos=htbp,width=\textwidth]
  \centering
  \caption{Comparative Experiments}
  \label{tab:Comparative-Experiments}
  \resizebox{\textwidth}{!}{%
    \begin{tabular}{cccccc}
    \toprule
    Category & Model & mDice($\uparrow$) & mIoU($\uparrow$) & mBF1($\uparrow$) & mGFPR($\downarrow$) \\
    \midrule
    \multirow{2}[1]{*}{1} 
          & U-Net~\citep{ronnebergerUnet2015} & 53.04\% & 39.32\% & 25.44\% & 2.28\% \\
          & nnU-Net~\citep{isenseeNnUNet2021} & 53.92\% & 39.50\% & 27.64\% & 1.86\% \\
    \midrule
    \multirow{2}[0]{*}{2} 
          & TransUNet~\citep{chenTransUNet2024} & 55.96\% & 41.27\% & 28.93\% & 1.70\% \\
          & SegNeXt~\citep{NEURIPS2022_08050f40} & 66.49\% & 52.04\% & 32.47\% & 1.26\% \\   
    \midrule
    \multirow{2}[0]{*}{3} 
          & U-Mamba~\citep{maUmamba2024} & 66.62\% & 52.20\% & 34.04\% & 1.35\% \\
          & U-KAN~\citep{liUKAN2025} & 62.45\% & 48.19\% & 31.51\% & 1.67\% \\
    \midrule
    \multirow{4}[0]{*}{4} 
          & SegDINO-SAT~\citep{yangSegDINO2025} & \underline{70.44\%} & \underline{56.63\%} & \underline{38.26\%} & 1.14\% \\
          & SegDINO-LVD~\citep{yangSegDINO2025} & 70.06\% & 56.21\% & \underline{38.26\%} & \underline{1.04\%} \\
          & DINOUnet-SAT~\citep{gaoDino2025} & 69.21\% & 55.02\% & 36.30\% & 1.32\% \\
          & DINOUnet-LVD~\citep{gaoDino2025} & 69.75\% & 55.63\% & 37.49\% & 1.37\% \\
    \midrule
    \multirow{2}[0]{*}{5} 
          & UTANet~\citep{luoRethinking2025} & 50.80\% & 38.54\% & 25.16\% & 1.42\% \\
          & ConDSeg~\citep{leiConDSeg2025} & 55.86\% & 43.13\% & 34.47\% & 1.05\% \\
    \midrule
    6     & HD-DinoMoE (Ours) & \textbf{72.11\% (+1.67\%)} & \textbf{58.44\% (+1.81\%)} & \textbf{41.40\% (+3.14\%)} & \textbf{1.02\% (-0.02\%)} \\
    \bottomrule
    \end{tabular}%
  }
\end{table}

\begin{figure}[pos=htbp,width=\textwidth]
	\centering
	\includegraphics[width=0.7\textwidth]{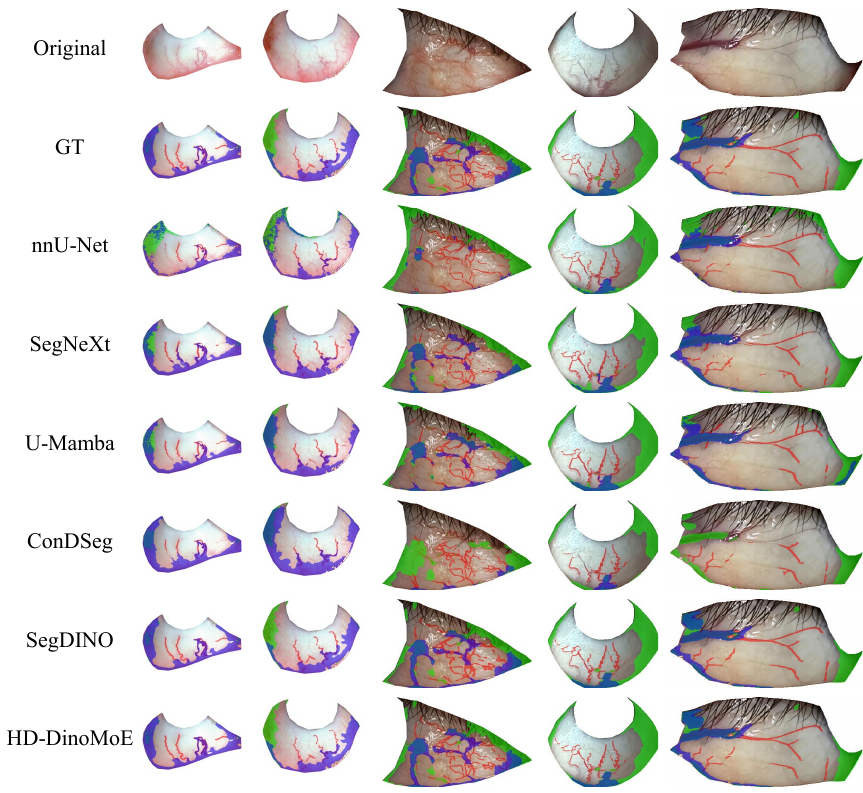}
	\caption{Visual comparison of representative segmentation methods on ML-SASD-Mix}
	\label{fig:Visual-comparison}
\end{figure}

As shown in Table~\ref{tab:Comparative-Experiments}, HD-DinoMoE achieves the best performance across all four metrics. 
Compared with the second-best baseline, HD-DinoMoE improves mDice, mIoU, and mBF1 by 1.67\%, 1.81\%, and 3.14\%, respectively, while suppressing mGFPR to 1.02\%. 
Among these metrics, the performance gain in mBF1 is particularly prominent, indicating that the proposed method enhances not only region-level overlap accuracy but also the boundary delineation capability for scleral anomalies.

Conventional medical image segmentation models show relatively limited performance on ML-SASD, with U-Net and nnU-Net achieving 53.04\% and 53.92\% mDice, respectively. 
Although hybrid and emerging architectures, such as TransUNet, SegNeXt, U-Mamba, and U-KAN, improve over classical CNN baselines, with U-Mamba reaching 66.62\% mDice and 52.20\% mIoU, their overall performance still lags behind visual foundation model-based methods. 
This underscores that, for multi-source heterogeneous scleral anomaly segmentation, large-scale visual pretrained features can provide more effective and transferable foundational representations.

Among the DINOv3-related baselines, SegDINO-SAT and SegDINO-LVD achieve 70.44\% and 70.06\% mDice, respectively, clearly outperforming most non-foundation-model counterparts. 
This indicates that DINOv3 features have strong adaptability to scleral anomaly segmentation. 
However, HD-DinoMoE further improves the performance to 72.11\% mDice and 58.44\% mIoU, suggesting that using a single DINOv3 feature stream or a simple decoding structure is still insufficient to fully address the morphological variations across multiple anomaly classes and the cross-source domain discrepancies in multi-source scleral images.

In terms of reflection-induced false positives, ConDSeg and SegDINO-LVD achieve relatively low mGFPR values of 1.05\% and 1.04\%, respectively; however, their holistic segmentation performance remains inferior to that of HD-DinoMoE. 
HD-DinoMoE achieves the lowest mGFPR and the best region-level segmentation performance, indicating that the proposed method enhances overall segmentation accuracy while further reducing the risk of SSR regions being falsely segmented as anomalous foreground.

To further compare the prediction behavior across methods, Fig.~\ref{fig:Visual-comparison} presents multi-label segmentation visualizations on several representative examples. 
In the figure, color blending is used to show regions with class overlap. 
As shown in Fig.~\ref{fig:Visual-comparison}, traditional segmentation models exhibit certain limitations in maintaining the continuity of elongated Ve structures and delineating complex boundaries. 
Visual foundation model-based methods generally produce more complete segmentation results, but misclassifications may still occur in regions with multi-class overlap and in the vicinity of specular reflection artifacts. 
In contrast, HD-DinoMoE maintains more continuous Ve structures in most examples and produces more stable predictions of the spatial extents of BS and YBS. 
These qualitative observations are generally consistent with the quantitative trends of mDice, mIoU, mBF1, and mGFPR reported in Table~\ref{tab:Comparative-Experiments}.

\subsection{Ablation Studies}

To facilitate subsequent quantitative analysis and table formatting, this section first defines a unified numeric mapping for the anomaly categories: Ve, YBS, and BS are denoted as Class 1, Class 2, and Class 3, respectively.

\subsubsection{Ablation Study of CA-DSGF}
\label{CA-DSGF-val}

To validate the effectiveness of CA-DSGF in the feature extraction phase, this study designs a set of ablation experiments on encoder fusion strategies. 
The Base strategy comprises a single-branch configuration initialized with SAT weights. 
The Dual-Stream Gated Fusion (DSGF) strategy also adopts two encoder branches, namely SAT-DINOv3-L and LVD-DINOv3-L, but does not introduce class-aware gating. Rather than learning class-specific gating weights, this baseline uses a shared dual-stream fusion function across all classes.

To isolate the effect of encoder fusion strategies on the experimental results, this experiment adopts the same decoder setting and fixes the DPT decoder as the decoding head. 
Both methods are trained and evaluated under three data settings, namely ML-SASD-Clinical, ML-SASD-Wild, and ML-SASD-Mix, to analyze the performance differences of dual-stream fusion strategies across heterogeneous data sources. The experimental results are shown in Table~\ref{tab:Comparison-CA-DSGF}.

\begin{table}[pos=htbp,width=\textwidth]
  \centering
  \caption{Comparison between CA-DSGF and DSGF}
    \begin{tabular}{cccc}
    \toprule
    Strategy & Dataset types & mDice($\uparrow$) & mIoU($\uparrow$) \\
    \midrule
    \multirow{3}[1]{*}{Base} & ML-SASD-Clinical & 65.94\% & 51.69\% \\
          & ML-SASD-Wild & 73.83\% & 60.39\% \\
          & ML-SASD-Mix & 70.26\% & 56.40\% \\
    \multirow{3}[0]{*}{DSGF} & ML-SASD-Clinical & \textbf{67.17\% (+1.23\%)} & \textbf{52.87\% (+1.18\%)} \\
          & ML-SASD-Wild & \textbf{74.84\% (+1.01\%)} & \textbf{61.66\% (+1.27\%)} \\
          & ML-SASD-Mix & 70.88\% (+0.62\%) & 57.20\% (+0.80\%) \\
    \multirow{3}[1]{*}{CA-DSGF} & ML-SASD-Clinical & 67.02\% (+1.08\%) & 52.69\% (+1.00\%) \\
          & ML-SASD-Wild & 74.81\% (+0.98\%) & 61.65\% (+1.26\%) \\
          & ML-SASD-Mix & \textbf{71.01\% (+0.75\%)} & \textbf{57.31\% (+0.91\%)} \\
    \bottomrule
    \end{tabular}%
  \label{tab:Comparison-CA-DSGF}%
\end{table}%

Overall, both dual-stream fusion strategies outperform the Base strategy, indicating that jointly introducing SAT and LVD pretrained representations can provide complementary information for scleral anomaly segmentation. 
However, the two fusion strategies exhibit different trends under different data-source settings. On ML-SASD-Clinical, the improvement margins of DSGF over the Base strategy exceed those of CA-DSGF by 0.15\% in mDice and 0.18\% in mIoU. 
On ML-SASD-Wild, DSGF also achieves slightly larger gains than CA-DSGF, with improvement-margin differences of 0.03\% in mDice and 0.01\% in mIoU. 
These results suggest that, under relatively homogeneous single-source settings, a shared MLP can learn a stable global fusion pattern; therefore, the advantage of class-level dynamic gating is not yet pronounced.

When the data-source setting switches to ML-SASD-Mix, CA-DSGF surpasses the shared dual-stream fusion strategy in accuracy metrics, showing better multi-source adaptability. Specifically, compared with DSGF, CA-DSGF achieves larger improvement margins over the Base strategy by 0.13\% in mDice and 0.11\% in mIoU. 
ML-SASD-Mix contains both Clinical and Wild sources, with differences in image style, resolution distribution, anomaly saliency, class ratio, and the degree of specular reflection interference. 
In such a multi-source heterogeneous scenario, a single shared fusion function may have limited capacity to simultaneously adapt to the feature distributions of different classes and sources. 
In contrast, CA-DSGF independently learns SAT/LVD fusion weights for each anomaly class, enabling the model to dynamically adjust the contribution of the two pretrained representations according to class-specific features. Therefore, its advantage is mainly reflected in more complex multi-source mixed data.

\begin{figure}[pos=htbp,width=\textwidth]
	\centering
	\includegraphics[width=0.5\textwidth]{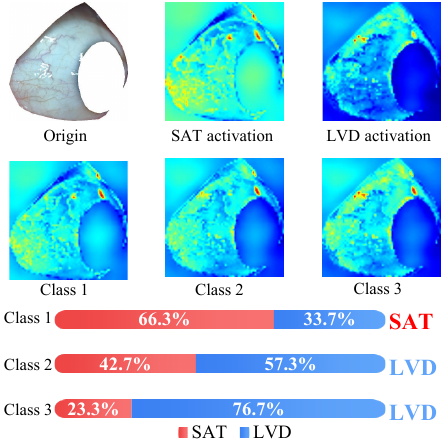}
	\caption{CA-DSGF encoder heatmap comparison}
	\label{fig:CA-DSGF-heatmap}
\end{figure}

The heatmap visualization further supports the above observations. As shown in Fig.~\ref{fig:CA-DSGF-heatmap}, SAT activation and LVD activation exhibit different activation preferences. The SAT branch shows more continuous responses over large-scale Ve structures, suggesting that it may better capture long-range spatial dependencies and macro-topological relationships. In contrast, the LVD branch produces activations that are more concentrated around local textures and boundary regions, suggesting that it may provide stronger local discriminative information. After CA-DSGF fusion, the feature maps of different classes show clearer class-related patterns, and the fusion ratios differ across classes. This indicates that the class-aware gating mechanism can perform class-specific fusion of dual-stream features at the encoding stage, rather than relying on a fixed global fusion strategy.

In summary, the advantage of CA-DSGF does not lie in fully replacing DSGF on single-source datasets, but in its stronger class-wise adaptability and feature-fusion flexibility under multi-source heterogeneous data. Therefore, this study retains this strategy for the Overall Ablation Study in Table~\ref{tab:Overall-Ablation}.

\subsubsection{Ablation Study of TS-BFRA}
\label{TS-BFRA-study}

To validate the effectiveness of TS-BFRA (Three-Stage Backbone-Frozen Routing Alignment), this study further compares the effects of different training strategies on the performance of HD-DinoMoE, focusing on how different optimization schedules affect the final segmentation performance. 
The experimental results are shown in Table~\ref{tab:Ablation-TS-BFRA}.

\begin{figure}[pos=htbp,width=\textwidth]
	\centering
	\includegraphics[width=0.8\textwidth]{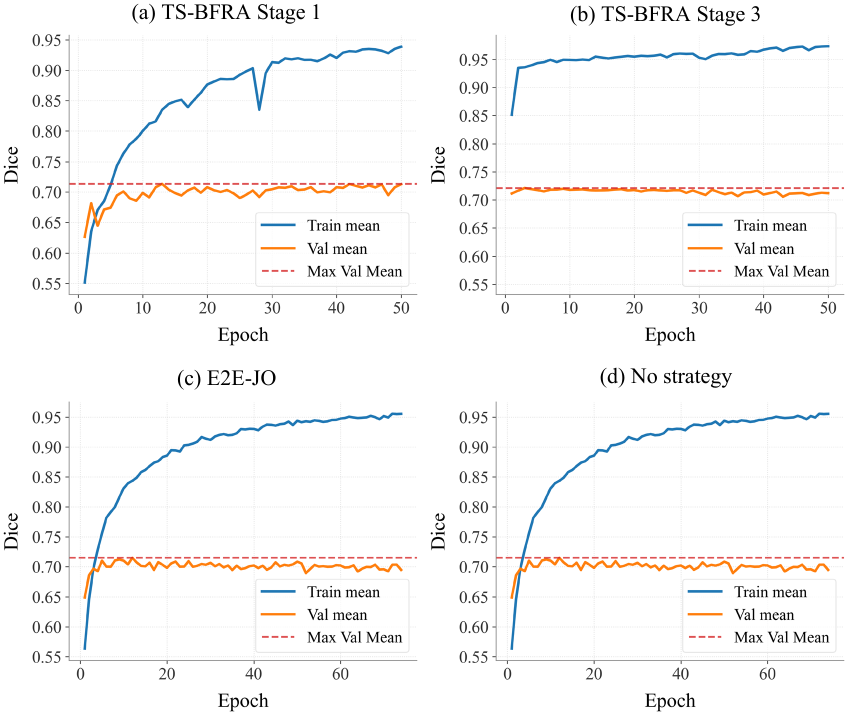}
	\caption{Dice coefficient curves under different training strategies}
	\label{fig:Dice-curves}
\end{figure}

Specifically, four training configurations are evaluated: ``No strategy'' denotes the baseline setting that uses single-branch SAT weights without introducing a dual-stream staged training strategy; 
E2E-JO (End-to-End Joint Optimization) jointly optimizes all trainable modules from the onset of training, without single-branch pre-adaptation or staged freezing; 
TS-FFT (Three-Stage Full Fine-Tuning) retains the single-branch pre-adaptation process in Stage 1 and Stage 2, but unfreezes all parameters for joint fine-tuning in Stage 3; 
and TS-BFRA, the strategy proposed in this study, adapts the SAT and LVD branches in Stage 1 and Stage 2, respectively, and freezes the two domain-adapted backbones in Stage 3 while fine-tuning the remaining trainable parameters. 
Fig.~\ref{fig:Dice-curves} shows the Dice coefficient curves under different training strategies.

\begin{table}[pos=htbp,width=\textwidth]
  \centering
  \caption{Ablation experiments on the TS-BFRA strategy}
    \begin{tabular}{cccc}
    \toprule
    Training Strategy & mDice ($\uparrow$) & mIoU ($\uparrow$) & Note \\
    \midrule
    No strategy  & 70.26\% & 56.40\% & Use SAT backbone \\
    E2E-JO & \underline{70.44\% (+0.18\%)} & \underline{56.54\% (+0.14\%)} & End-to-End Joint Optimization \\
    TS-FFT & 70.32\% (+0.06\%) & 56.40\% (+0.00\%) & Three-Stage Full Fine-Tuning (Unfrozen) \\
    TS-BFRA(Ours) & \textbf{71.01\% (+0.75\%)} & \textbf{57.31\% (+0.91\%)} & Three-Stage Backbone-Frozen Routing Alignment \\
    \bottomrule
    \end{tabular}%
  \label{tab:Ablation-TS-BFRA}%
\end{table}%

As shown in Table~\ref{tab:Ablation-TS-BFRA}, compared with No strategy, E2E-JO improves mDice by 0.18\% and mIoU by 0.14\%, indicating only negligible gains. 
As shown in Fig.~\ref{fig:Dice-curves}(c), the training Dice of E2E-JO continues to increase, whereas the validation Dice reaches a premature peak and subsequently plateaus, without improving synchronously with the training performance. 
This indicates that although end-to-end joint optimization can continue to fit the training set, its improvement on validation performance is limited, suggesting a certain tendency toward overfitting. 
One possible reason is that, during the early training stage, a large number of parameters are jointly optimized before CA-DSGF has completed data-domain adaptation. 
As a result, the gating network needs to learn fusion weights within a highly non-stationary feature space, making it difficult to form stable feature-fusion ratios.

Compared with No strategy, TS-FFT improves mDice by 0.06\%, while mIoU remains unchanged. 
Compared with E2E-JO, TS-FFT performs independent domain adaptation for the SAT and LVD branches in the first two stages; however, the joint fine-tuning in Stage 3 may disrupt the stability of the pre-adapted dual-branch feature spaces, making it difficult for the gated fusion module to complete effective alignment on stable representations. 
Therefore, TS-FFT still fails to fully exploit the capacity of CA-DSGF.

In contrast, TS-BFRA achieves the best results, improving mDice by 0.75\% and mIoU by 0.91\% compared with No strategy. 
As shown in Fig.~\ref{fig:Dice-curves}(a) and Fig.~\ref{fig:Dice-curves}(b), TS-BFRA preserves the best validation-performing single-branch backbone checkpoints in the first two stages and freezes the backbone parameters in Stage 3, directing subsequent optimization mainly toward the remaining non-backbone parameters. 
This strategy can reduce the risk of continued overfitting or feature drift in the backbones during joint fine-tuning, enabling routing alignment in a relatively stabilized feature space during Stage 3 and thereby achieving better performance.

In summary, the experimental results indicate that TS-BFRA is better suited to the CA-DSGF architecture. 
Therefore, this study retains this strategy for the Overall Ablation Study in Table~\ref{tab:Overall-Ablation}.

\subsubsection{Ablation Study of CS-MED}
\label{CS-MED-study}

In a multi-expert decoding structure, both expert types and topological configurations can affect the final performance. 
To validate the effectiveness of the CS-MED design, this study organizes the ablation experiments into two parts: selecting the expert candidate pool and comparing different CS-MED topological configurations.

For the expert candidate pool, this study selects seven representative decoding architectures, including DPT, SAM-MLP, D2S, UperNet, ASPP, SegFormer, and Linear\_Attn. 
To control variables, this experiment uniformly uses SAT weights for the backbone and independently evaluates the standalone performance of each decoder. 
The results are shown in Table~\ref{tab:Ablation-Decoders}.

\begin{table}[pos=htbp,width=\textwidth]
  \centering
  \caption{Ablation Study of Individual Decoders}
    \begin{tabular}{ccc}
    \toprule
    Decoder type & mDice($\uparrow$) & mIoU($\uparrow$) \\
    \midrule
    SAM-MLP & \underline{70.30\%} & \underline{56.48\%} \\
    D2S   & 70.05\% & 56.19\% \\
    UperNet & 53.20\% & 39.32\% \\
    ASPP  & 68.79\% & 54.81\% \\
    SegFormer & 70.00\% & 56.21\% \\
    DPT   & 70.23\% & \textbf{56.52\%} \\
    Linear\_Attn & \textbf{70.34\%} & 56.47\% \\
    \bottomrule
    \end{tabular}%
  \label{tab:Ablation-Decoders}%
\end{table}%

As shown in Table~\ref{tab:Ablation-Decoders}, different decoders show distinct performance profiles on ML-SASD. 
Among them, Linear\_Attn achieves the highest mDice of 70.34\%, while DPT obtains the highest mIoU of 56.52\%. 
SAM-MLP ranks second in both mDice and mIoU, indicating relatively stable decoding performance. In contrast, UperNet and ASPP achieve relatively lower results, suggesting that some classical decoding structures show limited compatibility with DINOv3 features. 
Considering both standalone decoding performance and functional complementarity among experts, this study finally selects DPT, SAM-MLP, D2S, and Linear\_Attn to form the expert pool of CS-MED, covering the decoding requirements of multi-scale structure recovery, semantic region discrimination, spatial detail reconstruction, and long-range dependency modeling, respectively.

\begin{figure}[pos=htbp,width=\textwidth]
	\centering
	\includegraphics[width=0.8\textwidth]{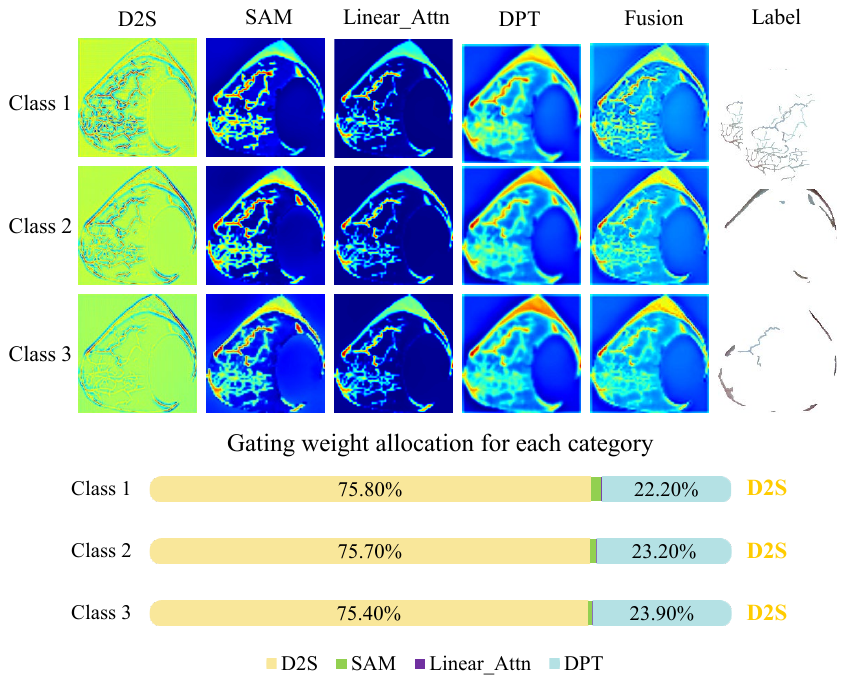}
	\caption{Visualization of expert feature activations and class-specific gating weights in CS-MED}
	\label{fig:Visualization-CS-MED}
\end{figure}

After establishing the expert pool, this study further compares different CS-MED topological configurations to analyze how the selected experts can be more effectively organized for class-specific pixel decoding. 
Specifically, the following experiments evaluate whether class-specific projection layers, class-specific gating networks, and class-specific expert sets can better adapt the expert pool to the structural and morphological heterogeneity of different anomaly categories. 
In addition, Fig.~\ref{fig:Visualization-CS-MED} provides a qualitative visualization analysis of expert activations and gating weights, further illustrating how different topological configurations affect expert utilization and class-specific decoding behavior.

\begin{table}[pos=htbp,width=\textwidth]
  \centering
  \caption{Ablation results of the three CS-MED configurations}
    \begin{tabular}{cccc}
    \toprule
    Category & Configuration & mDice($\uparrow$) & mIoU($\uparrow$) \\
    \midrule
    \multirow{4}[2]{*}{LVD} & DPT   & 70.26\% & 56.40\% \\
          & Shared Proj. + Multi-MoE & 70.94\% (+0.68\%) & 57.04\% (+0.64\%) \\
          & Multi-Proj. + Multi-MoE & \underline{71.04\% (+0.78\%)} & \underline{57.16\% (+0.76\%)} \\
          & Shared Proj. + Shared MoE & \textbf{71.11\% (+0.85\%)} & \textbf{57.27\% (+0.87\%)} \\
    \midrule
    \multirow{4}[2]{*}{SAT} & DPT   & 70.18\% & 56.39\% \\
          & Shared Proj. + Multi-MoE & \underline{71.27\% (+1.09\%)} & \underline{57.45\% (+1.06\%)} \\
          & Multi-Proj. + Multi-MoE & 70.87\% (+0.69\%) & 57.11\% (+0.72\%) \\
          & Shared Proj. + Shared MoE & \textbf{71.34\% (+1.16\%)} & \textbf{57.60\% (+1.21\%)} \\
    \bottomrule
    \end{tabular}%
  \label{tab:CS-MED-configurations}%
\end{table}%

Fig.~\ref{fig:Visualization-CS-MED} intuitively reveals the specialized roles and collaborative fusion mechanism of different decoding experts in CS-MED. 
It can be observed that experts with different inductive biases exhibit strong complementarity in feature extraction. 
In the expert routing strategy shown in this example, D2S serves as the dominant expert, with DPT acting as a complementary expert. 
For the representative Ve structures in the figure, the sub-pixel reconstruction-based D2S expert shows strong high-frequency detail-capturing capability, with the heatmap focusing on sharp and slender Ve boundaries but appearing relatively fragmented. 
In contrast, the DPT expert provides necessary multi-scale macro-structural support, where the heatmap outlines the overall structure of Ve but is less detailed. 
The gating network captures this image-specific characteristic and allocates a dominant weight fraction, exceeding 97\%, to these two experts. 
A comparison between the ``Fusion'' and ``Label'' panels shows that, after class-aware fusion, the fused feature maps exhibit stronger adaptability to different anomaly categories.

A noteworthy phenomenon is that the gating weights for the three classes in the \textit{Gating weight allocation for each category} panel are almost identical, whereas the final fused feature maps still show clear differences. 
Specifically, the ``Fusion'' map of Class 1 presents a more detailed distribution of Ve structures, while the upper anomaly regions in Class 2 and Class 3 display more intense activations. 
This is mainly because the expert feature maps before fusion have already been class-enhanced, as reflected by the clear differences among Class 1--3 in D2S. 
This class enhancement originates from the class-aware decoupling of the front-end CA-DSGF encoder, which ensures that the decoder receives specialized class-specific feature inputs.

Therefore, at the decoding stage, although the input features are derived from the same image, their semantics have already been decoupled by class. 
In this case, the gating network does not need to perform sharp routing adjustments across different classes, leading to nearly identical gating weights across the three anomaly categories. 
This observation further raises the question of whether the projection layer and the decoder also require class-aware specialization. 
Accordingly, this study designs progressive ablation experiments on ML-SASD-Mix based on the three topological configurations described in Section~\ref{CS-MED}. 
The results of the three configurations under SAT and LVD weights are reported in Table~\ref{tab:CS-MED-configurations}.

As shown in Table~\ref{tab:CS-MED-configurations}, all three configurations achieve different degrees of improvement over the baseline results in Table~\ref{tab:Data-Subsets}. 
However, unlike CA-DSGF, introducing stronger class-aware specialization does not lead to a substantial performance gain in CS-MED. 
Among the three configurations, the most complex Multi-Proj + Multi-MoE architecture ranks second under LVD weights and third under SAT weights. In contrast, Shared Proj + Shared MoE achieves the largest improvement and maintains the leading performance under both pretrained weight settings. 
These results indicate that overly rigid class-wise parameter compartmentalization does not necessarily bring additional benefits; instead, it may weaken cross-class information reuse in the shared latent space and reduce feature transferability under complex multi-source domain shifts. 
Therefore, this study retains the Shared Proj + Shared MoE architecture for the Overall Ablation Study in Table~\ref{tab:Overall-Ablation}.

\subsubsection{Ablation Study of PCP Loss}
\label{PCP-loss-ex}

To validate the suppressive effect of PCP Loss on false positives within SSR regions, this study conducts ablation experiments on its core hyperparameters. 
To isolate variables, the experiment uses SAT weights as the encoder, adopts DPT as the decoder, and keeps the remaining training settings invariant.
The dynamic penalty weight in PCP Loss is defined as:

\begin{center}
\refstepcounter{equation}\label{eq:pcp_weight_ablation}
\makebox[\linewidth]{%
\hfill
\makebox[0pt][c]{%
\(\displaystyle
w(x)
=
1+
(\lambda-1)\cdot
\hat{p}_d(x)^{\gamma}
\cdot
\mathbf{M}_g(x)
\)
}
\hfill
\llap{(\theequation)}%
}
\end{center}
where \(\lambda\) denotes the maximum penalty magnitude, and \(\gamma\) controls how rapidly the penalty changes as the prediction confidence increases. This study performs a grid-based ablation over \(\lambda\) and \(\gamma\), with the candidate sets defined as \(\lambda\in\{2,3,4\}\) and \(\gamma\in\{0.5,1.0,2.0\}\), respectively. The main objective of this experiment is to reduce mGFPR while maintaining relatively stable mIoU.

\begin{figure}[pos=htbp,width=\textwidth]
	\centering
	\includegraphics[width=0.8\textwidth]{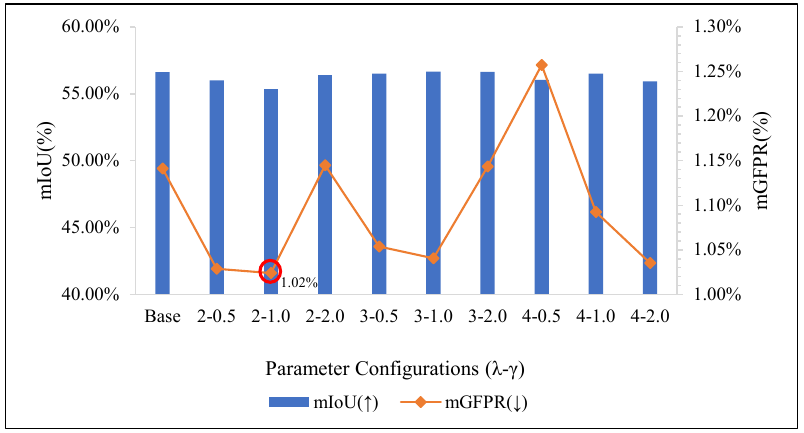}
	\caption{Effect of PCP Loss parameters on mIoU and mGFPR}
	\label{fig:PCP-parameters}
\end{figure}

\begin{table}[pos=htbp,width=\textwidth]
  \centering
  \caption{Ablation results of PCP Loss hyperparameters}
    \begin{tabular}{ccc}
    \toprule
    Parameter combination & mIoU($\uparrow$) & mGFPR($\downarrow$) \\
    \midrule
    Baseline & 56.63\% & 1.14\% \\
    $\lambda=2,\ \gamma=0.5$ & 56.01\% & \underline{1.03\%} \\
    $\lambda=2,\ \gamma=1.0$ & 55.36\% & \textbf{1.02\%} \\
    $\lambda=2,\ \gamma=2.0$ & 56.42\% & 1.14\% \\
    $\lambda=3,\ \gamma=0.5$ & 56.52\% & 1.05\% \\
    $\lambda=3,\ \gamma=1.0$ & 56.66\% & 1.04\% \\
    $\lambda=3,\ \gamma=2.0$ & 56.65\% & 1.14\% \\
    $\lambda=4,\ \gamma=0.5$ & 56.05\% & 1.26\% \\
    $\lambda=4,\ \gamma=1.0$ & 56.51\% & 1.09\% \\
    $\lambda=4,\ \gamma=2.0$ & 55.93\% & 1.04\% \\
    \bottomrule
    \end{tabular}%
  \label{tab:Ablation-PCP}%
\end{table}%

\begin{figure}[pos=htbp,width=\textwidth]
	\centering
	\includegraphics[width=0.5\textwidth]{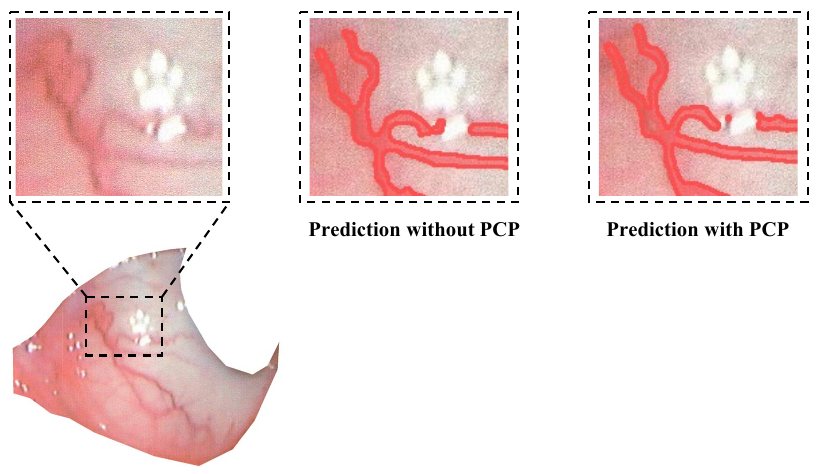}
	\caption{Visual comparison of the effect of PCP Loss on suppressing false positives in reflective regions}
	\label{fig:Visual-PCP}
\end{figure}

As shown in Table~\ref{tab:Ablation-PCP} and Fig.~\ref{fig:PCP-parameters}, after introducing PCP Loss, mIoU remains within a relatively compact range across different parameter combinations, indicating that PCP Loss does not substantially disrupt the baseline segmentation accuracy. 
Meanwhile, different combinations of \(\lambda\) and \(\gamma\) have different effects on mGFPR. 
An excessively large penalty magnitude or an inappropriate decay setting may affect the model's learning of true anomaly regions, whereas a relatively moderate penalty configuration is more conducive to achieving a better trade-off between segmentation fidelity and specular-reflection suppression.

When the parameters are set to \(\lambda=2\) and \(\gamma=1\), the model achieves the strongest interference-suppression performance, with mGFPR reaching the minimum value of 1.02\%. 
Considering the trade-off between segmentation accuracy and specular-reflection suppression, this parameter combination is selected as the final configuration and applied to the Overall Ablation Study in Table~\ref{tab:Overall-Ablation}.

As shown in Fig.~\ref{fig:Visual-PCP}, without PCP Loss, the model tends to produce segmentation spillover near strong specular reflection artifacts, where glare-affected regions around fragmented Ve structures are falsely segmented as anomalous foreground. 
After introducing PCP Loss, these localized false-positive responses are visibly reduced, while the predictions of the main anomaly structures remain largely stable.

\subsubsection{Ablation Study of CA-ASW}
\label{CA-ASW study}

To validate the effectiveness of the adaptive sample-class weighting scheme in CA-ASW, this study compares five weighting strategies on ML-SASD-Mix, namely Easy, Hard, Focal, Curriculum, and Balanced, and further conducts ablation experiments on the related hyperparameters. 
The Base configuration denotes the control baseline without CA-ASW. 
The experimental results are shown in Table~\ref{tab:comparison-strategies}.

\begin{table}[pos=htbp,width=\textwidth]
  \centering
  \caption{Performance comparison of Easy, Balanced, Hard, Focal, and Curriculum strategies}
    \begin{tabular}{cccc}
    \toprule
    Strategy & Parameter & mDice($\uparrow$) & mIoU($\uparrow$) \\
    \midrule
    Base  & -     & 70.26\% & 56.40\% \\
    \midrule
    \multirow{3}[2]{*}{Hard} & t=0.5 & 70.27\% (+0.01\%) & 56.47\% (+0.07\%) \\
          & t=1.0 & 70.34\% (+0.08\%) & 56.56\% (+0.16\%) \\
          & t=2.0 & 70.34\% (+0.09\%) & 56.56\% (+0.16\%) \\
    \midrule
    \multirow{3}[2]{*}{Balanced} & t=0.5 & 70.44\% (+0.19\%) & 56.66\% (+0.25\%) \\
          & t=1.0 & \textbf{70.79\% (+0.54\%)} & \textbf{56.97\% (+0.57\%)} \\
          & t=2.0 & 70.09\% (-0.17\%) & 56.30\% (-0.10\%) \\
    \midrule
    \multirow{3}[2]{*}{Easy} & t=0.5 & 70.37\% (+0.11\%) & 56.49\% (+0.09\%) \\
          & t=1.0 & 70.48\% (+0.22\%) & 56.75\% (+0.35\%) \\
          & t=2.0 & 70.18\% (-0.08\%) & 56.53\% (+0.13\%) \\
    \midrule
    \multirow{3}[2]{*}{Focal} & g=2.0 & 69.90\% (-0.36\%) & 56.03\% (-0.38\%) \\
          & g=3.0 & 69.83\% (-0.43\%) & 55.90\% (-0.51\%) \\
          & g=4.0 & 70.30\% (+0.04\%) & 56.53\% (+0.13\%) \\
    \midrule
    \multirow{9}[2]{*}{Curriculum} & w=5, t=0.5 & 70.14\% (-0.12\%) & 56.30\% (-0.10\%) \\
          & w=5, t=1.0 & 70.42\% (+0.16\%) & 56.72\% (+0.32\%) \\
          & w=5, t=2.0 & 70.66\% (+0.41\%) & 56.89\% (+0.49\%) \\
          & w=10, t=0.5 & \underline{70.68\% (+0.43\%)} & \underline{56.96\% (+0.56\%)} \\
          & w=10, t=1.0 & 70.48\% (+0.23\%) & 56.55\% (+0.15\%) \\
          & w=10, t=2.0 & 70.07\% (-0.19\%) & 56.24\% (-0.16\%) \\
          & w=15, t=0.5 & 70.25\% (-0.01\%) & 56.37\% (-0.03\%) \\
          & w=15, t=1.0 & 70.22\% (-0.04\%) & 56.51\% (+0.11\%) \\
          & w=15, t=2.0 & 69.89\% (-0.36\%) & 56.18\% (-0.22\%) \\
    \bottomrule
    \end{tabular}%
  \label{tab:comparison-strategies}%
\end{table}%

As shown in Table~\ref{tab:comparison-strategies}, different weighting strategies have different effects on model performance. 
The Hard and Focal strategies provide limited overall gains, and some settings even perform worse than the Base configuration. 
This may be because high-loss samples may contain confounding factors such as specular reflection artifacts, out-of-focus regions, or ambiguous annotation boundaries; therefore, indiscriminately increasing the weights of high-loss samples does not necessarily lead to stable improvements. 
The Easy strategy achieves a certain improvement under $t=1.0$, increasing mDice and mIoU by 0.22\% and 0.35\%, respectively, indicating that appropriately emphasizing low-loss samples can help stabilize training under noisy data conditions.

The Curriculum strategy performs well under some parameter settings. Under $w=10$ and $t=0.5$, mDice and mIoU increase by 0.43\% and 0.56\%, respectively. 
This suggests that a training strategy that gradually transitions from easy samples to hard samples can be effective to some extent. 
However, this strategy relies on a predefined training schedule and lacks the flexibility to adapt to the evolving sample-class difficulty recorded in the historical loss matrix.

In contrast, the Balanced strategy achieves the best result in this group of experiments under $t=1.0$, improving mDice and mIoU by 0.54\% and 0.57\%, respectively. 
This strategy uses the class-wise median of the global historical loss matrix as a dynamic anchor.
By measuring each sample-class combination against this global anchor, it moderately suppresses extreme noisy samples while preserving the contribution of relatively difficult samples.

Overall, the Balanced strategy achieves the best overall performance in this group of experiments and removes the need for a predefined training schedule. 
Therefore, this study adopts Balanced under $t=1.0$ as the default weight allocation strategy for CA-ASW and applies it directly to the Overall Ablation Study in Table~\ref{tab:Overall-Ablation}.

\subsubsection{Ablation Study on Stage-wise Injection Strategies for Optimization Objectives}

Since HD-DinoMoE adopts the TS-BFRA three-stage training strategy, the injection stage of different optimization objectives can affect feature adaptation and expert-routing learning.
Therefore, this study further conducts ablation experiments on the injection timing of CA-ASW and PCP Loss, with the tested settings including stage=1\&2, stage=3, and stage=1\&2\&3.
As shown in Table~\ref{tab:Ablation-stage-wise-CA-ASW}, under the selected Balanced configuration with $t = 1$, CA-ASW achieves the best results when injected in stage=1\&2, yielding the highest mDice and mIoU.
By contrast, injecting CA-ASW only in stage=3 or throughout stage=1\&2\&3 achieves smaller improvement margins.
This result suggests that, under the Balanced strategy, CA-ASW is better suited to the backbone domain-adaptation stage.
At this stage, sample-class-level weighting can directly guide the optimization focus toward hard classes and challenging samples during early feature learning.
If CA-ASW is postponed to stage=3, its ability to adjust backbone feature representations is limited because the backbone has already been frozen.

\begin{table}[pos=htbp,width=\textwidth]
  \centering
  \caption{Ablation results of the stage-wise injection strategy for CA-ASW}
    \begin{tabular}{ccccc}
    \toprule
    Strategy & Parameter & MoE-Stage & mDice($\uparrow$) & mIoU($\uparrow$) \\
    \midrule
    Base  & -     & -     & 70.26\% & 56.40\% \\
    \midrule
    \multirow{3}[2]{*}{Balanced} & \multirow{3}[2]{*}{t=1} & 1\&2  & \textbf{71.51\% (+1.25\%)} & \textbf{57.82\% (+1.42\%)} \\
          &       & 3     & 70.79\% (+0.53\%) & 57.17\% (+0.77\%) \\
          &       & 1\&2\&3 & 70.93\% (+0.67\%) & 57.28\% (+0.88\%) \\
    \midrule
    \multirow{3}[2]{*}{Curriculum} & \multirow{3}[2]{*}{w=10, t=0.5} & 1\&2  & 70.28\% (+0.02\%) & 56.56\% (+0.16\%) \\
          &       & 3     & 71.26\% (+1.00\%) & 57.54\% (+1.14\%) \\
          &       & 1\&2\&3 & \underline{71.44\% (+1.18\%)} & \underline{57.80\% (+1.40\%)} \\
    \bottomrule
    \end{tabular}%
  \label{tab:Ablation-stage-wise-CA-ASW}%
\end{table}%

\begin{table}[pos=htbp,width=\textwidth]
  \centering
  \caption{Ablation results of the stage-wise injection strategy for PCP Loss}
    \begin{tabular}{cccc}
    \toprule
    Parameter combine & MoE-Stage & mDice($\uparrow$) & mIoU($\uparrow$) \\
    \midrule
    Base  & -     & 70.26\% & 56.40\% \\
    $\lambda=2,\ \gamma=1.0$ & 1\&2  & 70.30\% (+0.04\%) & 56.61\% (+0.21\%) \\
    $\lambda=2,\ \gamma=1.0$ & 3     & \textbf{70.80\% (+0.54\%)} & \textbf{57.07\% (+0.67\%)} \\
    $\lambda=2,\ \gamma=1.0$ & 1\&2\&3 & \underline{70.42\% (+0.16\%)} & \underline{56.72\% (+0.32\%)} \\
    \bottomrule
    \end{tabular}%
  \label{tab:Ablation-stage-wise-PCP}%
\end{table}%

\begin{table}[pos=htbp,width=\textwidth]
  \centering
  \caption{Overall Ablation Study}
    \begin{tabular}{ccccccccc}
    \toprule
    ID    & A     & B     & C     & D     & mDice($\uparrow$) & mIoU($\uparrow$) & mBF1($\uparrow$) & mGFPR($\downarrow$) \\
    \midrule
    1     &       &       &       &       & 70.26\% & 56.40\% & 38.26\% & 1.14\% \\
    2     & \checkmark     &       &       &       & 71.01\% (+0.75\%) & 57.31\% (+0.91\%) & 39.76\% (+1.50\%) & 1.07\% (-0.07\%) \\
    3     &       & \checkmark     &       &       & 71.50\% (+1.24\%) & 57.86\% (+1.46\%) & 40.34\% (+2.08\%) & \underline{0.91\% (-0.23\%)} \\
    4     &       &       & \checkmark     &       & 70.85\% (+0.59\%) & 57.07\% (+0.67\%) & 36.95\% (-1.31\%) & 1.02\% (-0.12\%) \\
    5     &       &       &       & \checkmark     & 70.79\% (+0.53\%) & 56.97\% (+0.57\%) & 39.00\% (+0.74\%) & 1.06\% (-0.08\%) \\
    6     & \checkmark     & \checkmark     &       &       & 71.64\% (+1.38\%) & 58.12\% (+1.72\%) & 40.11\% (+1.85\%) & 1.04\% (-0.10\%) \\
    7     & \checkmark     &       & \checkmark     &       & 70.80\% (+0.54\%) & 57.07\% (+0.67\%) & 38.70\% (+0.44\%) & \textbf{0.89\% (-0.25\%)} \\
    8     & \checkmark     &       &       & \checkmark     & 71.51\% (+1.25\%) & 57.82\% (+1.42\%) & 39.18\% (+0.92\%) & 1.03\% (-0.11\%) \\
    9     &       & \checkmark     & \checkmark     &       & 71.33\% (+1.07\%) & 57.69\% (+1.29\%) & 40.85\% (+2.59\%) & 1.02\% (-0.12\%) \\
    10    &       & \checkmark     &       & \checkmark     & 71.23\% (+0.97\%) & 57.44\% (+1.04\%) & 40.61\% (+2.35\%) & 1.10\% (-0.04\%) \\
    11    &       &       & \checkmark     & \checkmark     & 70.26\% (+0.00\%) & 56.50\% (+0.01\%) & 38.58\% (+0.32\%) & 0.98\% (-0.16\%) \\
    12    & \checkmark     & \checkmark     & \checkmark     &       & \underline{71.82\% (+1.56\%)} & 58.15\% (+1.75\%) & \underline{41.01\% (+2.75\%)} & 0.97\% (-0.17\%) \\
    13    & \checkmark     & \checkmark     &       & \checkmark     & 71.77\% (+1.51\%) & 58.13\% (+1.73\%) & 40.18\% (+1.92\%) & 0.97\% (-0.17\%) \\
    14    & \checkmark     &       & \checkmark     & \checkmark     & 71.34\% (+1.08\%) & 57.65\% (+1.25\%) & 39.78\% (1.52\%) & 0.98\% (-0.16\%) \\
    15    &       & \checkmark     & \checkmark     & \checkmark     & 71.79\% (+1.53\%) & \underline{58.18\% (+1.78\%)} & 40.81\% (+2.55\%) & 1.01\% (-0.13\%) \\
    16    & \checkmark     & \checkmark     & \checkmark     & \checkmark     & \textbf{72.11\% (+1.85\%)} & \textbf{58.44\% (+2.04\%)} & \textbf{41.40\% (+3.14\%)} & 1.02\% (-0.12\%) \\
    \bottomrule
    \end{tabular}%
  \label{tab:Overall-Ablation}%
\end{table}%

For PCP Loss, Table~\ref{tab:Ablation-stage-wise-PCP} shows that it achieves the best results when injected at stage=3, improving mDice and mIoU by 0.54\% and 0.67\%, respectively.
By contrast, introducing PCP Loss at stage=1\&2 or stage=1\&2\&3 yields smaller gains.
This result suggests that PCP Loss is more suitable for injection after the model has learned relatively stable anomaly representations.
Since PCP Loss mainly imposes pixel-level penalties on high-confidence false positives within specular reflection regions, introducing it during the early training stages may limit its effectiveness, as the predictive confidence profiles are still unstable.
At stage=3, the model has already completed basic feature adaptation, making the spatial constraints of PCP Loss on reflection-induced false positives more effective.

In summary, this study adopts the stage-wise injection configuration selected through ablation: CA-ASW is enabled at stage=1\&2 for sample-class-level reweighting of training contributions, whereas PCP Loss is enabled at stage=3 for false-positive suppression in specular reflection regions.
This configuration is used in the Overall Ablation Study in Table~\ref{tab:Overall-Ablation}.

\subsubsection{Overall Ablation Study}
\label{Overall-Ablation-Study}

After independently validating each submodule and selecting the corresponding parameters, this study further conducts an overall ablation study on ML-SASD-Mix. 
The experiment examines the standalone contributions and combined configurations of four modules: A denotes the CA-DSGF encoder, B denotes the CS-MED expert decoder, C denotes PCP Loss, and D denotes CA-ASW. 
The overall ablation results are shown in Table~\ref{tab:Overall-Ablation}.

From the single-module results, A, B, C, and D all lead to performance changes to different extents. Among them, module B provides the most pronounced improvement, increasing mDice, mIoU, and mBF1 by 1.24\%, 1.46\%, and 2.08\%, respectively, indicating that CS-MED plays an important role in decoding fidelity and boundary localization. Module A also brings stable improvements, suggesting that CA-DSGF can improve the effectiveness of dual-stream feature fusion. Module C mainly reduces mGFPR, but its individual use is accompanied by a decrease in mBF1, indicating that PCP Loss is more oriented toward suppressing false positives in specular reflection regions and needs to be used together with structural modules. Module D brings smaller but consistent improvements in region-level metrics, suggesting that CA-ASW contributes to sample-class-level allocation of training contributions.

From the combination experiments, the structural module combination A+B achieves a substantial overall improvement, increasing mDice and mIoU by 1.38\% and 1.72\%, respectively.
By contrast, combining only the training-objective modules C+D yields limited improvements in region-level metrics.
This indicates that, in ML-SASD, loss reweighting alone is insufficient to fully address multi-source heterogeneity and class-wise morphological disparities, and that encoder and decoder structures remain the main sources of performance improvement.

Further observation of the three-module combinations shows that A+B+C, A+B+D, and B+C+D all achieve favorable performance.
Among them, combinations containing module B consistently achieve better results, further indicating the importance of CS-MED in the overall architecture.
After being combined with other modules, PCP Loss can reduce mGFPR, but its primary effect remains focused on suppressing reflection-induced false positives rather than independently improving all dense segmentation metrics.

Finally, the complete model A+B+C+D achieves the best overall segmentation performance, with mDice, mIoU, and mBF1 reaching 72.11\%, 58.44\%, and 41.40\%, respectively, yielding improvements of 1.85\%, 2.04\%, and 3.14\% over Base.
Meanwhile, the complete model obtains an mGFPR of 1.02\%, which is lower than the 1.14\% of Base.
These results indicate that the complete HD-DinoMoE improves region-level segmentation performance and boundary delineation fidelity while maintaining suppression of reflection-induced false positives.

In summary, the overall ablation study shows that CA-DSGF, CS-MED, PCP Loss, and CA-ASW contribute from four complementary aspects: feature fusion, expert decoding, reflection suppression, and sample-class-level weighting.
When used together, the four modules achieve the best overall segmentation results; therefore, this study adopts A+B+C+D as the final configuration of HD-DinoMoE.

\subsection{Generalization Experiment}

\begin{table}[pos=htbp,width=\textwidth]
  \centering
  \caption{Generalization Experiment Results}
    \begin{tabular}{ccccc}
    \toprule
    Category & Model & Dice($\uparrow$) & IoU($\uparrow$) & BF1($\uparrow$) \\
    \midrule
    \multirow{2}[2]{*}{1} & U-Net~\citep{ronnebergerUnet2015} & 38.87\% & 24.64\% & 37.75\% \\
          & nnU-Net~\citep{isenseeNnUNet2021} & 46.22\% & 30.86\% & 46.07\% \\
    \midrule
    \multirow{2}[2]{*}{2} & TransUnet~\citep{chenTransUNet2024} & 41.19\% & 26.61\% & 42.52\% \\
          & SegNeXt~\citep{NEURIPS2022_08050f40}  & 39.96\% & 25.33\% & 38.04\% \\
    \midrule
    \multirow{2}[2]{*}{3} & U-Mamba~\citep{maUmamba2024} & 45.45\% & 30.03\% & 38.04\% \\
          & U-KAN~\citep{liUKAN2025} & 48.41\% & 32.42\% & 38.57\% \\
    \midrule
    \multirow{4}[2]{*}{4} & Segdino-sat~\citep{yangSegDINO2025} & 40.23\% & 25.90\% & 32.04\% \\
          & Segdino-lvd~\citep{yangSegDINO2025} & 39.84\% & 25.65\% & 32.37\% \\
          & DINOUnet-sat~\citep{gaoDino2025} & \underline{48.54\%} & \underline{33.05\%} & 38.47\% \\
          & DINOUnet-lvd~\citep{gaoDino2025} & 47.79\% & 32.48\% & \underline{47.58\%} \\
    \midrule
    \multirow{2}[2]{*}{5} & UTANet~\citep{luoRethinking2025} & 43.61\% & 28.46\% & 38.07\% \\
          & ConDSeg~\citep{leiConDSeg2025} & 46.28\% & 31.54\% & 40.53\% \\
    \midrule
    6     & HD-DinoMoE(Ours) & \textbf{49.13\% (+0.59\%)} & \textbf{33.45\% (+0.40\%)} & \textbf{53.09\% (+5.51\%)} \\
    \bottomrule
    \end{tabular}%
  \label{tab:Generalization-Experiment}%
\end{table}%

\begin{figure}[pos=htbp,width=\textwidth]
	\centering
	\includegraphics[width=0.4\textwidth]{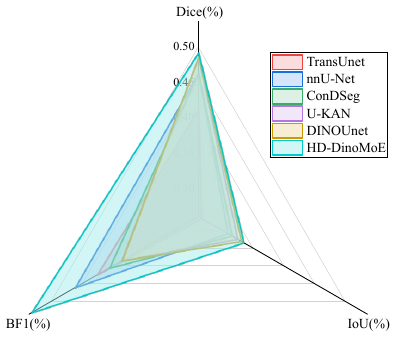}
	\caption{Radar chart comparison of the best-performing models from each group}
	\label{fig:Radar-group-SBPVI}
\end{figure}

\begin{figure}[pos=htbp,width=\textwidth]
	\centering
	\includegraphics[width=0.9\textwidth]{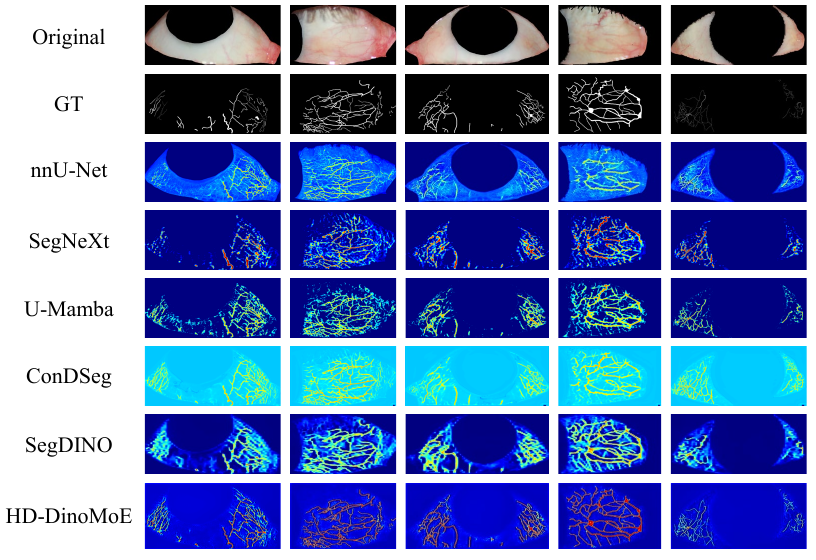}
	\caption{Comparison of prediction heatmaps on the SBVPI Vessels dataset}
	\label{fig:Comparison-SBVPI-Ve}
\end{figure}

Given the scarcity of public benchmarks for scleral anomaly segmentation, this study introduces the public SBVPI dataset to evaluate model performance under an external-domain data distribution.
Since the Vessels subset in SBVPI provides scleral vasculature annotations and is most closely aligned with the target task of this study, only this subset is extracted for testing in this section.
The experimental configuration is kept consistent with the benchmark evaluation protocol in Section~\ref{Comparison-Methods}, and Dice, IoU, and BF1 are adopted as the evaluation metrics.
The experimental results are shown in Table~\ref{tab:Generalization-Experiment}.

As shown in Table~\ref{tab:Generalization-Experiment}, HD-DinoMoE achieves the highest Dice and IoU on the SBVPI Vessels subset, reaching 49.13\% and 33.45\%, respectively.
Compared with the closest-performing DINOUnet-SAT, HD-DinoMoE improves Dice and IoU by 0.59\% and 0.40\%, respectively.
This result indicates that the proposed model can still maintain favorable region-level segmentation performance on external Vessels data.

For the boundary metric BF1, HD-DinoMoE shows a more evident advantage, reaching 53.09\%, exceeding the second-best DINOUnet-LVD by 5.51 percentage points.
This indicates that HD-DinoMoE provides better boundary localization capability for Vessels structures in cross-dataset testing.
Considering the domain discrepancies between SBVPI and ML-SASD in image sources, acquisition environments, and annotation styles, these results suggest that the proposed model has a certain degree of external-domain generalization potential for Vessels structure segmentation.

Compared with the other baseline models, the DINOv3 variants and U-KAN show strong performance in region-level metrics, suggesting that visual foundation model features and emerging network architectures are helpful for cross-domain feature transfer.
However, HD-DinoMoE achieves better results in both region-overlap and boundary metrics, indicating that the joint design of CA-DSGF and CS-MED is not limited to ML-SASD, but also maintains relatively stable performance on the public Vessels subset.

To further observe the prediction behavior of the model on external Vessels data, Fig.~\ref{fig:Comparison-SBVPI-Ve} presents the heatmap visualization results on the SBVPI Vessels subset.
It can be observed that some models can localize the main Vessels regions, but they still exhibit response fragmentation in fine branches, insufficient boundary continuity, and background activation in some cases, such as the brighter overall background response of ConDSeg.
By contrast, HD-DinoMoE produces more contiguous activation responses along the Vessels network, with relatively lower responses in background regions.
This observation is consistent with the advantage of HD-DinoMoE in the BF1 metric reported in Table~\ref{tab:Generalization-Experiment}, suggesting that the proposed method has better boundary preservation capability in external Vessels structure segmentation.

In summary, the SBVPI external experiment shows that HD-DinoMoE still achieves favorable Vessels segmentation results under cross-domain distribution shift, with a particular advantage in boundary quality.
Since SBVPI only provides Vessels-related annotations, this experiment mainly validates the external generalization capability of the model for vascular anomaly segmentation.
For the cross-dataset generalization of Yellow and Black Spots and Blood Spots, more extensive publicly annotated data will still be required for comprehensive validation in future work.

\section{Conclusion}

This study proposes the technical framework of the TAO system to support the intelligent and quantifiable application of TCM ocular inspection.
This framework consists of four components: ocular image acquisition, scleral anomaly segmentation, eye-organ mapping and regional scoring, and knowledge-enhanced report generation.
The present study mainly focuses on ocular image acquisition and scleral anomaly segmentation, while providing technical planning for the subsequent development of the third and fourth components.

The main work of this study includes two aspects: data-basis construction and segmentation-model design.
At the data level, ML-SASD advances scleral image analysis from conventional scleral contour extraction to  multi-label surface-anomaly parsing, enabling Ve, YBS, BS, and reflection/glare interference to be modeled and evaluated within a unified annotation framework.
At the model level, HD-DinoMoE constructs a class-aware modeling framework from encoding representations, decoding paths, and training losses to address multi-source heterogeneity, class-wise morphological differences, and reflection interference in complex acquisition scenarios.
Compared with methods that rely only on a single vision backbone or a fixed decoding structure, this design places greater emphasis on fine-grained differentiated processing for different anomaly classes in terms of feature representation, boundary morphology, and training difficulty.

The experimental results show that HD-DinoMoE achieves strong overall segmentation performance on ML-SASD-Mix and outperforms most representative segmentation methods in region overlap, boundary quality, and reflection-induced false-positive control.
The comparison experiments and ablation studies further indicate that DINOv3-based visual foundation features can provide effective and transferable dense representations for scleral anomaly segmentation, while class-aware dual-stream fusion, multi-expert decoding, reflection-region penalization, and sample-class-level weighting jointly improve the model's adaptability to complex anomaly morphologies and reflection interference.
The generalization experiment on the public SBVPI dataset also shows that the proposed method has a certain degree of cross-dataset transferability for Vessels-related anomaly segmentation.
However, because existing public datasets lack annotations for other anomaly categories such as YBS and BS, external generalization validation for these two categories remains limited in this study.

Future work will be advanced along three directions: data expansion, model deployment, and further improvement of the TAO system.
First, the scale of ML-SASD will be further expanded to cover more acquisition devices, illumination conditions, age distributions, and clinical sources.
External validation data for different scleral anomaly categories will also be supplemented to improve the comprehensiveness of model evaluation.
Second, considering the relatively high computational footprint of the dual DINOv3 backbones and MoE-based decoding structure in HD-DinoMoE, future work will explore algorithmic compression, model lightweighting, knowledge distillation, parameter-efficient fine-tuning, and mobile deployment strategies to reduce computational cost in practical applications.
Finally, at the TAO system level, future studies will further improve the polar-coordinate eye-organ mapping and regional scoring mechanisms, and validate their stability and interpretability by incorporating physician annotations, clinical priors, and follow-up data.
Meanwhile, a knowledge base related to TCM ocular inspection will be constructed, and Retrieval-Augmented Generation (RAG) or multimodal large language models (LLMs) will be integrated to generate auxiliary analytical reports.
The consistency, readability, and clinical reference value of these reports will be further evaluated through physician assessment.

Overall, this study offers a feasible technical path for supporting the transition of TCM ocular inspection from empirical observation toward intelligent and quantifiable auxiliary analysis.




\section{Data availability}
The source code, annotation protocols, data splits, evaluation scripts, and releasable ML-SASD data materials are available at \url{https://github.com/FX-CMX/HD-DinoMoE}. 
For prospectively collected ocular images, all participants provided informed consent for the use of their original ocular images and pixel-wise annotations in academic research, manuscript publication, and public dataset release. 
For data materials involving third-party source images, access is subject to the permissions associated with the original sources and may be provided upon reasonable request where redistribution is permitted.

\section*{Declaration of generative AI and AI-assisted technologies in the manuscript preparation process}

During the preparation of this work, the authors used ChatGPT to assist with language polishing, grammar checking, wording refinement, LaTeX-related troubleshooting, and manuscript organization. After using this tool, the authors reviewed and edited the content as needed and take full responsibility for the content of the published article.

\printcredits

\bibliographystyle{cas-model2-names}

\bibliography{eye6}



\end{document}